\documentclass{article}

\usepackage[main, final]{neurips_2026}

\usepackage[utf8]{inputenc} 
\usepackage[T1]{fontenc}    
\usepackage{hyperref}       
\usepackage{url}            
\usepackage{booktabs}       
\usepackage{amsfonts}       
\usepackage{nicefrac}       
\usepackage{microtype}      
\usepackage{xcolor}         

\usepackage{amsmath,amsfonts,bm}

\def\eqref#1{equation~\ref{#1}}

\def\1{\bm{1}}

\DeclareMathAlphabet{\mathsfit}{\encodingdefault}{\sfdefault}{m}{sl}
\SetMathAlphabet{\mathsfit}{bold}{\encodingdefault}{\sfdefault}{bx}{n}

\newcommand{\R}{\mathbb{R}}

\usepackage{url}
\usepackage{todonotes}
\usepackage{acro}
\usepackage{booktabs}
\usepackage{csquotes}
\usepackage{multirow}
\usepackage{siunitx}
\usepackage{tcolorbox}
\tcbuselibrary{breakable,listings,skins}
\tcbset{
  grokstyle/.style={colback=orange!8, colframe=orange!70},
  geministyle/.style={colback=teal!8, colframe=teal!70},
  deepseekstyle/.style={colback=blue!5, colframe=blue!70},
  gptstyle/.style={colback=olive!8, colframe=olive!65},
  opusstyle/.style={colback=purple!6, colframe=purple!65},
}
\acsetup{make-links=false}
\usepackage{newfloat}      
\usepackage{caption} 
\usepackage{subcaption}
\usepackage{minted}        
\usepackage{xcolor}        
\usepackage{titletoc}
\usepackage{csquotes}
\usepackage{wrapfig}

\usepackage{algorithm}
\usepackage{algpseudocode}
\usepackage{amssymb}

\newfloat{trace}{tbp}{cod}
\floatname{trace}{Trace}
\definecolor{lightgray}{gray}{0.95}
\setminted{
  bgcolor=lightgray,
  fontsize=\small,
  breaklines=true,
  linenos=false
}
\DeclareAcronym{rl}{
  short = RL,
  long  = reinforcement learning
}
\DeclareAcronym{zsc}{
    short = ZSC,
    long = zero-shot coordination
}
\DeclareAcronym{llm}{
    short = LLM,
    long = large language model,
}

\DeclareAcronym{sp}{
    short = SP,
    long = self-play
}

\DeclareAcronym{xp}{
    short = XP,
    long = cross-play
}

\DeclareAcronym{icl}{
    short = ICL,
    long = in-context learning
}

\DeclareAcronym{xp_mat}{
    short = XP-matrix,
    short-plural-form = XP-matrices,
    long = cross-play matrix,
    long-plural-form = cross-play matrices
}

\DeclareAcronym{sp_score}{
    short = SP-score,
    long = self-play score,
}

\DeclareAcronym{xp_score}{
    short = XP-score,
    long = cross-play score,
}

\DeclareAcronym{cfd}{
    short = CFD,
    long = coordination-friendly definition,
}

\DeclareAcronym{mdp}{
    short = MDP,
    long = Markov decision process,
    long-plural = es
}

\DeclareAcronym{dec_pomdp}{
    short = Dec-POMDP,
    long = Decentralised Partially Observable Markov Decision Process,
    long-plural = es
}

\DeclareAcronym{pomdp}{
    short = POMDP,
    long = partially observable Markov decision process,
    long-plural = es
}

\DeclareAcronym{op}{
    short = OP,
    long = other-play
}

\DeclareAcronym{obl}{
    short = OBL,
    long = off-belief learning
}

\DeclareAcronym{aoh}{
    short = AOH,
    long = action-observation history
}

\DeclareAcronym{nlp}{
    short = NLP,
    long = natural language processing
}

\DeclareAcronym{sft}{
    short = SFT,
    long = supervised fine-tuning
}

\DeclareAcronym{rlhf}{
    short = RLHF,
    long = reinforcement learning from human feedback
}

\DeclareAcronym{grpo}{
    short = GRPO,
    long = group relative policy optimisation
}

\DeclareAcronym{ppo}{
    short = PPO,
    long = proximal policy optimisation 
}

\DeclareAcronym{dpo}{
    short = DPO,
    long = discrete prompt optimisation
}

\DeclareAcronym{vlm}{
    short=VLM,
    long=vision-language model
}
\DeclareAcronym{nle}{
    short=NLE,
    long= NetHack learning environment
}

\DeclareAcronym{ape}{
    short=APE,
    long=Automatic Prompt Engineer
}

\DeclareAcronym{ts-mp}{
    short=TS-MP,
    long=too-strict meta-problem,
    long-plural-form=too-strict meta-problems

}

\DeclareAcronym{cf-mp}{
    short=CF-MP,
    long=coordination-failure meta-problem,
    long-plural-form=coordination-failure meta-problems
}

\DeclareAcronym{tom}{
    short=ToM,
    long=theory of mind,
}

\DeclareAcronym{cai}{
    short=CAI,
    long=Constitutional AI,
}

\newcommand{\acb}[1]{\acl{#1} (\acs{#1})}
\newcommand{\acbp}[1]{\aclp{#1} (\acsp{#1})}
\newcommand{\Acb}[1]{\Acl{#1} (\acs{#1})}
\newcommand{\Acbp}[1]{\Aclp{#1} (\acsp{#1})}

\newcommand{\sxpgap}{\acs{sp}-\acs{xp} gap}

\newcommand{\ie}{i.e.\ }

\renewcommand{\sectionautorefname}{Section}
\renewcommand{\subsectionautorefname}{Section}
\renewcommand{\subsubsectionautorefname}{Section}

\newcommand{\appref}[1]{Appendix~\ref{#1}}

\newcommand{\meansem}[2]{\ensuremath{#1\:\!{\scriptstyle\pm#2}}}

\usepackage{tikz}
\usetikzlibrary{calc}

\newcommand{\payoff}[2]{%
  \ifcase\numexpr#1-1\relax
    \ifcase\numexpr#2-1\relax 1.0\or 0.0\or 1.01\or 0.0\fi
  \or
    \ifcase\numexpr#2-1\relax 1.01\or 1.0\or 0.0\or 0.0\fi
  \or
    \ifcase\numexpr#2-1\relax 0.0\or 0.0\or 1.0\or 1.01\fi
  \or
    \ifcase\numexpr#2-1\relax 0.0\or 1.01\or 0.0\or 1.0\fi
  \fi
}

\title{Zero-Shot Coordination among LLM Agents}

\author{%
  Adrian Hayler\thanks{Equal contribution.},
  Shashank Reddy Chirra\footnotemark[1],
  Andrei Lupu, \\
  \textbf{Johannes Forkel, Bidipta Sarkar, Siheng Feng, Jakob Foerster} \\
  BOLD, University of Oxford
}

\begin{document}

\maketitle

\begin{abstract}
We study zero-shot coordination (ZSC), where independently developed agents must coordinate at test time. While ZSC has been well studied in the RL literature, far less is known about the performance of LLM agents despite their increasing deployment in such settings. Existing work on LLM coordination often relies on specialised scaffolding that independently developed agents are unlikely to share in practice. Moreover, evaluations on complex environments (e.g., Hanabi) make it difficult to pinpoint the sources of coordination failure. In contrast, we focus on simple, general-purpose scaffolds in minimal environments designed to isolate specific coordination challenges. Our results show that even in these controlled settings, frontier LLM agents struggle to coordinate, largely due to limited understanding of the coordination problem and weak reasoning about their partner’s beliefs. While LLMs introduce \textit{semantic information} as an additional axis for coordination, they nonetheless fail to exploit this structure effectively. Towards this, we propose \textit{\Acbp{cfd}} as a principled approach for enabling robust coordination among LLM agents. Finally, we show that \acp{cfd} can be discovered automatically, removing the need for manual engineering. Code is available at \url{https://github.com/ahayler/zsc-llm-agents}. 
\end{abstract}

\section{Introduction}
\label{sec:intro}

Rapid advances in AI have led to the deployment of autonomous agents in environments that require robust coordination with humans and other independently developed agents \citep{human_ai_coordination_dafoe, robotaxis_kirkham, browser_use_mueller, other_play_hu}. While humans often readily coordinate with unseen partners by reasoning about their behaviour, referred to as \textit{\acb{tom}} \citep{theory_of_mind_premack}, autonomous agents struggle to do the same \citep{decrypto_lupu, foerster_waymo_twitter_video}. A classic real-world example is that of autonomous driving, where vehicles developed independently by companies such as Waymo, Tesla, and Wayve must coordinate during on-road interactions~\citep{robotaxis_kirkham}. 

With the rapid growth in \acbp{llm} capabilities, \acs{llm} agents are increasingly deployed in real-world applications that require coordination with other independently developed agents. For example, in AI-assisted coding software such as Cursor, agents powered by different \acs{llm} providers are used together to collaboratively generate code \citep{cursor_scaling_agents_2026}. Similar settings also arise in robotics \cite{ji2026genswarm, Liang2022CodeAP, szot2024large} and autonomous driving \cite{ishida2024langprop}, where \acs{llm} agents generate policies as code that must coordinate with independently developed partners at deployment.


This problem is formally known as \Acb{zsc}~\citep{other_play_hu}, where agents are trained independently and evaluated together at test time without prior communication or re-training. Although \acs{zsc} prohibits agents from sharing policies, it permits prior agreement on a game-agnostic learning rule used during training. The central objective of \acs{zsc} is therefore to identify learning rules that encode general coordination principles applicable across a wide range of settings, while avoiding common failure modes. For example, the other-play \cite{other_play_hu} learning rule ensures that the policy is equivariant with respect to the symmetries of the environment.

\acs{zsc} literature has predominantly focused on traditional deep \ac{rl} training pipelines~\citep{other_play_hu, obl_hu, muglich2025expected}. More recently, however, the emergence of \acp{llm} has given rise to a new class of autonomous agents~\citep{browser_use_mueller, surveywebagents_ning, alphaevolve} that differ substantially from this paradigm. Rather than requiring a large number of environment interactions \cite{forkel2025entropy}, \acs{llm} agents can coordinate zero-shot from a textual description of the environment via \acb{icl}, without any task-specific training. These textual descriptions expose an additional axis for coordination through the \textit{semantics} of the environment, a signal that has proven difficult for traditional \acs{rl} agents to exploit \citep{ma2023learningintuitivepoliciesusing}. As illustrated in \autoref{fig:combined2_new}, agents can use semantic cues (e.g., the “red arrow”) to coordinate their actions. 


We present the first systematic study of \acs{llm} agents in the \acs{zsc} setting, beginning with a formalisation of two of the most common scaffolds: text agents and code agents. We then introduce a suite of simplified cooperative environments, inspired by prior work \citep{baysian_action_decoder_foerster, obl_hu, trajectory_diversity_lupu, hu2022selfexplainingdeviationscoordination}, each designed to isolate a specific coordination challenge. We augment these environments with text descriptions to support interaction with LLM agents. Together, these environments cover a broad range of challenges an agent is likely to encounter in the real world. Consequently, we posit that strong performance on this suite is \textbf{necessary but not sufficient} for robust real-world coordination. 
 
We find that both classes of agents struggle to coordinate reliably in these simplified environments, driven in part by a weak understanding of the underlying coordination challenge and weak \acs{tom} abilities. Next, we find that agents have a strong bias toward leveraging \textit{environment semantics} for coordination. This can yield improvements over traditional \acs{rl} agents that struggle to exploit such cues; however, semantics also introduce coordination strategies that compete with existing ones, and agents that fail to consistently arbitrate between the two see degraded coordination. Overall, our results highlight the need for a new perspective on coordination with \acs{llm} agents.

Finally, inspired by the \acs{zsc} principle of a shared learning rule among coordinating entities, we introduce the notion of a \textit{\Acb{cfd}}, a natural language specification of general coordination principles shared across agents. This mirrors analogous approaches in other domains, such as AI safety, where high-level rules are codified as a “constitution” to promote adherence to common safety standards \citep{bai2022constitutionalaiharmlessnessai,Huang_2024}. We show that incorporating a \acs{cfd} improves coordination for both text and code agents across a range of environments, and that meta-learned principles within the \acs{cfd} can outperform hand-crafted ones. Taken together, these results position \acs{cfd} as a principled mechanism for enhancing coordination in \acs{llm} agents and a promising direction for future research.

\section{Background and Notation}
\label{sec:background}
\begin{figure*}[t]
    \centering
    \begin{subfigure}[c]{0.38\textwidth}
        \centering
        \includegraphics[width=\linewidth]{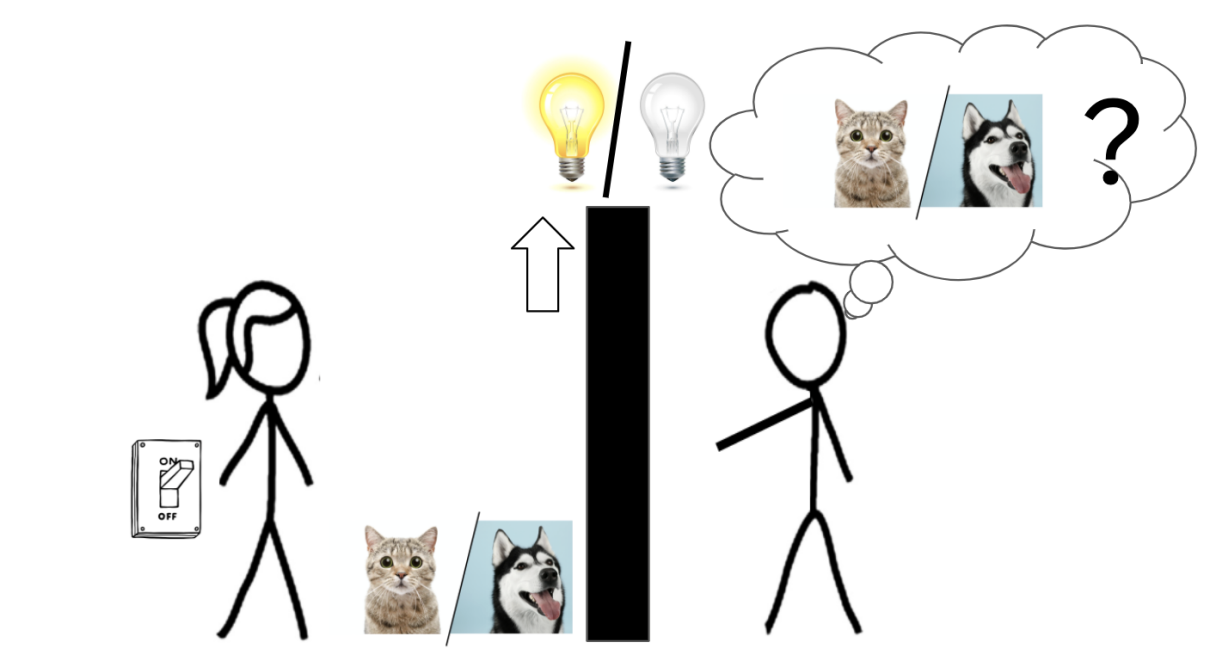}
    \end{subfigure}
    \hspace{0.02\textwidth}
    \begin{subfigure}[c]{0.58\textwidth}
        \centering
        \includegraphics[width=\linewidth]{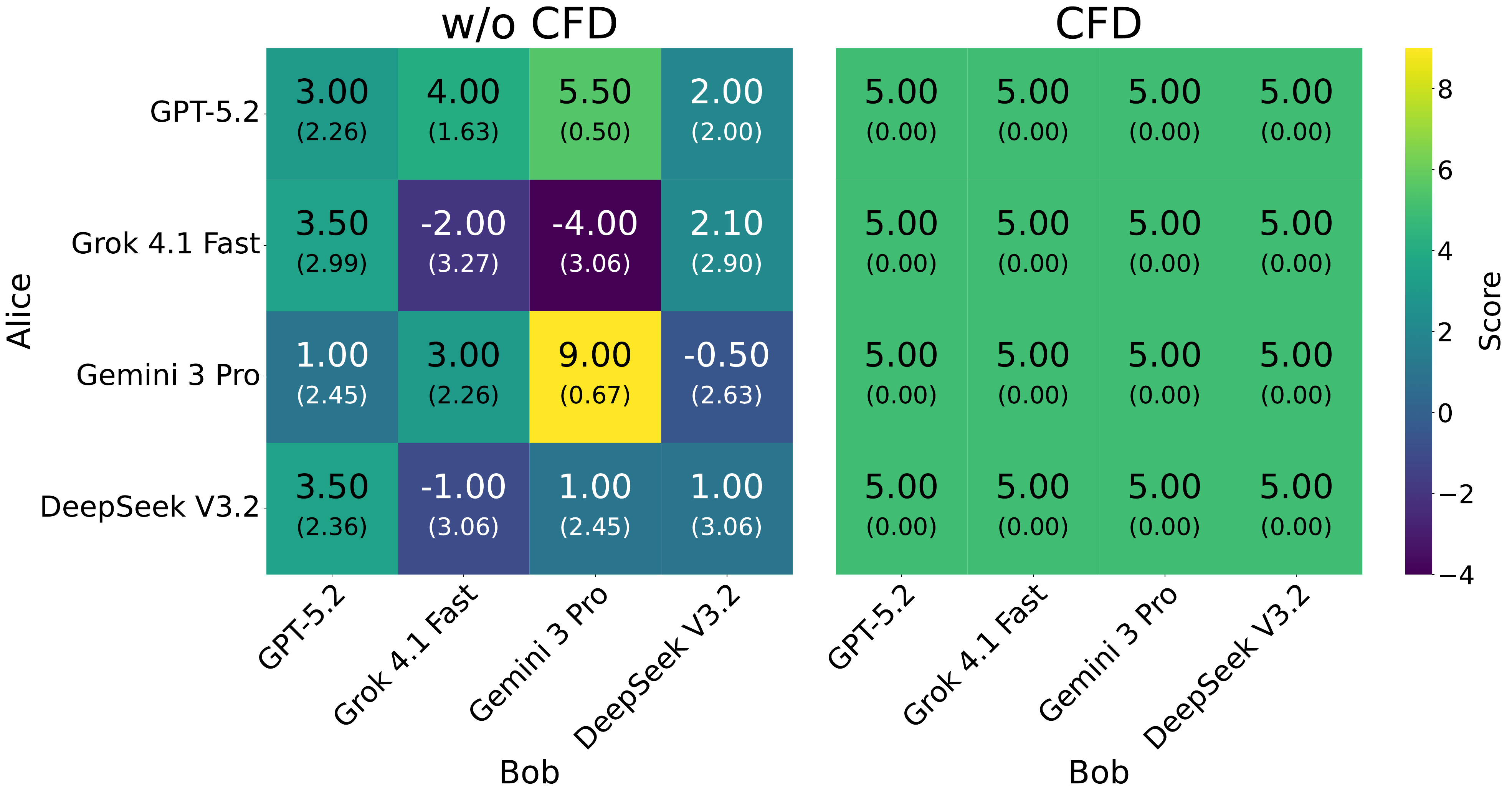}
    \end{subfigure}
    \caption{\textbf{Cat \& Dog coordination game.} \textit{(Left)}. Alice observes an animal (cat/dog). She can either bail and end the game for a joint reward of 1, remove the wall for a cost of 5, or use the light bulb to signal for free. Bob either observes the state of the light bulb or the animal if the wall is removed. He can guess the animal (+10/-10) or bail for a reward of 0.5. \textit{(Right)} XP of text agents in the variable \acs{llm} setting with and without \acs{cfd}. Each cell contains the average score of 10 episodes. Adding the \acs{cfd} leads the agents to consistently pick the safe fallback for a guaranteed reward of $+5$.}
    \label{fig:combined}
    \vspace{-3mm}
\end{figure*}

\begin{figure*}[t]
\centering
\begin{tcolorbox}[
  colback=white,
  colframe=black!70,
  sharp corners,
  boxrule=0.8pt,
  width=\textwidth,
  left=1mm,
  right=1mm,
  top=1mm,
  bottom=1mm,
  before skip=6pt,
  after skip=6pt,
  enhanced,
]

\begin{tcolorbox}[
  breakable,
  geministyle,
  title={Gemini 3 Pro (Alice)},
  fonttitle=\bfseries\small,
  boxrule=0.6pt,
  left=1mm,
  right=1mm,
  top=0.8mm,
  bottom=0.8mm,
  enhanced,
]
\scriptsize\ttfamily
\setlength{\parindent}{0pt}
\raggedright
There is no inherent reason to prefer one over the other, which makes this risky. However, usually, ``Action'' (Light On) maps to ``Option A'' (Cat) and ``No Action'' (Light Off) maps to ``Option B'' (Dog). Let's reconsider the remove\_wall option. A guaranteed 5 is very safe. But a potential 10 is much better. Is there a way to be sure? No. But if I am an AI playing with another AI or a rational player, we likely try to maximize the 10.
\end{tcolorbox}


\begin{tcolorbox}[
  breakable,
  gptstyle,
  title={GPT-5.2 (Bob)},
  fonttitle=\bfseries\small,
  boxrule=0.6pt,
  left=1mm,
  right=1mm,
  top=0.8mm,
  bottom=0.8mm,
  enhanced,
]
\scriptsize\ttfamily
\setlength{\parindent}{0pt}
\raggedright
Alice deliberately turned the light on as a one-bit signal of which animal is present. In a cooperative setting, the natural convention is ``light on = dog'' (active warning/action), so I should guess dog to try to secure the +10 joint reward.
\end{tcolorbox}
\end{tcolorbox}
\caption{\textbf{Reasoning traces from a failed coordination in the \textit{Cat and Dog} environment.} Gemini~3~Pro and GPT-5.2 independently arrive at incompatible strategies.}
\label{trace:cat_and_dog}
\vspace{-6mm}
\end{figure*}

\paragraph{\Acsp{dec_pomdp}} We focus on fully cooperative settings which can be represented via the \acb{dec_pomdp} framework~\citep{bernstein_dec_pomdp, dec_pomdp_book_amato}. A Dec-POMDP $\mathcal{M}$ consists of $M$ agents, each assigned a role $r \in \mathcal{R} = [M]$, a state space $\mathcal{S}$, action space $\mathcal{A}$, per-agent observation spaces $\mathcal{O}_i$, transition function $\mathcal{T}: \mathcal{S} \times \mathcal{A} \rightarrow \Delta(\mathcal{S})$, discount factor $\gamma \in[0,1]$ and horizon $H \in \mathbb{N}$, which represents the maximum length of an episode. Agents share a common reward function $R: \mathcal{S} \times \mathcal{A} \rightarrow \Delta(\mathbb{R})$. Each agent observes the environment via its observation function $Z_i: \mathcal{S} \rightarrow \Delta\left(\mathcal{O}_i\right)$. The action space can be written as a product $\mathcal{A} = A_1 \times \cdots \times A_M$, where $A_i$ corresponds to agent $i$'s action space. A (joint) policy $\pi$ is an $M$-tuple of local policies $\pi_i: \mathcal{H}_i \rightarrow \Delta(A_i)$, where $\mathcal{H}_i$ is the space of possible action-observation histories (\Acsp{aoh}) of player $i$, each of the form $\tau_t^i = \{o_0^i, a_0^i, \ldots, a_{t-1}^i, o_t^i\}$. The joint return is defined as $J(\pi)=\mathbb{E}_{\pi}[\sum_{t=0}^H\gamma^t R_t]$.
We assume access to natural language descriptions $\mathcal{D} = (\mathcal{D}_{\mathcal{M}}, \mathcal{D}_{\mathcal{R}}, \{ \mathcal{D}_{\mathcal{H}_i}\}_{i \in [M]})$ of the \acs{dec_pomdp}. $\mathcal{D}_{\mathcal{M}} \in \Sigma^*$ is a description of the \acs{dec_pomdp}, with $\Sigma^*$ denoting all strings over a suitable finite alphabet $\Sigma$. $\mathcal{D}_{\mathcal{H}_i}: \mathcal{H}_i \rightarrow \Sigma^*$ and $\mathcal{D}_{\mathcal{R}}: \mathcal{R} \rightarrow \Sigma^*$ are injective functions, describing each AOH $\tau_i$ and role $r$ in natural language. We use the term \textit{role} as a synonym for player index; the distinction is purely terminological and reflects how player identity is communicated to LLM agents via $\mathcal{D}_{\mathcal{R}}$.
    



We define an \acs{llm} agent as a system that leverages an \acs{llm} and \acb{icl} to generate a local or joint policy $\pi$ for a \acs{dec_pomdp}. Crucially, \acs{icl} does not involve a separate training phase with gradient-based parameter updates; instead, all the ``learning'' occurs through conditioning on the prompt during the model's forward pass. We consider two classes of LLM agents as illustrated in \autoref{fig:text_and_code_agents}: \textbf{text agents} and \textbf{code agents}. Text agents are assigned to a role $r \in \mathcal{R}$, take as input $(\mathcal{D}_{\mathcal{M}}, \mathcal{D}_{\mathcal{R}}(r), \mathcal{D}_{\mathcal{H}_i}(\tau_i))$ and return an action $a_t^i$, whereas code agents take as input $\mathcal{D}_{\mathcal{M}}$ and return a joint policy $\pi$ as executable code. Since a joint policy specifies the behaviour of all players simultaneously, code agents require no role assignment via $\mathcal{D}_{\mathcal{R}}$; the roles, together with their observations and actions, are already specified in $\mathcal{D}_{\mathcal{M}}$. A worked example of $\mathcal{D}$ for Cat \& Dog is given in \appref{app:worked_example}. Hence, code agents induce a distribution over joint policies. In contrast, text agents do not produce an explicit joint policy; instead, they act on an implicit local policy $\pi^r$ assigned to their role. As a result, their behaviour is harder to interpret or modify, and prompting an \acs{llm} for each action makes them slower and more expensive to evaluate.

\paragraph{Zero-shot Coordination} 

The standard approach for training fully cooperative multi-agent \acs{rl} systems is self-play (\acs{sp})~\citep{checkers_sp_samuel, Tesauro1994TDGammonAS}, wherein agents are trained alongside the same counterparts they will encounter at test time. In contrast, the \acs{zsc} setting~\citep{other_play_hu} assumes that independently trained policies are paired together at test time without prior communication or adaptation, a setting known as cross-play (\acs{xp}). To enable coordination, entities are willing to coordinate on (part of) a Dec-POMDP-agnostic learning rule for training their agents. Examples of \acs{zsc} learning rules make use of known game symmetries~\citep{other_play_hu} or explicitly control for how agents reason about their partner~\citep{obl_hu}.

\section{Problem Formulation}



Consider a \acs{dec_pomdp} with two \acs{llm} agents $i$ and $j$, and write $\pi_i^r$ for the local policy of agent $i$ assigned role $r \in \{1, 2\}$. Code agents independently generate joint policies ${\pi_i,\pi_j}$, with the \ac{sp} score given as $\mathbb{E}_{\pi_i}[J(\pi_i)]$ and $\mathbb{E}_{\pi_j}[J(\pi_j)]$. Their \acs{xp} score is defined as
\begin{align}
\label{eq:xp_definition}
J(i, j) = \tfrac{1}{2}\mathbb{E}_{(\pi_i, \pi_j)}\big[ J(\pi_i^1, \pi_j^2) + J(\pi_j^1, \pi_i^2) \big]
\end{align}

\autoref{eq:xp_definition} also applies for text agents where assigning agent $i$ to role $r$ induces the local policy $\pi_i^r$.

We consider three evaluation variants. \textbf{Self-play} (\acs{sp}) pairs a text agent with itself; for code agents, it is the score of the generated joint policy. The \textbf{fixed \acs{llm}} setting ($\text{XP}_{\text{seed}}$) applies only to code agents, which induce a distribution over joint policies: two policies are sampled independently from the same agent and evaluated under \acs{xp}, measuring compatibility within a single distribution. The \textbf{variable \acs{llm}} setting ($\text{XP}_{\text{LLM}}$), applicable to both code and text agents, instead pairs two distinct \acs{llm} agents under XP. A schematic of the evaluation regimes is provided in \autoref{fig:XP_schematic} in the Appendix.

We treat a single \acs{icl} generation as the analogue of a training run in tabula-rasa \acs{rl}. For code agents, one generated joint policy therefore corresponds to one self-play training run, and pairing independently generated joint policies mirrors standard \acs{xp} evaluation~\citep{other_play_hu}. For text agents, which only induce local policies, the distinction is less clear-cut. We refer to pairing two instances of the same \acs{llm} with the same prompt (but different roles) as \acs{sp}, reflecting that providers may optimise models to work well with instances of themselves, e.g., as subagents within a shared harness. As we show in \autoref{sec:zsc_performance}, even such local policies are not guaranteed to be mutually compatible.

\section{ZSC performance of LLM agents}
\begin{figure*}
    \centering    \includegraphics[width=0.9\linewidth]{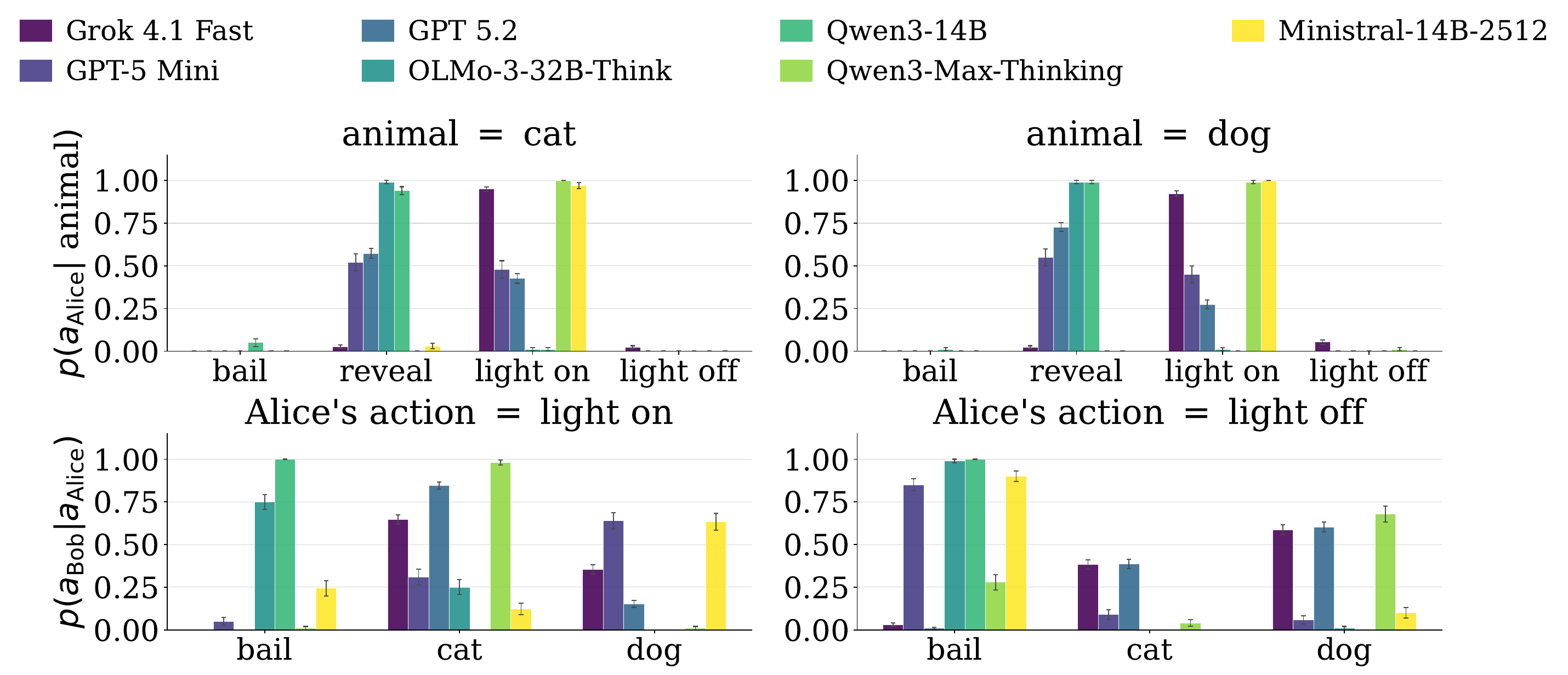}
    \caption{\textbf{Local policies induced by text agents on Cat \& Dog.} Most agents induce degenerate signalling policies; e.g, Qwen3-Max-Thinking and Ministral-14B-2512 turn the light on regardless of the observed animal. Qwen3-14B and OLMo-3-32B-Think induce the \acs{zsc}-optimal policy.
    }
    \label{fig:local_policies_cat_dog}
    \vspace{-3mm}
\end{figure*}
\label{sec:zsc_performance}

It is natural to ask whether \acs{llm} agents coordinate effectively out of the box, rendering further study unnecessary. Towards this, we conduct experiments across nine diverse coordination problems. Many of these are adapted from the \acs{zsc} literature, including Cat \& Dog~\citep{obl_hu}, Tiny Hanabi~\citep{baysian_action_decoder_foerster} and a variant of the lever game~\citep{obl_hu, lfc_treutlein}. Our choice of environments is motivated by two reasons: (a) each environment \textbf{isolates a core coordination capability} under \acs{zsc} evaluation in simplified settings, enabling a clean assessment of whether \acs{llm} agents exhibit the targeted behaviour; and (b) collectively, they span a diverse set of challenges that arise in real-world coordination. These include, for example, selecting a safe fallback in the absence of shared conventions, forming conventions at test time, and achieving coordinated symmetry breaking. A detailed description of the environments and the challenges they capture is provided in \appref{app:environments}. We emphasise that strong performance across these environments constitutes a \textbf{necessary but not sufficient condition} for effective real-world coordination. For instance, the narrow tunnel environment isolates the problem of tie-breaking, an issue that has been observed in deployed autonomous driving systems such as Waymo \cite{foerster_waymo_twitter_video}. If LLM agents are used to generate navigation policies for autonomous vehicles \cite{ishida2024langprop}, failures in this environment would likely translate to analogous failures in the real world.


To ensure a fair evaluation, we include an explanation of the \acs{xp} evaluation framework used, as well as provide explicit instructions to the agents to optimise their policies for high \acs{xp} scores. The prompt templates used for both types of agents are reported in \appref{app:prompt}. While we evaluate agents across all environments, we conduct a detailed analysis on the Cat and Dog toy environment~\citep{obl_hu} and a simplified variant of Hanabi \cite{hanabi}, a widely used coordination benchmark. As we show later, coordination failure modes observed in Cat and Dog persist in the more complex game of Hanabi as well. In Cat \& Dog, as shown in \autoref{fig:combined}, there exist two incompatible conventions (light on $\Rightarrow$ cat and light on $\Rightarrow$ dog), both of which achieve the maximum reward of 10. However, without any prior communication, the (\acs{zsc}-optimal) strategy is to remove the wall, yielding a reward of 5. For code agents, we employ SOTA reasoning models, namely Grok~4.1~Fast, GPT-5.2, Gemini~3~Pro, Claude Opus 4.5 and DeepSeek~V3.2. Due to cost reasons, we evaluate text agents in the \acs{sp} setting across all games using GPT-5.2 and Grok~4.1~Fast, and restrict the variable \acs{llm} setting to Cat \& Dog.

\paragraph{Text agents} 
\begin{figure*}
    \centering
    \includegraphics[width=0.9\textwidth]{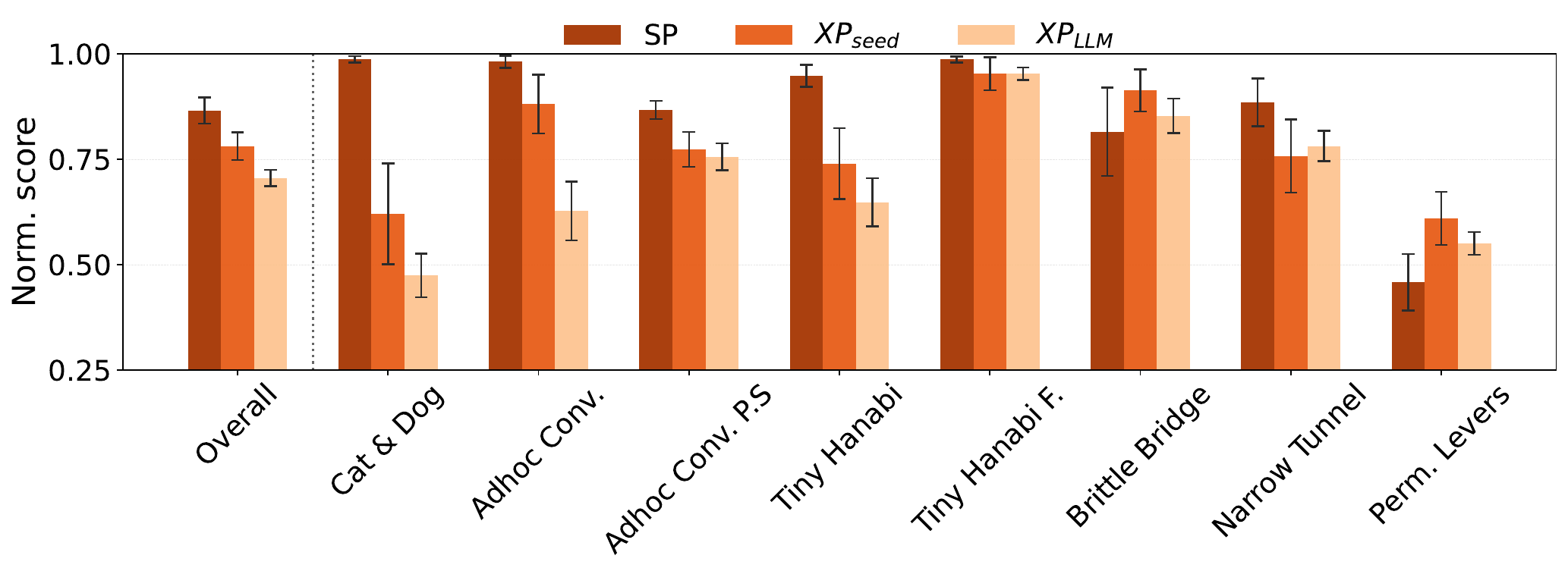}
\vspace{-0.25cm}
\caption{
\textbf{\acs{sp} and \acs{xp} scores of code agents, aggregated across \acsp{llm}.} Code agents exhibit a clear \sxpgap{}, indicating that independently generated policies use incompatible strategies.}
    \label{fig:code_agents_xp}
\vspace{-3mm}
\end{figure*}
As shown in \autoref{fig:combined}, text agents coordinate poorly in Cat and Dog, resulting in \acs{xp} scores well below the safe guaranteed reward of $5$ obtained by removing the wall. We find that Gemini~3~Pro consistently adopts the convention \textit{light on} $\Rightarrow$ \textit{cat}, enabling it to coordinate with itself. Since this convention is arbitrary, it leads to poor coordination with other agents. An example failure case is shown in \autoref{trace:cat_and_dog}, which records the reasoning traces from an episode between Gemini~3~Pro and GPT~5.2. When acting as Alice, the Gemini agent \textit{recognises} that any coordination convention is \textit{arbitrary}, yet nevertheless selects one to avoid the $-5$ penalty and subsequently rationalises this choice as “natural”. In contrast, when acting as Bob, the GPT agent independently rationalises the opposite convention to be “natural”; importantly, it does not consider bailing as an option.

Next, we estimate the empirical local policies of text agents by repeatedly prompting them $100$ times from a fixed game state, as illustrated in \autoref{fig:local_policies_cat_dog}. As Alice, we observe that Grok, Ministral, and Qwen3-Max consistently turn the light on regardless of the observed animal, yielding a degenerate signalling policy. GPT models exhibit similar behaviour, though with lower probability, opting to reveal in the remaining cases. We observe similarly degenerate behaviour as Bob. Ministral tends to bail when the light is off but guesses “dog” with high probability when the light is on, whereas Qwen3-Max guesses “cat” when the light is on and “dog” otherwise. Overall, their behaviour is largely arbitrary and degenerate. Surprisingly, smaller models such as Qwen3-14B and OLMo-32B-Think recover the optimal ZSC policy: revealing as Alice and bailing when observing a signal as Bob, despite being weaker than frontier models like Grok-4.1-Fast, GPT-5.2, and Qwen3-Max-Thinking at mainstream tasks such as Math and Coding. Notably, Qwen3-Max-Thinking ($\sim\!1$T parameters) performs significantly worse than its smaller counterpart, Qwen3-14B. We observe a similar pattern for smaller models such as Grok~4.1~Fast and GPT-5~mini (see \autoref{tab:reasoning_effort} in the Appendix), where increased reasoning effort correlates with lower coordination performance due to more frequent hallucinations of a shared convention. However, the relationship is mixed: for frontier models such as Claude Opus~5 and Grok~4.5, higher reasoning effort increases the reveal rate, whereas the newer GPT-5.6 Sol reveals far less often than GPT-5.2 (\appref{sec:app/reasoning_effort}). Greater model capability or reasoning effort therefore does not reliably translate into stronger \acs{zsc}.

Next, we examine how semantics, specifically, textual descriptions of the environment, affect the agents’ local policies. To this end, we consider two reformulations of the Cat \& Dog game. First, we rename the signalling actions from “light on/off” to “red/blue light.” Second, we translate the entire environment description $\mathcal{D}$ into German. The latter reflects settings where coordinating agents may access the same environment (e.g., a website) in different languages. Note that the payoffs in each of these cases is unchanged. We find that semantics can significantly influence the agents’ local policies as shown in \autoref{fig:local_policies_all} (Top-Right and Bottom-Left) in the Appendix. For instance, GPT models become more grounded, choosing to reveal with higher probability when we relabel the actions to red light and blue light. In contrast, Qwen3-14B and OLMo-3-32B exhibit a reduced tendency to reveal, with Qwen3-14B increasingly opting to bail as Alice-an effect that is further amplified under the German relabeling. Meanwhile, Grok, Ministral, and GPT-5-Mini continue to exhibit degenerate policies across these modified environments. These results indicate that the \textbf{models are unable to robustly reason about the game and are arbitrarily influenced by \textit{semantics}}.


As discussed in \autoref{sec:intro}, successful coordination requires theory of mind (\acs{tom})~\citep{theory_of_mind_premack}, since agents must reason about which conventions, if any, their partners are likely to adopt. In the absence of an established convention, a well-calibrated agent should maintain uncertainty over how its partner will interpret a signal~\citep{obl_hu}; selecting the ambiguous signal anyway reflects a failure of this reasoning. We note, however, that failures in Cat \& Dog do not by themselves causally separate weak \acs{tom} from a failure to recognise that no shared convention exists, and we view both as contributing factors. We observed that the agents are strongly averse to incurring the penalty for revealing, which pushes them to adopt an arbitrary signalling convention (also seen in \autoref{trace:cat_and_dog}). Moreover, agents rationalise their chosen convention as the unique "natural" convention, something that the Gemini agent exploits for successful coordination (\autoref{fig:combined}). We test whether this is a general strategy that other agents can also exploit. For this, we add the following instruction to the prompt: "You are playing a cooperative game with an identical instance of yourself. The same language model is used for both players, but each instance is assigned a different role in the game." If the model has a consistent notion of what it considers a "natural" convention, it should be able to coordinate on that convention and thereby achieve the maximum reward. That said, as shown in \autoref{fig:local_policies_all} (Bottom-Right) in the Appendix, agents are unable to exhibit such capabilities. Instead, coordination performance deteriorates. For example, GPT-5.2, which initially revealed in about $75\%$ of cases, now adopts a degenerate policy of always turning the light on, as Alice, revealing in only about $5\%$ of cases. Similarly, Qwen3-14B and OLMo-32B-Think reduce their reveal rates to around $50\%$ when observing a dog, down from over $95\%$. As Bob, they also shift toward guessing the animal more frequently, rather than bailing as they did previously. These results suggest that \textbf{agents behave as though a consistent internal notion of a “natural” convention exists; however, in practice, no such consistency is observed}. 
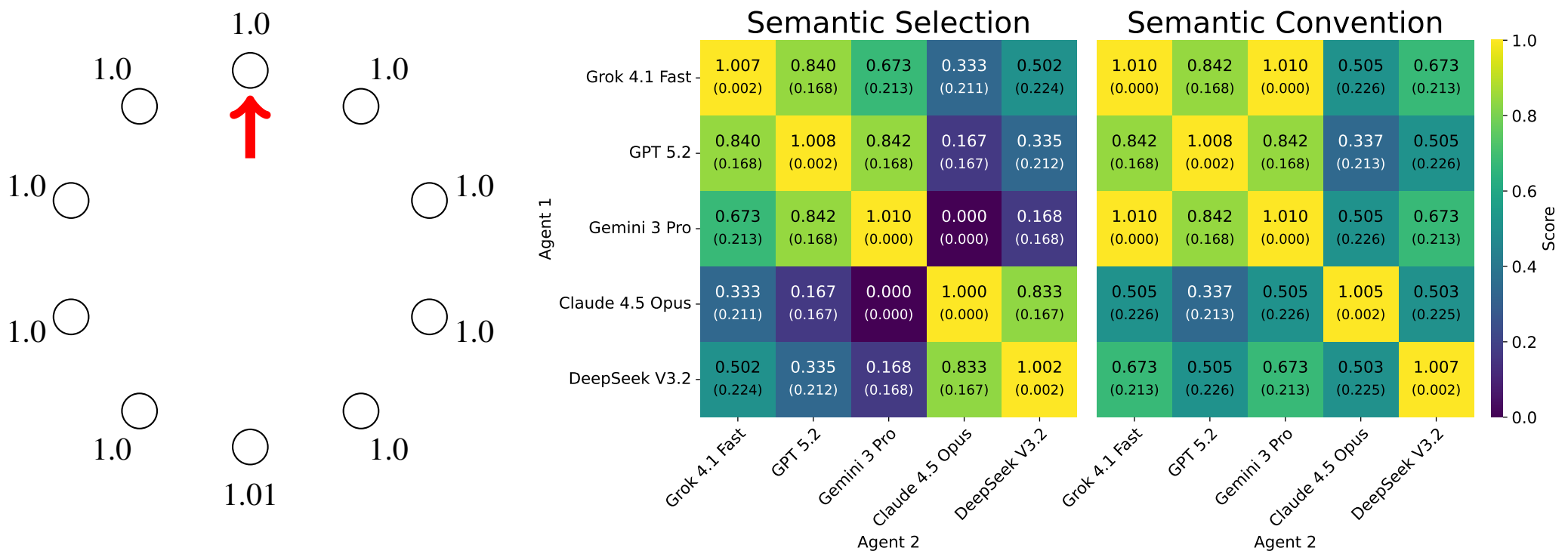
\begin{figure}
    \centering
    \begin{subfigure}[c]{0.3\textwidth}
        \centering
        \begin{tikzpicture}[scale=0.55, transform shape]

\def\R{3.0}      
\def\r{0.28}     
\def\L{0.75}     

\def\arrowW{2.5pt}   
\def\arrowStart{1.4} 
\def\arrowGap{0.12}  

\def\vals{{1.0,1.0,1.0,1.0,1.0,1.01,1.0,1.0,1.0,1.0}}

\foreach \k [evaluate=\k as \ang using 90-36*\k] in {0,...,9} {
  \draw (\ang:\R) circle (\r);

  \pgfmathtruncatemacro{\idx}{\k} 
  \node[font=\Large] at (\ang:\R+\L) {\pgfmathparse{\vals[\idx]}\pgfmathresult};
}

\draw[->, red, line width=\arrowW] (90:\R-\arrowStart) -- (90:\R-\r-\arrowGap);

\end{tikzpicture}
    \end{subfigure}
    \hspace{0.02\textwidth}
    \begin{subfigure}[c]{0.65\textwidth}
        \centering
        \includegraphics[width=\linewidth]{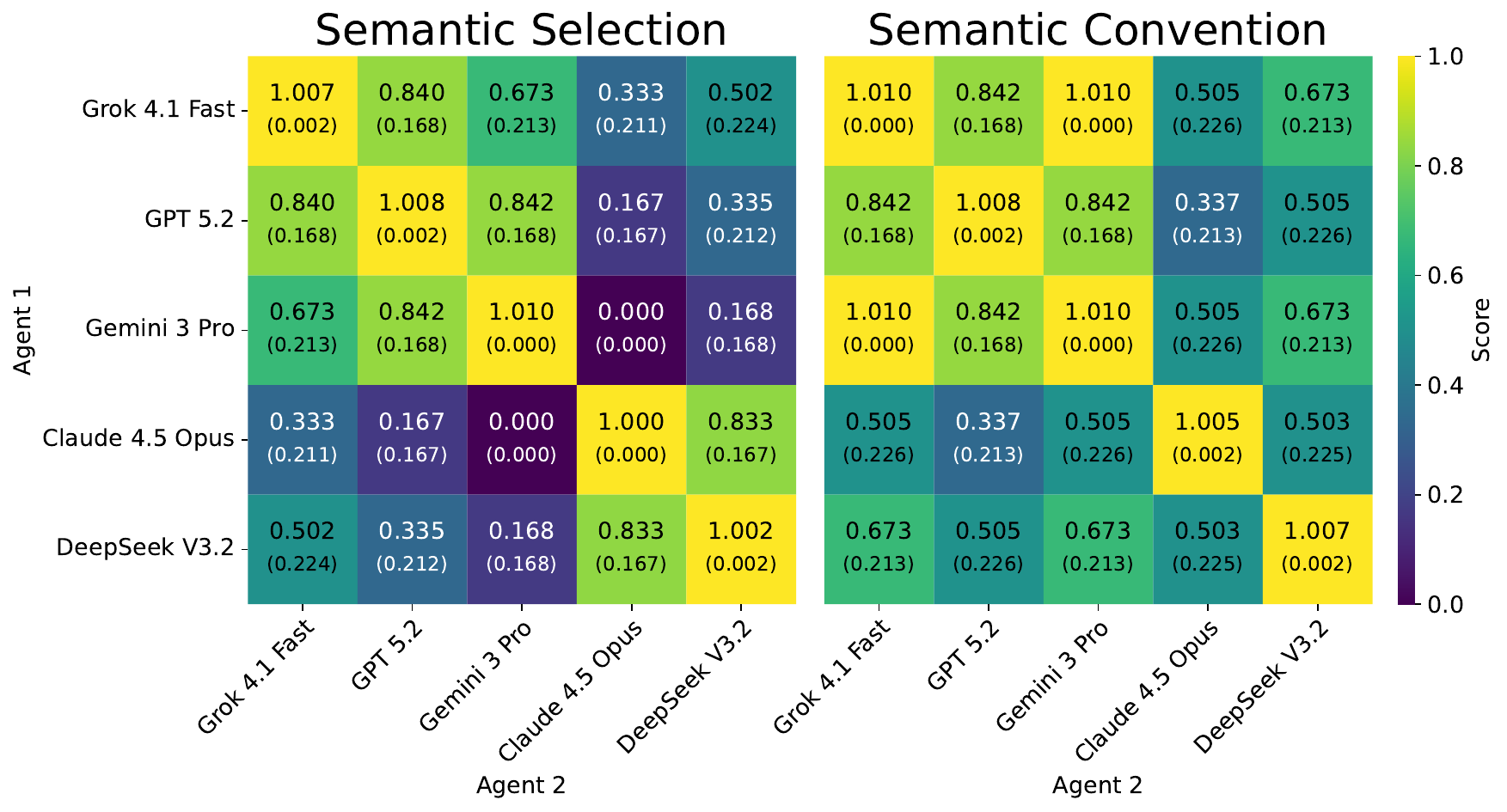}
    \end{subfigure}
    \vspace{-0.25cm}
    \caption{\textit{(Left)} \textbf{Semantic Selection game.} Agents could either coordinate on the semantic focal point (top) or Dec-POMDP-based focal point (bottom). \textit{(Right)} \textbf{\acs{xp} scores of code agents on both games.} The \sxpgap{} reveals that agents inconsistently choose between the two strategies.} 
    \label{fig:combined2_new}
    \vspace{-3mm}
\end{figure}

Finally, we find that coordination failures in text agents are not limited to the Cat \& Dog environment. As summarised in (\autoref{tab:text_xp_summary} in the Appendix), Grok~4.1~Fast and GPT~5.2 exhibit high failure rates in $7/9$ tasks. Since these simplified environments are designed to isolate coordination failure modes, we examine whether the same failure modes persist in more complex settings. This helps validate our claim that strong performance in simplified environments is \textbf{necessary but not sufficient} for real-world coordination. We evaluate text agents on a 3x3 variant of Hanabi \cite{hanabi}, of which Cat \& Dog serves as an abstraction. The game mechanics are described in \appref{app:environments} in the Appendix, and results are shown in \autoref{fig:hanabi_xp} in the Appendix. We find that most failures stem from mismatched assumptions of shared conventions, closely mirroring our findings in Cat \& Dog. For instance, a predominant failure mode arises when the clue receiver assumes the “Good Touch Principle” (i.e., that all hinted cards will be played), while the clue giver does not (\autoref{trace:hanabi_good_touch_failure} in the Appendix). 




\paragraph{Code agents} 

We evaluate the \acs{sp} and \acs{xp} performance of code agents both in the fixed \acs{llm} ($\text{SP}, \text{XP}_{\text{seed}}$) and variable \acs{llm} setting ($\text{XP}_{\text{LLM}}$) and report the aggregated results in \autoref{fig:code_agents_xp}, providing detailed results in \appref{sec:app/code_agents_with_cfd}. Similar to text agents, code agents although \textit{aware} of the safe reveal strategy are averse to the penalty associated with revealing. Adding to this we observe a strong bias toward adopting conventions based on \textit{semantics}, primarily via \textit{lexicographical ordering}. Yet, the agents frequently fail to agree on the ordering itself. For instance, assumptions such as cat$<$dog and on$<$off make mappings like cat $\rightarrow$ light on a common choice. However, many agents instead interpret light off/on as a binary signal, equating it with $0/1$, which yields light off$=$cat resulting in coordination failure. More broadly, lexicographical ordering is a brittle \acs{zsc} coordination rule. For example, if agents use textual descriptions from different languages (as discussed in \autoref{sec:zsc_performance}), the ordering may invert. For example, \textit{Hund} (dog) precedes \textit{Katze} (cat) in German.

Similar coordination failures are observed across the remaining environments. This stems from code agents generating joint policies that achieve high self-play returns but coordinate poorly with other independently generated policies. Across both fixed and variable \acs{llm} settings, we observe a significant gap relative to \acs{sp} returns (see \autoref{fig:code_agents_xp}). Qualitative analysis (\appref{sec:app/reasoning_traces}) further indicates that these policies are brittle and sheds light on the dominant failure modes across environments. Although the evaluation settings of code and text agents are not directly comparable, we qualitatively find that code agents are markedly more likely to resort to such brittle policies, while text agents exhibit more robust reasoning. Overall, our experiments demonstrate that \textbf{both sets of agents perform poorly across the evaluated environments}. This finding is particularly important in light of recent work deploying both text and code agents in real-world \acs{zsc} settings \cite{Liang2022CodeAP, szot2024large, ishida2024langprop}.

\section{Semantics and Coordination}
\label{sec:semantics_impact}
\definecolor{darkgold}{RGB}{212,175,55}
\begin{figure}
\centering
\begin{tcolorbox}[
  colback=green!8,
  colframe=green!60!black,
  rounded corners,
  arc=3mm,              
  boxrule=0.8pt,
  width=\linewidth,
  title={Hand-crafted CFD},
  fonttitle=\bfseries,
  left=1.5mm,
  right=1.5mm,
  top=1mm,
  bottom=1mm,
  before skip=6pt,
  after skip=6pt
]
\scriptsize\ttfamily
\noindent
An action/policy is \emph{not coordination-friendly} if it satisfies \emph{any} of the following:

\begin{itemize}
  \setlength\itemsep{0.2em}
  \setlength\parskip{0em}
  \item It assumes the existence of arbitrary conventions. To assess this, ask:
    \begin{itemize}
      \setlength\itemsep{0.15em}
      \setlength\parskip{0em}
      \item Is it using a signalling convention that could be replaced by a different, equivalent convention?
      \item Is there a role assignment that could be flipped? (This applies when two players are symmetric.)
    \end{itemize}
  \item \textit{Note:} Most actions/policies rely on some convention—judge whether that convention is arbitrary or essential.
\end{itemize}

\noindent
Otherwise, the action/policy is coordination-friendly.

\vspace{1mm}
\noindent
Before suggesting an action/policy, evaluate the criteria above; if any apply, the action should be avoided.

\end{tcolorbox}
\begin{tcolorbox}[
  colback=darkgold!10,
  colframe=darkgold!80!black,
  rounded corners,
  arc=3mm,              
  boxrule=0.8pt,
  width=\linewidth,
  title={Meta-learned CFD},
  fonttitle=\bfseries,
  left=1.5mm,
  right=1.5mm,
  top=1mm,
  bottom=1mm,
  before skip=6pt,
  after skip=6pt
]
\scriptsize\ttfamily
\noindent
Prefer actions that would still be reasonable under several plausible partner interpretations of the shared observation, and avoid actions that are good only if a partner infers one specific unstated rationale.
\end{tcolorbox}

\caption{\textbf{Hand-crafted and Meta-learned \acp{cfd}.}}
\label{fig:cfd_candidate}
\vspace{-3mm}
\end{figure}

\acs{llm} agents rely on textual descriptions of the \acs{dec_pomdp} to access the environment, thereby introducing \textit{semantics} as an additional axis for coordination. Semantics have important implications for coordination, enabling agents to leverage semantic relationships between observations and actions that have been difficult for traditional \acs{rl} methods to exploit \citep{ma2023learningintuitivepoliciesusing}. Moreover, semantics introduce new coordination strategies that compete with those grounded in the structure of the Dec-POMDP. As discussed in the previous section, \acs{llm} agents exhibit a strong bias toward leveraging semantic features for coordination; here, we investigate this more systematically, using environments specifically designed to isolate the role of semantics in coordination.


\paragraph{Exploiting semantics for coordination}
Code agents achieve near-perfect \acs{xp} returns on Tiny Hanabi Fixed, despite its close similarity to the Cat \& Dog setting, which features a safe fallback and two symmetric but incompatible Nash equilibria. Inspection of the reasoning traces reveals that agents independently converge on a unique convention induced by the \textit{lexicographic} mapping $(A \mapsto 1, C \mapsto 2)$ (see \autoref{trace:tiny_hanabi_fixed_convergence} in the Appendix). Traditional \acs{zsc} algorithms, which cannot exploit such semantic structure, instead default to the safe fallback and incur lower rewards~\cite{baysian_action_decoder_foerster, obl_hu}.

To further probe this ability, we evaluate text and code agents on the Hint–Guess game \citep{ma2023learningintuitivepoliciesusing} (\appref{app:environments}), a task where success requires exploiting semantic relationships between observations and actions and where agents can coordinate through human-like reasoning patterns such as mutual exclusivity \citep{implicature1} and implicature \citep{implicature2} (\autoref{fig:hint_guess_scenarios} in the Appendix). Text agents perform strongly in this environment (\autoref{fig:hanabi_xp} in the Appendix), matching the human baselines reported in \citep{ma2023learningintuitivepoliciesusing} and adopting the same coordination strategies as human players (\autoref{tab:hint_guess_implicatures} in the Appendix); however, they occasionally suffer coordination failures that are unlikely to arise in human play (\autoref{trace:hint_guess_grok_gpt_failure} in the Appendix). Code agents, in contrast, perform poorly (\autoref{fig:hanabi_xp} in the Appendix): rather than grounding coordination in semantic features, they fall back on arbitrary conventions such as lexicographic sorting or positional indexing (\autoref{trace:hint_guess_gpt_canonical_index} in the Appendix), and none of the resulting policies successfully leverage mutual exclusivity or implicature. This is consistent with our findings in \autoref{sec:zsc_performance}, where code agents perform worse than text agents across the range of coordination tasks.


\paragraph{Semantics introduce competing strategies} 
Conversely, code agents achieve lower \acs{xp} scores in the Tiny Hanabi environment, despite the presence of a unique optimal strategy $(A \mapsto 2, C \mapsto 1)$. Analysis of the reasoning traces reveals that semantics introduce an additional coordination strategy based on lexicographic ordering, as discussed above, albeit with a lower joint return. The resulting competition between these strategies degrades XP performance, whereas traditional \acs{zsc} algorithms, which do not exploit such semantic structure, are not affected \cite{baysian_action_decoder_foerster}.

To further investigate this competition, we introduce the \textit{Semantic Selection Game} (\autoref{fig:combined2_new}) and the \textit{Semantic Convention Game} (\appref{app:environments}). Broadly, agents must choose between two competing and incompatible coordination strategies: coordinating on the uniquely \acs{sp}-optimal joint policy, or coordinating based on the semantic relationship between the observations and actions. For example, as shown in \autoref{fig:combined2_new}, agents can either coordinate by selecting the lever with reward 1.01 or by choosing the lever marked with a “big red arrow”. \autoref{fig:combined2_new} demonstrates that the \acs{llm} agents fail to resolve this competition consistently, resulting in low \acs{xp} scores. We find that Gemini 3 Pro is the only model that reliably selects the \acs{sp}-optimal policy across all games. In contrast, the other \acs{llm} agents alternate between the two competing strategies. We find that Claude Opus 4.5 is particularly biased toward semantic alignment, selecting the semantic-aligned joint policy in all six runs of the semantic selection and in three out of six runs of the semantic convention game.

\paragraph{Avoiding local optima}
Direct reasoning about the environment allows \acs{llm} agents to identify optimal strategies that are difficult for traditional \acs{rl} methods to discover through Monte Carlo rollouts, which are often prone to local minima. As shown in \appref{sec:app/tiger_trampline_game}, \acs{llm} agents reliably discover \textit{self-explaining deviations} in the trampoline–tiger coordination challenge~\citep{hu2022selfexplainingdeviationscoordination}, while widely used MARL algorithms such as MAPPO~\citep{yu2022mappo} and QMIX~\citep{rashid2018qmix} fail to uncover these solutions.

\section{Coordination Friendly Definitions}
\label{sec:cfd}
\begin{figure}
    \centering
    \includegraphics[width=0.9\linewidth]{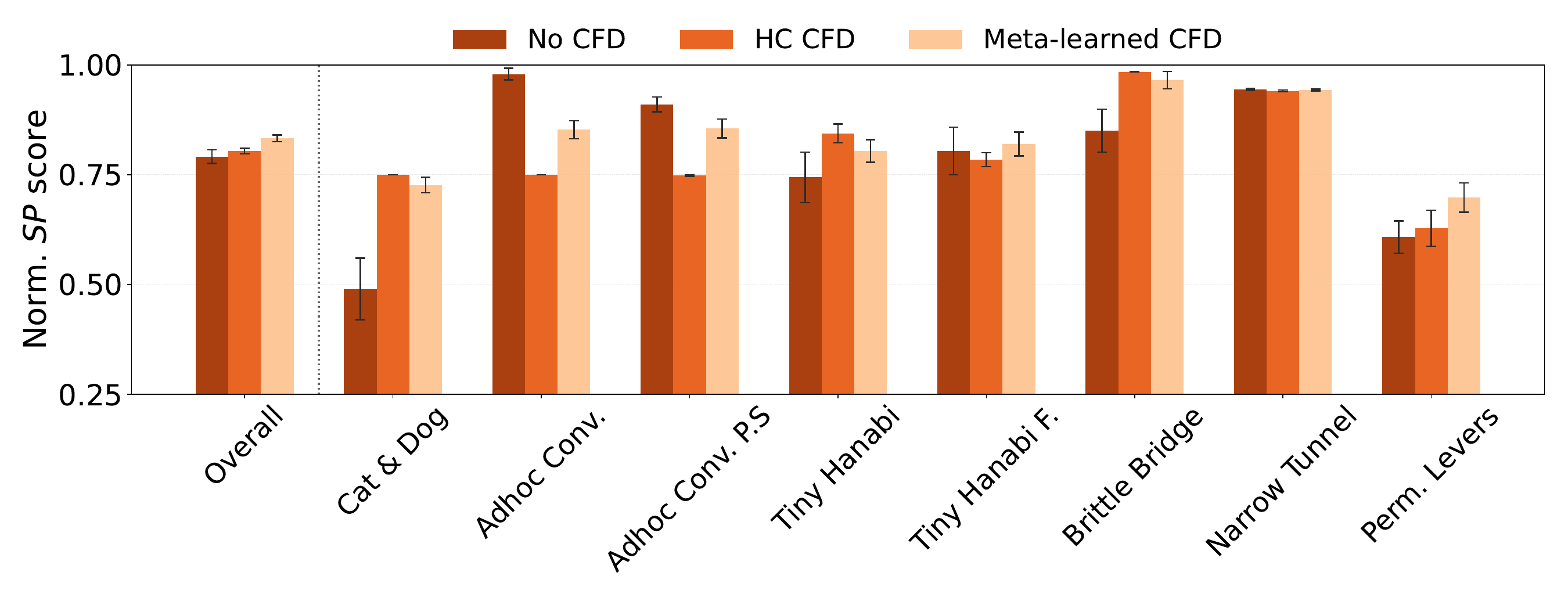}
    \caption{\textbf{Self-play performance of Grok~4.1~Fast as text agent.} Incorporating the meta-learned CFD outperforms the hand-crafted CFD, which in turn improves over the baseline.}
    \label{fig:text_sp}
    \vspace{-3mm}
\end{figure}

The results in \autoref{sec:zsc_performance} and \autoref{sec:semantics_impact} show that \acs{llm} agents fail even in simple settings, highlighting the need to adapt \acs{llm} agents for \acl{zsc}. Towards this, we propose a simple approach inspired by the core principle underlying \acs{zsc} algorithms. Prior work \citep{other_play_hu} motivates the idea of a ZSC learning rule as follows: “suppose that multiple independent AI designers will construct agents that have to interact in various but ex-ante unknown Dec-POMDPs without being able to coordinate beforehand, what learning rule should these designers agree on?” In the context of LLM agents, we introduce the analogous notion of a \acb{cfd}. A \acs{cfd} comprises a set of textual instructions that encode general coordination principles used to guide agent behaviour. We assume that independent developers can agree on a shared CFD to facilitate coordination among their agents. Given the diversity of scaffolds, including the text and code agents described in \autoref{sec:background}, we emphasise that CFDs should remain simple to ensure compatibility across implementations. 


The \acs{cfd} must influence the behaviour of the LLM agent, for example, by shaping the induced local policies of text agents or the joint policy generation process of code agents. In this work, we focus on the most straightforward application: incorporating the \acs{cfd} directly into the prompt. We also briefly explore a specialised scaffold for code agents that introduces a policy critic to modify or reject generated joint policies based on compliance with the \acs{cfd}. More advanced approaches, such as \acb{rlhf} \cite{stiennon2022learningsummarizehumanfeedback, bai2022traininghelpfulharmlessassistant} based on CFD compliance (similar to \cite{bai2022constitutionalaiharmlessnessai}), remain a promising direction for future work but are beyond the scope of this work.




\paragraph{Hand-crafted \acs{cfd}} We experiment with a simple hand-crafted \acs{cfd} as shown in \autoref{fig:cfd_candidate}, which is based on preventing arbitrary conventions and role assignments~\citep{other_play_hu, lfc_treutlein}. As shown in \autoref{fig:combined} and \autoref{tab:text_xp_summary}, \autoref{tab:code-xp-summary} and \autoref{tab:xp-summary-failure}, the \acs{cfd} improves the performance of both agents in the Cat \& Dog environment, where agents now consistently choose to remove the wall. We find that the \acs{cfd} has a stronger effect on text agents and smaller models, reducing the coordination failure rate in $6/9$ environments for Grok~4.1~Fast and $5/9$ for GPT-5.2 (\autoref{tab:text_xp_summary} in the Appendix). That said, the \acs{cfd} can be too restrictive in some cases. In the ad-hoc conventions environments, the \acs{cfd} biases the agents toward the safe fallback, preventing them from forming conventions at test time. 

The effect of the \acs{cfd} is less pronounced for code agents, where it fails to reduce the \sxpgap{} beyond the Cat \& Dog and the Semantic Selection Game environments. We attribute this to the generally weaker coordination abilities of code agents, as documented in \autoref{sec:zsc_performance} and~\autoref{sec:semantics_impact}. Further examination of execution traces revealed that, while code agents adjust their generated policies in response to the \acs{cfd}, these policies rarely account for edge cases, leading to coordination failures. Taken together, these results show that \textbf{conditioning \acs{llm} agents on general coordination principles via a \acs{cfd} is a simple and effective method for improving coordination.}

\paragraph{Automatic discovery of \acp{cfd}}
The hand-crafted \acs{cfd} is only an initial attempt and, as noted above, does not apply to all coordination challenges. Rather than refining it by hand, we explore whether \acp{cfd} can be discovered automatically via meta-learning (``learning to learn'') \cite{vanschoren2018metalearningsurvey, hospedales2020metalearningneuralnetworkssurvey, beck2025tutorialmetareinforcementlearning}, which has been shown to outperform human design choices.

To this end, we implement a meta-learning loop to autonomously discover CFDs using LLM-driven evolution. Our method is loosely based on GEPA~\cite{agrawal2026gepareflectivepromptevolution}, which employs Pareto-based candidate selection to improve sample efficiency in the evolutionary process. We combine this with other optimizations which increase the diversity of proposals in the evolutionary process and reduce the cost of ``meta-step''. We provide further implementation details in \appref{sec:app/implementation_details/meta_learning}. We meta-train the CFD on Cat \& Dog and Brittle Bridge using Grok-4.1-Fast as the code agent, $\text{XP}_{\text{seed}}$ as the measure of fitness and GPT 5.4 as the meta-optimisation agent. We additionally modify the code agent scaffold by introducing a policy critic that refines generated joint policies based on compliance with the \acs{cfd}, while leaving the code generation process itself unconditioned on the \acs{cfd} to preserve diversity. Empirically, this leads to more diverse solutions and reduced variance in the $\text{XP}_{\text{seed}}$ score of generated policies. Further implementation details regarding the policy critic are provided in \appref{sec:app/implementation_details/policy_critic}.

The discovered \acs{cfd} is roughly two-thirds shorter than the hand-crafted variant (\autoref{fig:cfd_candidate}) and generalises beyond its training conditions. Although meta-training only uses Grok~4.1~Fast as a code agent (direct meta-training with text agents would have raised token cost and runtime by two orders of magnitude), the discovered \acs{cfd} also improves Grok~4.1~Fast operating as a text agent (\autoref{fig:text_sp}). We further observe modest $\text{XP}_{\text{LLM}}$ gains across other models (GPT~5.2, Claude~4.5, Gemini~3~Pro) and the generic code-agent scaffold (\autoref{fig:code_xp_llm}), even though meta-training optimised $\text{XP}_{\text{seed}}$ in the critic-based scaffold; we do not, however, observe $\text{XP}_{\text{seed}}$ gains across LLM types (\autoref{fig:code_xp_seed_cfd}). Overall, \acsp{cfd} can be efficiently meta-learned with inexpensive scaffolds while generalising to more capable agents for which direct optimisation would be infeasible.

\section{Conclusion and Future Work}
\label{sec:conclusion}
We study the \acs{zsc} performance of \acs{llm} agents across a range of coordination tasks under the commonly used \textit{text} and \textit{code} scaffolds, and find that they consistently fail to coordinate effectively, even in simplified environments. Although the two scaffolds are not directly comparable, we qualitatively observe that code agents are more prone to brittle, arbitrary conventions than text agents, an observation worth noting given their increasing deployment in real-world settings \cite{Liang2022CodeAP, ji2026genswarm, ishida2024langprop}. We identify several factors that contribute to these failures: a limited understanding of the coordination problem, a failure to recognise that no shared convention has been established, and weak \textit{\Acb{tom}} reasoning about how a partner will interpret one's actions. Our experiments do not causally isolate these factors from one another; rather, they jointly characterise the observed failures. We additionally find that \acs{llm} agents are biased toward exploiting \textit{semantic representations} of the environment, which can both aid and impede coordination. Next, we introduce \textit{\Acb{cfd}}, a simple and general mechanism for improving coordination that can be readily integrated into existing agent scaffolds, and show that \acp{cfd} can be automatically discovered through meta-learning, reducing the need for manual design.

\paragraph{Limitations} Both the hand-crafted and meta-learned \acp{cfd} remain relatively simple and do not apply to every coordination challenge. For the meta-learned \acs{cfd}, we attribute this in part to the limited number of meta-training environments, in line with prior meta-learning work showing that broader task distributions yield more robust generalisation \cite{lpo, sota-lpo, discobench}; the hand-crafted \acs{cfd}, being designed manually, would similarly benefit from refinement on a wider range of coordination settings. In particular, \acp{cfd} can lead agents to \textit{under-coordinate}: in the ad-hoc conventions environments, agents repeatedly choose the safe fallback instead of forming a convention at test time. This failure mode is not unique to our approach. \Ac{obl}~\cite{obl_hu}, for instance, would behave similarly, since it reinterprets the interaction history as though the partner had followed a random policy and therefore discards any ad-hoc convention formed during the interaction. Avoiding under-coordination while retaining robustness remains an open problem in \acs{zsc}. Nevertheless, we view under-coordination as preferable to the prevailing failure mode of \textit{over-coordination}, in which agents rely on brittle, arbitrary conventions that lead to catastrophic coordination failures. Finally, due to cost, our variable-\acs{llm} evaluation of text agents is less extensive than that of code agents and concentrates on Cat \& Dog, Hint-Guess, and the 3$\times$3 Hanabi variant.

\paragraph{Future work} Alternative integration mechanisms beyond prompting, such as \acs{rlhf}-based \acs{cfd} compliance \cite{bai2022constitutionalaiharmlessnessai}, are discussed in \autoref{sec:cfd}. In addition, principles from traditional \acs{zsc} algorithms \cite{obl_hu, other_play_hu} may be adapted in conjunction with LLM-RL methods such as GRPO~\cite{guo2025deepseekr1} to improve coordination in LLM agents. We find a mixed relationship between reasoning capacity and \acs{zsc} performance (\appref{sec:app/reasoning_effort}): more capable models or higher reasoning effort do not reliably yield better coordination, and understanding when they do is an important open question. Beyond our minimal environments, \acs{llm} agents should be evaluated in more complex settings such as Overcooked~\cite{gessler_overcooked_v2}, where agents can adapt to their partner within an episode. Two related settings fall outside the standard \acs{zsc} framework but capture important real-world challenges: adaptation to a partner across repeated episodes, and coordination given a small number of demonstrations of successful coordination (e.g., autonomous vehicles learning from human drivers); our few-shot baseline (\appref{sec:app/few_shot}) is a first step in this direction. Finally, our analysis focuses on textual descriptions; extending it to Vision–Language Models (VLMs), which can leverage visual cues for coordination, is a natural direction for future work.

\begin{ack}
Shashank Reddy is partially supported by the Cooperative AI PhD Fellowship from the Cooperative AI Foundation. Bidipta Sarkar is supported by the Clarendon Fund Scholarship in partnership with a Department of Engineering Science Studentship for his Oxford DPhil. Andrei Lupu was partially funded by a Fonds de recherche du Québec doctoral training scholarship. Jakob N. Foerster is partially funded and Johannes Forkel is funded by the UKRI grant EP/Y028481/1 (originally selected for funding by the ERC). The authors thank Darius Muglich, Antonio León Villares, and Michael Beukman for their thoughtful feedback on the manuscript.
\end{ack}

\clearpage
\bibliographystyle{plain}  
\bibliography{neurips_2026}

\clearpage
\appendix
\renewcommand{\sectionautorefname}{Appendix}
\renewcommand{\subsectionautorefname}{Appendix}
\renewcommand{\subsubsectionautorefname}{Appendix}

\clearpage

\section*{Appendix}
\addcontentsline{toc}{section}{Appendix}

\startcontents[app]
\begingroup
\titlecontents{section}[1.5em]{\smallskip\bfseries}{\contentslabel{1.5em}}{}{\hfill\thecontentspage}[\smallskip]
\titlecontents{subsection}[3.0em]{}{\contentslabel{2.3em}}{}{\dotfill\thecontentspage}
\begin{tcolorbox}[colback=white, colframe=black!30, title={\bfseries Table of Contents}, fonttitle=\small, sharp corners, boxrule=0.6pt]
\printcontents[app]{}{1}[3]{}
\end{tcolorbox}
\endgroup

\section{Related Work}
\label{sec:relatedwork}

Prior work has evaluated \acs{llm} agents across a variety of multi-agent settings, including embodied interaction \citep{li-etal-2023-theory, zhang2024building, guo2024embodied}, \acb{tom} \citep{decrypto_lupu, le-etal-2019-revisiting, sclar2025explore}, and games \citep{zhang-etal-2024-agent, xu2024exploringlargelanguagemodels, Wang2024BattleAgentBenchAB}. Closest to our setting, work on fully cooperative tasks has largely focused on spatial reasoning environments such as Overcooked~\citep{hierachical_ai_agents, proagent_zhang} and mazes~\citep{davidson2025collaborationgap}, often relying on specialised scaffolds to achieve performance competitive with traditional \acs{rl} methods. However, in the \acb{zsc} setting, independently developed agents are unlikely to share identical model-scaffold pairings, an implicit assumption in much of prior work, with \citep{davidson2025collaborationgap, decrypto_lupu} being notable exceptions. Moreover, the choice of the scaffold can substantially influence performance~\citep{meireles2025the}, making it difficult to disentangle inherent model capabilities from scaffold-induced effects, a challenge that has been noted in other domains~\citep{gringras2026safetyscaffoldingevaluationconditions}.

In contrast, a key advantage of using simplified environments, as in this work, is that they make failure modes explicit and easier to diagnose. In complex benchmarks such as Hanabi \cite{hanabi}, the core challenge is to communicate private information indirectly through actions and signals, which requires a combination of abilities, including theory of mind, convention formation, and long-horizon planning based on past interactions that can span dozens of timesteps. As a result, diagnosing coordination failures is challenging, as they may stem from deficiencies in any of the required abilities. Even among human players, post-game analysis involves reconstructing beliefs at each timestep of the game to identify where coordination broke down; doing so with \acs{llm} agents is even more challenging, given the difficulty of interpreting the reasons behind their actions. \citep{llm_coordination_agashe} attempt to address this by querying models at fixed game states with limited success. More recently, \citep{ramesh2026sparks} benchmark \acs{llm} agents in Hanabi and report \acs{xp} scores between different \acsp{llm} that improve with model strength. Their strongest agents average around 15 points, but rely on scaffolding with programmatic deductions and do not assign a score of zero to games that end after all life tokens are lost, which is the convention in prior \acs{zsc} work~\citep{hanabi}. Even so, these agents remain well below experienced human players (20+) and state-of-the-art \acs{zsc} methods (24+)~\citep{forkel2025entropy}, consistent with our findings. More broadly, complex environments require capabilities beyond coordination, for instance, strong planning in Hanabi or adequate state coverage in Overcooked \cite{gessler_overcooked_v2}, making it difficult to explicitly test for coordination abilities.

\section{Broader Impact Statement}
\label{sec:app/impact_statement}

This work studies coordination among independently developed \acs{llm} agents in a zero-shot setting, characterising both their strengths and failure modes. Although insights into multi-agent coordination are dual-use in principle, our focus on fully cooperative simplified environments substantially limits the risk of misuse, and the proposed \acsp{cfd} serve as an alignment-oriented mechanism for safer, more transparent coordination among independently developed agents.

\section{Licenses for Existing Assets}
\label{sec:app/licenses}

\paragraph{Environments.} The figures used to illustrate the Cat \& Dog~\citep{obl_hu}, Tiny Hanabi~\citep{baysian_action_decoder_foerster}, Hint--Guess~\citep{ma2023learningintuitivepoliciesusing}, Trampoline--Tiger~\citep{hu2022selfexplainingdeviationscoordination} and Matrix Game~\citep{trajectory_diversity_lupu} environments are reproduced from the corresponding arXiv submissions, which are released under the CC BY 4.0 license. We reimplement each of these environments from scratch based on the descriptions in the original papers.

\paragraph{\acsp{llm} accessed via API.} Grok~4.1~Fast and Grok~4.5 (xAI); GPT-5.2, GPT-5~Mini, GPT-5.4, and GPT-5.6 Sol (OpenAI); Gemini~3~Pro and Gemini~3.1~Pro (Google); Claude Opus~4.5, Claude Opus~5, Claude Sonnet~4.6, and Claude Haiku~4.5 (Anthropic); DeepSeek~V3.2 (DeepSeek); Qwen3-Max-Thinking (Alibaba); and Ministral-14B-2512 (Mistral) are accessed through OpenRouter and used in accordance with OpenRouter's terms of service and the underlying providers' usage policies.

\paragraph{Open-weight \acsp{llm}.} The token-level probability experiments use Qwen3-14B, OLMo-3-32B-Think, and Ministral-14B locally, all released under the Apache~2.0 license by their respective authors.

\section{Environments}
\label{app:environments}
\subsection{Navigation games}

We design a set of navigation challenges intended to capture the kinds of coordination challenges that arise in robotics and autonomous driving. A real-world example, documented in \href{https://x.com/j_foerst/status/1811676036418937208?lang=en}{this
 video}~\citep{foerster_waymo_twitter_video}, shows two Waymo vehicles hesitating at a narrow passage, each attempting to yield to the other. While such cautious behaviour is highly effective when interacting with proactive human drivers, it results in deadlock when both autonomous agents act symmetrically. More generally, navigation games highlight a central coordination problem: two symmetric players must \enquote{break symmetry} by adopting distinct, complementary roles in order to progress.

\paragraph{Narrow Tunnel} formalises the coordination challenge illustrated in the above example. The environment consists of a tunnel, laid out left to right, with two agents positioned at opposite ends, tasked with exchanging positions in as few turns as possible. The setting is turn-based, with agents selecting actions simultaneously from the action set $\{\texttt{LEFT}, \texttt{RIGHT}, \texttt{STAY}\}$. The tunnel is composed of two types of tiles: \emph{wide} tiles, which can accommodate both agents simultaneously, and \emph{narrow} tiles, which can host at most a single agent.

Collision rules restrict movement into narrow tiles. Specifically, (i) if two agents attempt to swap positions within a single turn and at least one of them occupies a narrow tile, the swap is blocked, and no movement occurs; (ii) if both agents attempt to enter the same narrow tile simultaneously, neither move is executed. Given these restrictions, at narrow points in the tunnel, agents must assume asymmetric roles to successfully traverse it (leader and follower). To incentivise minimising the number of turns, agents incur a per-step joint penalty of $-1$ until both agents have successfully exchanged positions.

The environment is designed to support any custom layout via a simple hyperparameter at initialisation. The default layout, shown below, mimics the coordination challenge illustrated in the previously shown video:\footnote{$\texttt{start\_left}$ and $\texttt{start\_right}$ are considered wide tiles.}

\[
\left[ \texttt{start\_left}, \texttt{wide}, \texttt{narrow}, \texttt{wide}, \texttt{start\_right} \right]
\]

\paragraph{Brittle Bridge} is inspired by cooperative platformer video games in which successful progression requires players to adopt complementary roles, such as one agent acting while the other temporarily waits. Although the assignment of such leader–follower roles is arbitrary, it is essential for successful task completion. Two agents are initially located on the same side of a bridge of length $n$. Their joint objective is to traverse the bridge such that both agents reach the opposite side in the minimal number of turns. The environment is turn-based, with agents selecting actions simultaneously from the set $\{\texttt{forward}, \texttt{backward}, \texttt{stay}\}$. No collision constraints are imposed, so multiple agents may occupy the same tile at any given time. The coordination challenge stems from the bridge's fragility. If both agents remain on the bridge, regardless of their exact positions, for more than a fixed number of consecutive turns, given by the parameter \texttt{bridge\_break\_threshold} (typically set to 1), the bridge collapses. When this happens, the episode ends and both agents receive a large negative reward of $-1000$. Importantly, the timer associated with the fragility constraint resets whenever at least one agent steps off the bridge. Consequently, when \texttt{bridge\_break\_threshold} equals 1, one agent may safely cross the bridge in its entirety while the other agent alternates between entering the first bridge tile and retreating to land, thereby preventing collapse. For our experiments, we fix $n$ to 5 and the \texttt{bridge\_break\_threshold} to 1. To incentivise minimising the number of turns, agents incur a per-step joint penalty of $-1$ until both agents have crossed the bridge.

\subsection{Convention games}

Convention games typically assign players fixed asymmetric roles and proceed sequentially. The coordination challenge arises from partial observability: the first player has access to some information that must be conveyed to the second player, who then makes a guess. To achieve a high reward, the players must establish a convention that maps possible signals to possible guesses.

\begin{figure}
    \centering
    \includegraphics[width=0.6\textwidth]{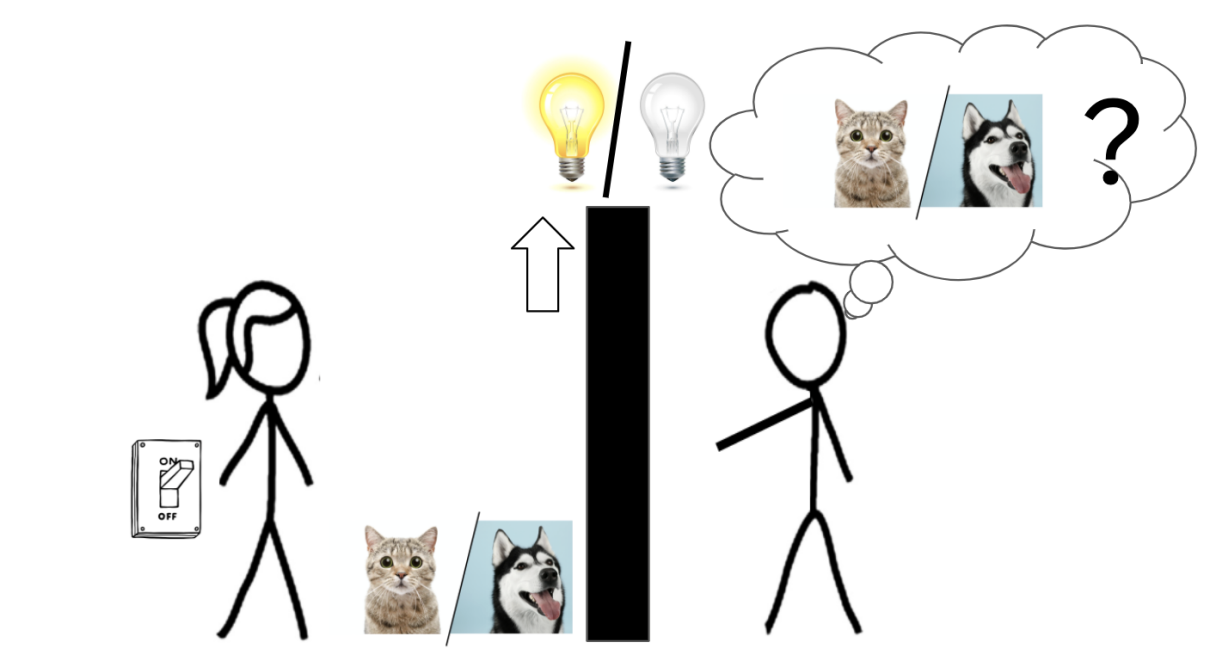}
    \caption{\textbf{An illustration of the Cat \& Dog game taken from~\cite{obl_hu}.}}
    \label{fig:app/envs/cat_and_dog}
\end{figure}

\paragraph{Cat \& Dog} is a two-player, fully cooperative coordination game introduced by~\citep{obl_hu}. The setup is straightforward: two players, \emph{Alice} and \emph{Bob}, are initially separated by a wall. At the beginning of the game, Alice privately observes a randomly chosen animal, either a cat or a dog, on her side of the wall. She must then select one of the following actions:

\begin{itemize}
    \item Turn on a light bulb on Bob’s side of the wall at no cost,
    \item Leave the light bulb off (the default; explicitly available as an action) at no cost,
    \item Remove the wall at a cost of $-5$,
    \item Bail, yielding a joint reward of $1$, terminating the game immediately.
\end{itemize}

Subsequently, it is Bob’s turn. He observes the state of the light bulb and, if Alice has removed the wall, also observes the animal. Bob then chooses one of the following actions:

\begin{itemize}
    \item Guess the correct animal (\emph{Cat} or \emph{Dog}). If his guess is correct, both players receive a joint reward of $10$; otherwise, they incur a joint penalty of $-10$. In expectation, a random guess yields a reward of $0$.
    \item Bail, yielding a smaller but guaranteed reward of $0.5$.
\end{itemize}

The term \emph{Cat \& Dog} refers specifically to the one-shot version of the game. The iterated version is examined in the following section. Since there are two equivalent but arbitrary signalling conventions (either \texttt{light\_on} $\rightarrow$ \emph{Cat} or \texttt{light\_on} $\rightarrow$ \emph{Dog}), independent agents without prior communication cannot coordinate on a convention within a single round. Therefore, removing the wall is considered ``optimal'' in the \acs{zsc} setting.

\paragraph{Worked example of $\mathcal{D}$.}
\label{app:worked_example}
\tcbset{
  workedexample/.style={
    colback=white,
    colframe=black!60,
    sharp corners,
    boxrule=0.8pt,
    fonttitle=\bfseries\small,
    colbacktitle=gray!15,
    coltitle=black,
    left=1.5mm, right=1.5mm, top=1mm, bottom=1mm,
    enhanced,
  },
}
\newcommand{\promptfield}[1]{\colorbox{blue!8}{\strut\texttt{\$\,#1\,\$}}}

\begin{figure}[t]
\centering
\begin{tcolorbox}[workedexample, width=\textwidth, title={Game Description $\mathcal{D}_{\mathcal{M}}$ — Cat \& Dog (shared by both players)}]
\scriptsize\setlength{\parindent}{0pt}\raggedright
You are playing a game of cat and dog with one other player. One of you is Alice and the other is Bob.
Alice is separated from Bob by a wall. The goal of the game is maximize your (joint) reward.
On Alice's side of the wall, there is either a cat or a dog.
Additionally, there is a light switch on Alice's side of the wall, which she can turn on to illuminate a bulb on Bob's side of the wall.

\smallskip
Alice goes first and can do exactly one of the following:\\
-- bail for a (joint) reward of 1 (this ends the game immediately)\\
-- remove the wall for a (joint) reward of $-5$\\
-- turn on the light bulb (no reward)\\
-- do nothing, i.e.\ the light bulb stays off (no reward)

\smallskip
If Alice didn't bail, it is Bob's turn. Bob observes the light status and the animal on Alice's side of the wall, if the wall was removed by Alice.
Bob can do exactly one of the following:\\
-- bail for a (joint) reward of 0.5\\
-- or guess the animal, i.e.\ either cat or dog\\
\hspace*{1em}-- if Bob guesses correctly, he receives a (joint) reward of 10\\
\hspace*{1em}-- if Bob guesses incorrectly, he receives a (joint) reward of $-10$

\smallskip
The players cannot communicate with each other except the options presented above (i.e.\ light status and removing the wall).
\end{tcolorbox}

\vspace{1mm}
\begin{minipage}[t]{0.49\linewidth}
\begin{tcolorbox}[workedexample, title={Turn 1 — Alice}, equal height group=catdogturns]
\scriptsize\setlength{\parindent}{0pt}\raggedright
\promptfield{Game History}\\[1pt]
You are Alice. {\color{gray}($\mathcal{D}_{\mathcal{R}}$)} On your side of the wall, there is a cat. {\color{gray}($\mathcal{D}_{\mathcal{H}}$)}

\smallskip
\promptfield{Legal Actions}\\[1pt]
\texttt{bail}, \texttt{remove\_wall}, \texttt{turn\_on\_light}, \texttt{do\_nothing}

\smallskip
\textbf{Response:}\\[1pt]
\texttt{REASONING: \ldots}\\[1pt]
\texttt{ACTION JSON: \{"action": "remove\_wall"\}}

\smallskip
\textbf{Outcome:} The wall is removed. Immediate joint reward: $-5$.
\end{tcolorbox}
\end{minipage}\hfill
\begin{minipage}[t]{0.49\linewidth}
\begin{tcolorbox}[workedexample, title={Turn 2 — Bob}, equal height group=catdogturns]
\scriptsize\setlength{\parindent}{0pt}\raggedright
\promptfield{Game History}\\[1pt]
You are Bob. {\color{gray}($\mathcal{D}_{\mathcal{R}}$)} The light bulb on your side of the wall is off. Furthermore, the wall has been removed. You can see a cat on the other side of the wall. {\color{gray}($\mathcal{D}_{\mathcal{H}}$)}

\smallskip
\promptfield{Legal Actions}\\[1pt]
\texttt{bail}, \texttt{guess\_dog}, \texttt{guess\_cat}

\smallskip
\textbf{Response:}\\[1pt]
\texttt{REASONING: \ldots}\\[1pt]
\texttt{ACTION JSON: \{"action": "guess\_cat"\}}

\smallskip
\textbf{Outcome:} Bob's guess is correct. Immediate joint reward: 10.
\end{tcolorbox}
\end{minipage}
\caption{\textbf{Worked example of the natural language descriptions $\mathcal{D} = (\mathcal{D}_{\mathcal{M}}, \mathcal{D}_{\mathcal{R}}, \{\mathcal{D}_{\mathcal{H}_i}\})$ for Cat \& Dog with text agents.} Both players receive the same game description $\mathcal{D}_{\mathcal{M}}$ (top). Each text agent additionally receives its role $\mathcal{D}_{\mathcal{R}}(r)$ and a description of its action-observation history $\mathcal{D}_{\mathcal{H}_i}(\tau_i)$, which together fill the \texttt{Game History} field of the text-agent prompt template (\autoref{fig:prompt_text}). The game description is taken verbatim from our implementation. In this episode, Alice plays the \acs{zsc}-optimal action of removing the wall, and the players obtain a total joint reward of $-5 + 10 = 5$. Code agents receive only $\mathcal{D}_{\mathcal{M}}$ and generate a joint policy for both roles.}
\label{fig:cat_and_dog_worked_example}
\end{figure}

\autoref{fig:cat_and_dog_worked_example} illustrates the natural language descriptions introduced in \autoref{sec:background} for a single episode of Cat \& Dog. The game description $\mathcal{D}_{\mathcal{M}}$ is shared by both players and specifies the roles, actions, and rewards. A text agent is additionally told which role it plays via $\mathcal{D}_{\mathcal{R}}$ and what it has observed so far via $\mathcal{D}_{\mathcal{H}_i}$; for example, Bob's history description reports the status of the light and, since Alice removed the wall, the animal. A code agent only receives $\mathcal{D}_{\mathcal{M}}$, since it generates a joint policy that specifies the behaviour of both Alice and Bob.

\paragraph{Ad-Hoc Conventions.} Inspired by \cite{gessler_overcooked_v2}, we introduce games that require protocol formation at test time. \emph{Ad-Hoc Conventions} is an iterated version of Cat \& Dog. In contrast to the one-shot version discussed in the previous section, iteration enables the formation of conventions that can lead to guaranteed high payoffs in later rounds. We replace the animals with generic levers that can be assigned labels (in our case $\{\texttt{blue}, \texttt{red}\}$) and the lightbulb is replaced by labelled signals (in our case $\{\texttt{square}, \texttt{circle}\}$).
Removing the wall is now represented by revealing the current correct lever. While we do not make further use of it in our experiments, we support an arbitrary number of signals and levers to increase the difficulty of the game. Additionally, the lever labels, signal labels, and the number of rounds can be specified as a hyperparameter. In our experiments, we play ten rounds per Ad-Hoc Conventions game.

\paragraph{Ad-Hoc Conventions with permuted signals.} We also introduce a harder variant of Ad-Hoc Conventions, called \emph{Ad-Hoc Conventions with permuted signals} (A.C.P.S.). Here we distinguish between \emph{buttons} and \emph{signals}, each with its own set of labels (${\texttt{apple}, \texttt{banana}}$ and ${\texttt{square}, \texttt{circle}}$ in our case). The only change per round is that, instead of directly revealing a signal, Alice can now press one of the buttons (or choose reveal or bail), which then displays a signal to Bob. The connection between buttons and signals is given by a random bijection unknown to both players. Importantly, Alice does not see the resulting signal, and Bob does not see the pressed button. Without prior communication, it is therefore impossible to recover the exact bijection between signals and levers. Note that the button to signal mapping is fixed within the episode.
As a result, the game cannot be solved through a simple lexicographical order strategy: Alice cannot know which button will trigger which signal for Bob. Nevertheless, in both the permuted and non-permuted versions, the players retain a simple fallback strategy, continuously revealing (\ie removing the wall), which guarantees a reward of 5 per round, for a total \enquote{safe} return of 50 across the 10 rounds we typically consider in our experiments.

\begin{figure}
    \centering
    \includegraphics[width=0.8\textwidth]{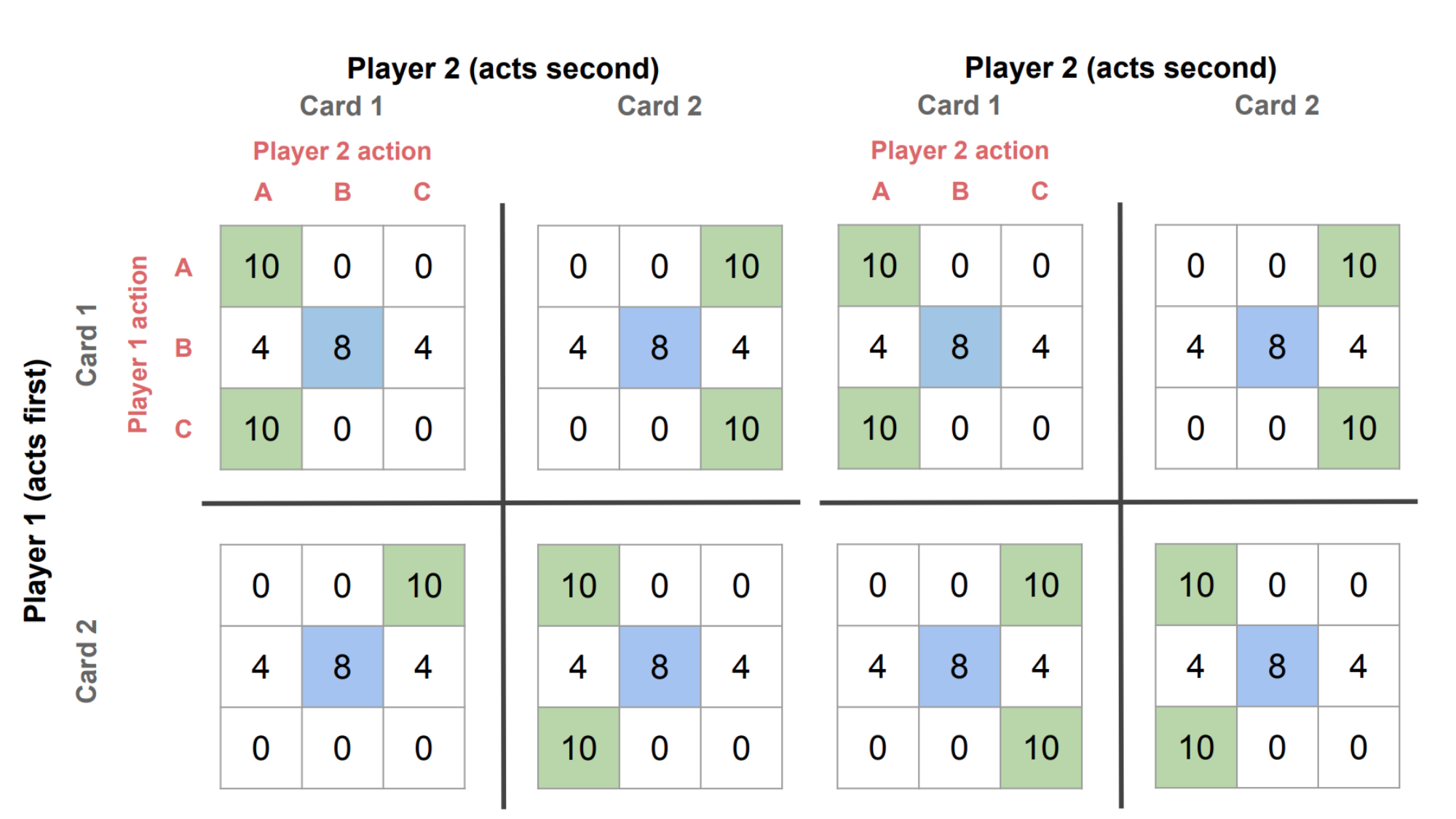}
    \caption{\textbf{Payoff matrices for Tiny Hanabi adapted from \cite{baysian_action_decoder_foerster}.} On the left side, we show the matrix from the original paper, and on the right side, the matrix of our version with a coordination challenge.}
    \label{fig:app/envs/tiny_hanabi}
\end{figure}

\paragraph{Tiny Hanabi} is a toy-game introduced in~\citep{baysian_action_decoder_foerster}. It consists of two players, who are each dealt a random card at the start of the game. The final joint reward depends on both the players’ cards and their actions throughout the game. The second player observes the first player’s actions before picking his action, which allows the first player’s action to be interpreted as a signal. We show the payoff matrix of Tiny Hanabi in \autoref{fig:app/envs/tiny_hanabi}. Due to the asymmetry of the payoff matrix, the original version of Tiny Hanabi has a unique \acs{sp} optimal policy, yielding a joint reward of 10. This policy is given by the first player picking action $A$ if he has card 2 and $C$ otherwise, with the second player responding accordingly. To see why it is unique, note that a joint reward of 10 requires that neither player ever plays $B$, and that the second player conditions on the first player's action, since no column of the payoff matrices consists entirely of 10s. The first player must therefore play different actions for the two cards. If the first player instead plays $C$ with card 2 while the second player holds card 1, the joint reward is 0 regardless of the second player's response. Hence, the first player must map card 2 to $A$ and card 1 to $C$. In Tiny Hanabi Fixed, the second player could respond to this case with $C$ and still obtain a reward of 10, which gives rise to two symmetric \acs{sp}-optimal policies.

To create a version of Tiny Hanabi with a coordination challenge, we symmetrize the payoff matrix, also shown in \autoref{fig:app/envs/tiny_hanabi}. As in the previous convention games, there is a safe fallback option, which does not require any conventions: Playing $B$ as the first player leads to a reward of $8$ as long as the second player responds rationally. Hence Tiny Hanabi Fixed is similar to Cat \& Dog wherein there exists two symmetrical conventions with identical payoffs and a safe fallback option.

\paragraph{Hanabi}
In the standard game of Hanabi, each colour has a deck of 10 cards: three 1s, two 2s, two 3s, two 4s, and one 5, for a total of 50 cards across 5 colours. The game is fully cooperative and features 5 ranks and 5 colours, with each card defined by a rank–colour pair. Players share a limited pool of 3 life tokens and 8 hint tokens. At the start, each player is dealt 5 cards; crucially, players cannot see their own hands but can observe those of their teammates. The objective is to build “fireworks” by stacking cards of each colour in increasing rank order. On each turn, a player may either play a card, discard a card, or provide a hint to another player. Hints can specify either a rank or a colour, indicating all cards of that type in the recipient’s hand, and each hint consumes one token. An incorrect play—i.e., one that cannot be placed on the fireworks—costs a life token. Discarding a card restores a hint token, up to the maximum. After playing or discarding, the player draws a new card from the deck, maintaining a constant hand size while cards remain. Players receive +1 for each correctly played card, and the game ends with a score of 0 if all life tokens are lost.

In this work, we consider a simplified 3×3 variant with 3 colours and 3 ranks, where each colour has three 1s, two 2s, and one 3, for a total maximum score of 9. The game includes a single life token and a limit of 3 hint tokens. The central challenge of Hanabi arises from the limited and costly communication channel provided by hints. Effective play therefore requires agents to use hints efficiently, often relying on implicit conventions to convey information about hidden states to their partners.

\subsection{Simultaneous selection games}

In \emph{simultaneous selection games}, symmetric players have to coordinate on one of many similar, often symmetric, choices to achieve a high joint reward. We study both one-shot versions of these games (\emph{Matrix Game}) and iterated versions (\emph{Permuted Levers}).

\paragraph{Permuted levers} Lever Games are common in \acs{zsc} literature~\citep{other_play_hu, lfc_treutlein}. For our purposes, we focus on iterated versions of lever games with $n$ levers all with identical payoff. In each round, both players simultaneously chose one of the levers. If players manage to pick the same lever, they receive a joint reward of 1. To prevent arbitrary focal points introduced by numbering the levers (which is needed for the \acs{llm} agents to interact with the environment), we introduce a further complication inspired by the \acl{op} objective~\citep{obl_hu}: Each player observes and acts on the levers through a random permutation of the lever numbers, which stays fixed across the rounds of the game. While players are aware of the fact, they do not know their own or others’ permutations.

Both the number of levers $k$ and the number of rounds $n$ are hyperparameters, which can be set at initialisation. For our experiments, we fix $k=5$ and $n=10$.

\begin{figure}
    \centering
    \includegraphics[width=0.3\textwidth]{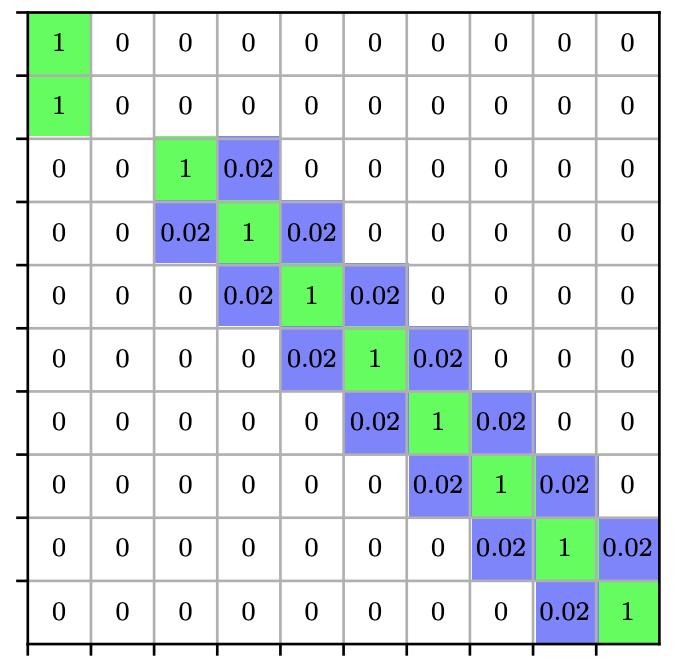}
    \caption{\textbf{Payoff matrices for the Matrix Game} adapted from \cite{trajectory_diversity_lupu}. 
    Players simultaneously pick a row and column, respectively, and receive the reward at the selected cell.}
    \label{fig:app/envs/matrix_game}
\end{figure}

\paragraph{Matrix game} Our version of the \emph{Matrix Game} was introduced by \cite{trajectory_diversity_lupu}. In this game, two players simultaneously choose a row/column and receive the reward in the selected cell. The matrix used to play this game is shown in \autoref{fig:app/envs/matrix_game}. For humans, there is an easy-to-identify focal point given by the first column and first/second row, which can be exploited. Our version of the matrix game uses a $11 \times 11$ matrix extending the diagonal pattern by 1. This yields another possible focal point at the middle of the diagonal pattern (see also \autoref{trace:matrix_game_failure}). Matrix game differs from the other coordination challenges presented in this section in that there is no salient policy that could be considered ``\acs{zsc}-optimal''. Successful coordination in the matrix game largely depends on players forming the same arbitrary conventions to form a focal point in the matrix, exactly the inverse of what \acs{zsc} is trying to achieve. Therefore, we have chosen to exclude Matrix game from the main experiments of the paper, but report results for completeness in \autoref{sec:app/text_agents_with_cfd} and \autoref{sec:app/code_agents_with_cfd}.

\subsection{Semantic games}
\begin{figure}[t]
\centering
\begin{tikzpicture}[
  cellsize/.store in=\cellsize, cellsize=0.95cm,
  fontsize/.store in=\fs, fontsize=\large,
  x=\cellsize, y=\cellsize,
  scale=1.15, transform shape]
]
  \definecolor{mygreen}{RGB}{90,255,90}
  \definecolor{myblue}{RGB}{120,130,255}

  \def\rows{4}
  \def\cols{4}

  \foreach \i in {1,...,\rows} {
    \pgfmathtruncatemacro{\y}{\rows-\i} 
    \foreach \j in {1,...,\cols} {

      \edef\v{\payoff{\i}{\j}}

      \def\fillcol{white}
      \pgfmathparse{abs(\v-1.0)<0.0005 ? 1 : 0}
      \ifnum\pgfmathresult=1 \def\fillcol{mygreen}\fi
      \pgfmathparse{abs(\v-1.01)<0.0005 ? 1 : 0}
      \ifnum\pgfmathresult=1 \def\fillcol{myblue}\fi

      \path[fill=\fillcol, draw=black!25, line width=0.6pt]
        (\j-1,\y) rectangle (\j,\y+1);

      \node[font=\fs] at (\j-0.5,\y+0.5) {\pgfmathprintnumber[fixed,precision=2]{\v}};
    }
  }

  \draw[line width=1.2pt] (0,0) rectangle (\cols,\rows);

  \foreach \i in {1,...,\rows} {
    \pgfmathtruncatemacro{\y}{\rows-\i}
    \pgfmathtruncatemacro{\idx}{\i-1}
    \edef\rlabel{\ifcase\idx A\or B\or C\or D\fi}
    \node[font=\fs] at (-0.45,\y+0.5) {\rlabel};
  }

  \foreach \j in {1,...,\cols} {
    \node[font=\fs] at (\j-0.5,\rows+0.45) {\j};
  }

\end{tikzpicture}
\caption{\textbf{Semantic Convention Game.} Alice and Bob chose a row and column in the matrix sequentially. Additionally, only Alice is informed of the currently active column, which is chosen at random. Only if the agents end up selecting a cell in the active column do they receive the reward in that cell. Agents must choose between the lexicographical mapping convention (green) and the \acs{sp}-optimal mapping convention (blue).}
\label{fig:semantic-convention-game}
\end{figure}
\begin{figure*}
    \centering
    \begin{minipage}[t]{0.49\linewidth}
        \centering
        \includegraphics[width=\linewidth]{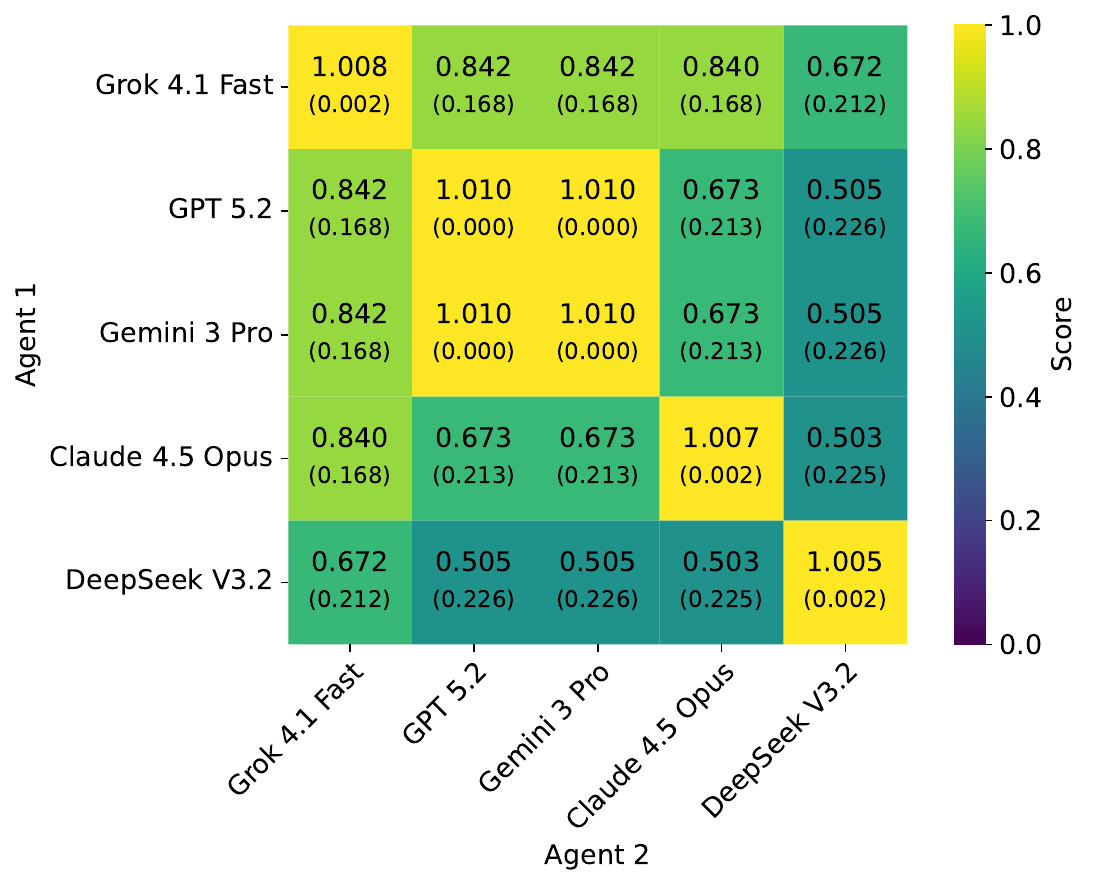}
        \phantomsubcaption\label{fig:semantic/convention_w_cfd_xp}
    \end{minipage}
    \hfill
    \begin{minipage}[t]{0.49\linewidth}
        \centering
        \includegraphics[width=\linewidth]{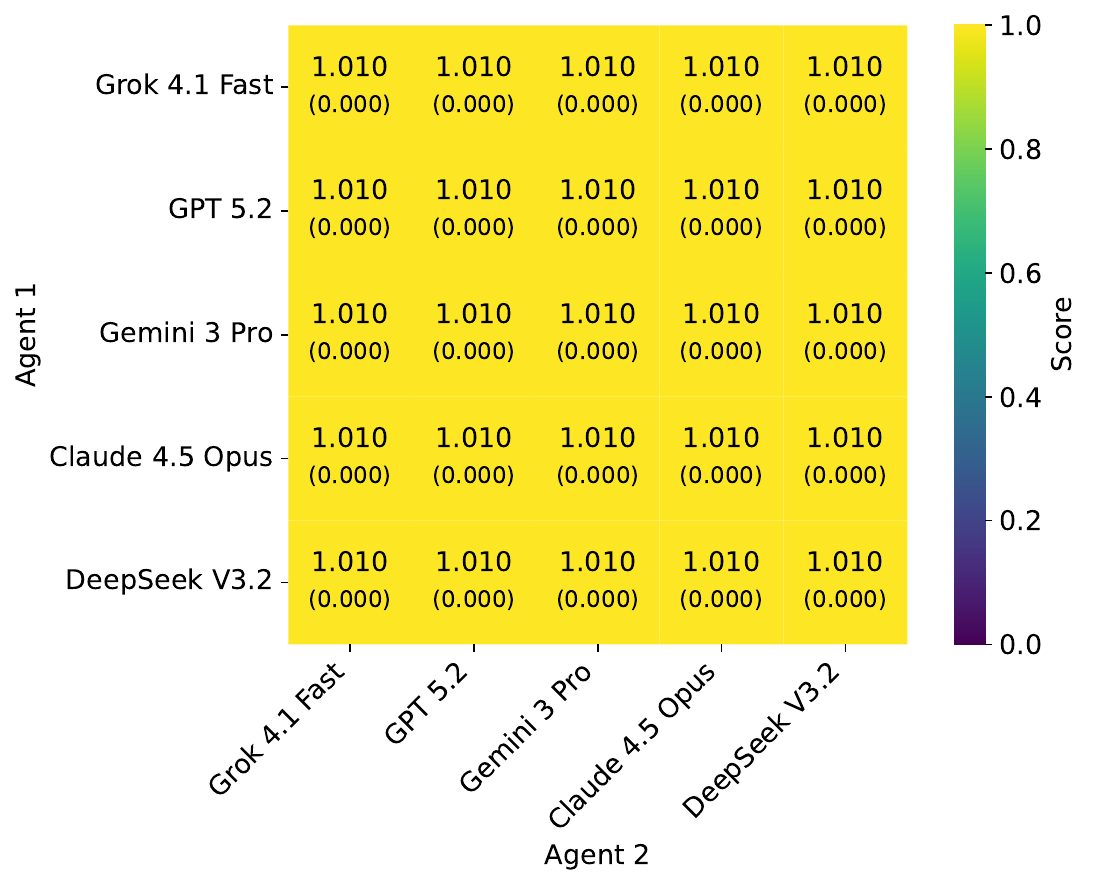}
        \phantomsubcaption\label{fig:semantic/selection_w_cfd_xp}
    \end{minipage}
    \caption{\textbf{\acs{xp} scores of code agents in the semantic games using the hand-crafted \acs{cfd}.} (Left) Semantic convention game. (Right) Semantic selection game. Scores are averaged across six i.i.d.\ \acs{xp} matrices; where each cell contained the mean return over 100 episodes.}
    \label{fig:semantic/cfd_xp_combined}
\end{figure*}

\paragraph{Semantic selection and convention games} demonstrate that the addition of semantics can create coordination challenges in games that previously did not exhibit them, as they have a unique \acs{sp} optimal policy. In particular, the semantic games presented in this paper, the \textit{semantic selection game} (\autoref{fig:combined2_new}) and the \textit{semantic convention game} (\autoref{fig:semantic-convention-game}), force players to choose between two mutually incompatible meta-strategies: either playing the \acs{sp} optimal strategy or following a strategy that is aligned with the game semantics. \autoref{sec:zsc_performance} has demonstrated that \acs{llm} agents often struggle to consistently interpret semantics and fail to properly reason about these difficulties. Semantic games, on the other hand, show that even when the semantic interpretation is clear across agents, semantics can still hinder the coordination abilities of agents, just by the fact that they introduce another possible meta-strategy.

\autoref{fig:combined2_new} demonstrates that \acs{llm} agents are generally not able to consistently choose among these competing meta-strategies, leading to coordination failures. Interestingly, even though the \acs{cfd} does not specify how to decide among the competing meta-strategies, it does lead agents to consistently opt for the \acs{sp} optimal policy in the semantic selection game. In the semantic convention game, the \acs{cfd} does not yield significant improvement in the agents' coordination as shown in \autoref{fig:semantic/cfd_xp_combined} (Left).

\begin{figure}[h]
    \centering
    \includegraphics[width=0.8\linewidth]{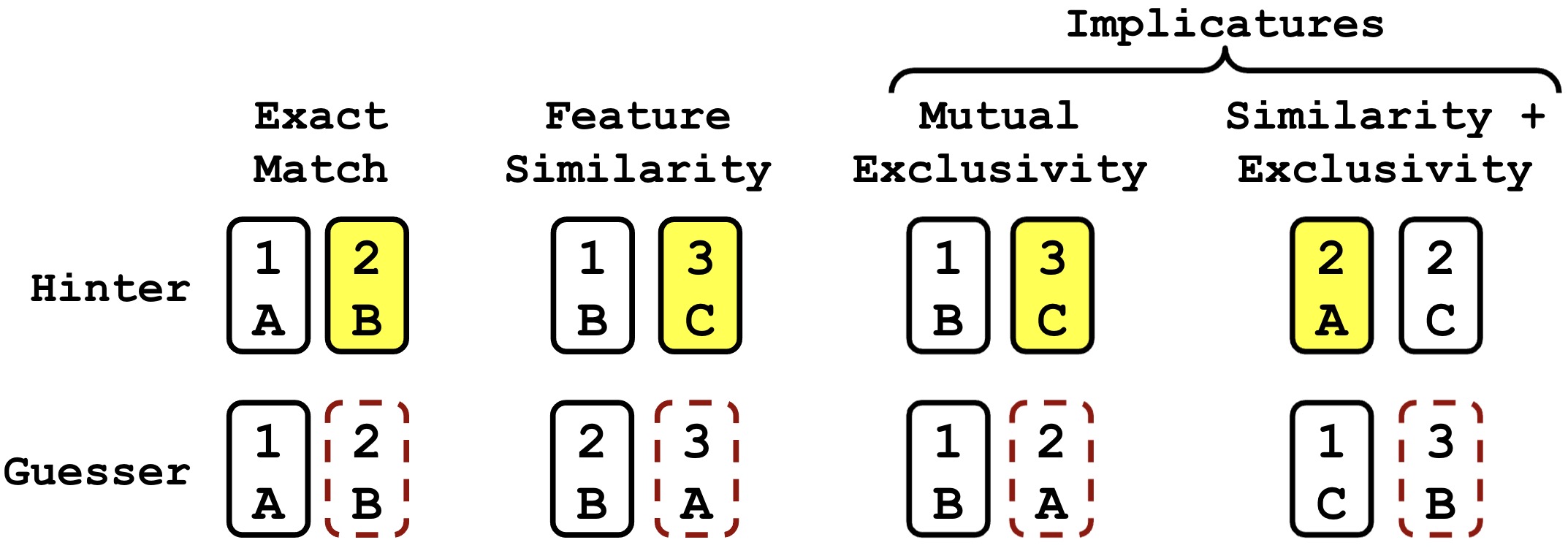}
    \caption{\textbf{Example hint-guess scenarios, adapted from Figure~2 of~\cite{ma2023learningintuitivepoliciesusing}.} Each panel shows a distinct coordination challenge: exact feature match, partial feature similarity, mutual exclusivity, and a combination of the latter two, which represents implicatures~\cite{implicature1, implicature2}. The card outlined in red is the target; the card highlighted in yellow is the human-compatible hint choice. The two rightmost scenarios require reasoning about implicature, where the intuitive hint shares no direct feature with the target.}
    \label{fig:hint_guess_scenarios}
\end{figure}

\paragraph{Hint-Guess}
The \textit{hint-guess} game was introduced in \cite{ma2023learningintuitivepoliciesusing} to study whether agents can exploit shared semantic structure between actions and observations for zero-shot coordination. The game involves two players, a \emph{hinter} and a \emph{guesser}, each dealt a hand of $N$ cards, where every card is described by two discrete features. One card from the guesser's hand is secretly designated the target, and its identity is revealed only to the hinter. The hinter must select a card from their \emph{own} hand to show the guesser; the guesser then picks which card in their hand corresponds to the target. Both players receive a reward of $1$ if the guess is correct and $0$ otherwise. The environment is visualised in \autoref{fig:hint_guess_scenarios}. Interestingly, implicatures~\cite{implicature1, implicature2} represent sophisticated patterns of human reasoning, making them a compelling test of whether LLM agents can exhibit similar capabilities. As shown in \autoref{tab:hint_guess_implicatures}, text agents consistently exhibit human-like reasoning and, notably, demonstrate the ability to coordinate via implicatures.

\section{Schematic diagram of \acs{llm} agents}
\begin{figure*}
    \centering
    \includegraphics[width=0.8\textwidth]{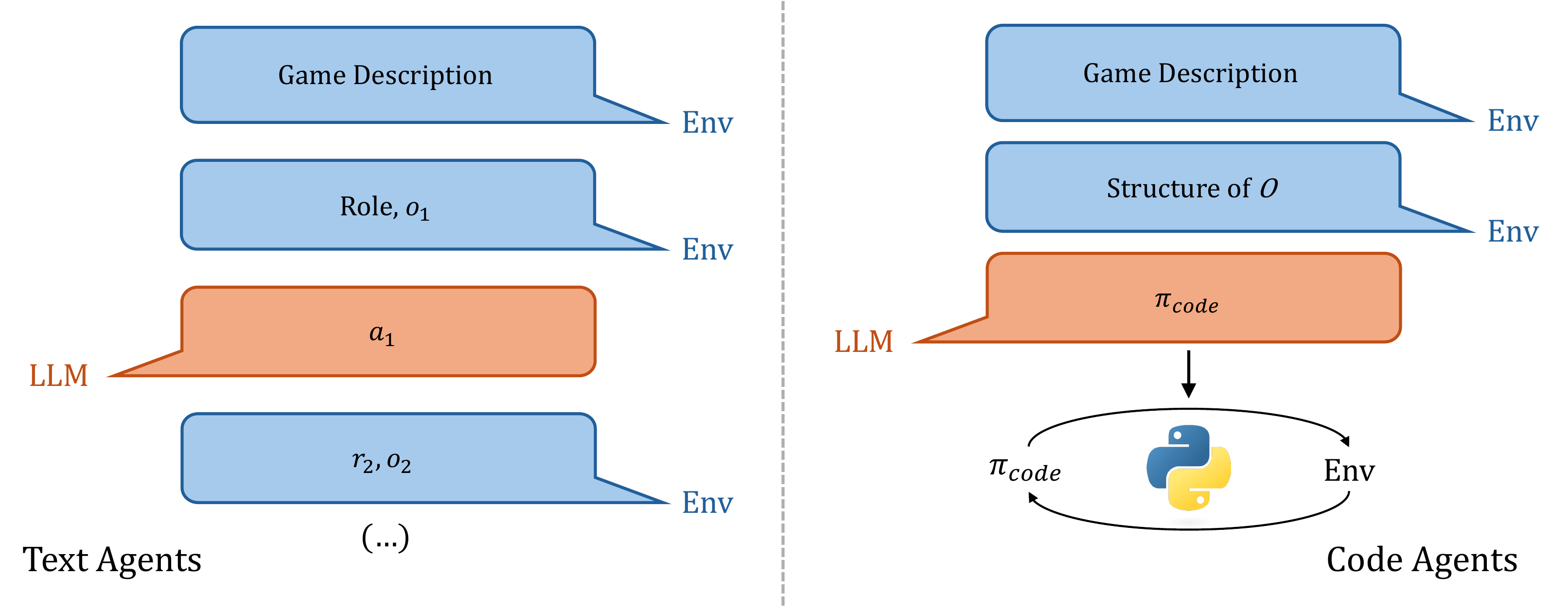}
    \caption{
    \textbf{Illustration of text and code agents.} While text agents only act out the local policy associated with their assigned roles, each code agent produces a joint policy in Python code. Code agents receive additional information on the structure of the observation space.
    }
    \label{fig:text_and_code_agents}
\end{figure*}
\begin{figure}
    \centering
    \includegraphics[width=0.8\linewidth]{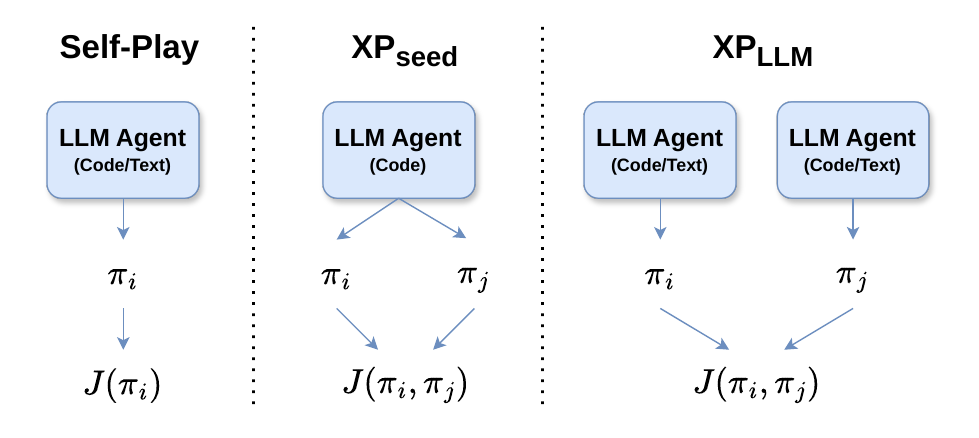}
    \caption{\textbf{Illustration of the evaluation protocols for text and code agents.} Note that $\text{XP}_{\text{seed}}$ is defined only for code agents, as they generate joint policies, whereas text agents induce only local policies. While text agents induce a joint policy which can be obtained by varying the role assignment (see \autoref{fig:text_and_code_agents}), they tend to significantly underperform code agents in \acl{sp} as the implicit policies are often not self-compatible (see \autoref{fig:local_policies_all}).}
    \label{fig:XP_schematic}
\end{figure}
\label{sec:app/schematic}
We present a schematic diagram of the two agents scaffolds in \autoref{fig:text_and_code_agents} and the different evaluation schemes in \autoref{fig:XP_schematic}.

\section{Additional details}

\subsection{Coordination Failure Rate}
In addition to reporting joint rewards, we measure the coordination failure rate, which captures the frequency of catastrophic coordination failures arising from paired local policies. For example, in Tiny Hanabi, a grounded joint policy that consistently reaches $(B,B)$ achieves a guaranteed reward of $8$. In contrast, we observe paired policies that attain the maximum reward of $10$ in $80\%$ of episodes but receive a reward of $0$ in the remaining $20\%$, resulting in the same XP score. Despite the same average score, the grounded policy is clearly more robust. To capture this, we report coordination failure rates. An episode is classified as a failure if the agents’ joint reward is at or below the threshold specified in \autoref{tab:thresholds}.

\subsection{Text Agents with \acs{cfd}}
\label{sec:app/text_agents_with_cfd}
\begin{table*}[t]
\centering
\scriptsize
\setlength{\tabcolsep}{8pt}
\caption{\acs{sp} performance comparison of text agents (mean $\pm$ SEM) across environments, evaluated with and without the \acs{cfd}. Results are shown for Grok~4.1~Fast and GPT-5.2. The coordination failure rate captures the frequency of catastrophic coordination breakdowns. Given the simplicity of these environments, we require coordination failure rates to be near zero in order to consider coordination robust. We report the mean and standard error over 100 episodes for single-turn games and 50 episodes for multi-turn games. We find that, with the exception of Narrow Tunnel and Permuted Levers, both Grok 4.1 Fast and GPT-5.2 exhibit high coordination failure rates.}
\begin{tabular}{llcccc}
\toprule
 &  & \multicolumn{2}{c}{w/o CFD} & \multicolumn{2}{c}{w/ CFD} \\
\cmidrule(lr){3-4} \cmidrule(lr){5-6}
Env & Model & SP Score & Coord.\ failure rate & SP Score & Coord.\ failure rate \\
\midrule
\multirow{2}{*}{Cat and Dog}
& Grok 4.1 Fast
& \meansem{-0.75}{0.99} & \meansem{0.53}{0.05}
& \meansem{5.00}{0.00} & \meansem{0.00}{0.00} \\
& GPT 5.2
& \meansem{4.50}{0.49} & \meansem{0.09}{0.03}
& \meansem{5.00}{0.00} & \meansem{0.00}{0.00} \\[4pt]

\multirow{2}{*}{Ad Hoc Conv.}
& Grok 4.1 Fast
& \meansem{95.80}{2.65} & \meansem{0.04}{0.03}
& \meansem{50.00}{0.00} & \meansem{0.00}{0.00} \\
& GPT 5.2
& \meansem{82.71}{4.01} & \meansem{0.16}{0.05}
& \meansem{50.00}{0.00} & \meansem{0.00}{0.00} \\[4pt]

\multirow{2}{*}{Ad Hoc Conv. PS}
& Grok 4.1 Fast
& \meansem{82.01}{3.30} & \meansem{0.08}{0.04}
& \meansem{49.64}{0.28} & \meansem{0.04}{0.03} \\
& GPT-5.2
& \meansem{76.95}{3.81} & \meansem{0.14}{0.05}
& \meansem{49.91}{0.09} & \meansem{0.02}{0.02} \\[4pt]

\multirow{2}{*}{Tiny Hanabi}
& Grok 4.1 Fast
& \meansem{7.38}{0.41} & \meansem{0.23}{0.04}
& \meansem{8.22}{0.21} & \meansem{0.05}{0.02} \\
& GPT-5.2
& \meansem{8.92}{0.30} & \meansem{0.10}{0.03}
& \meansem{7.64}{0.37} & \meansem{0.18}{0.04} \\[4pt]

\multirow{2}{*}{Tiny Hanabi Fixed}
& Grok 4.1 Fast
& \meansem{7.80}{0.40} & \meansem{0.21}{0.04}
& \meansem{7.78}{0.14} & \meansem{0.03}{0.02} \\
& GPT-5.2
& \meansem{7.88}{0.40} & \meansem{0.20}{0.04}
& \meansem{7.20}{0.29} & \meansem{0.13}{0.03} \\[4pt]

\multirow{2}{*}{Brittle Bridge}
& Grok 4.1 Fast
& \meansem{-149.69}{48.5} & \meansem{0.14}{0.05}
& \meansem{-15.86}{0.49} & \meansem{0.30}{0.06} \\
& GPT-5.2
& \meansem{-13.00}{0.48} & \meansem{0.10}{0.04}
& \meansem{-13.72}{0.50} & \meansem{0.10}{0.04} \\[4pt]

\multirow{2}{*}{Narrow Tunnel}
& Grok 4.1 Fast
& \meansem{-5.58}{0.17} & \meansem{0.00}{0.00}
& \meansem{-5.94}{0.19} & \meansem{0.00}{0.00} \\
& GPT-5.2
& \meansem{-8.92}{0.48} & \meansem{0.02}{0.02}
& \meansem{-8.38}{0.45} & \meansem{0.02}{0.02} \\[4pt]

\multirow{2}{*}{Matrix Game}
& Grok 4.1 Fast
& \meansem{0.62}{0.05} & \meansem{0.34}{0.05}
& \meansem{0.97}{0.02} & \meansem{0.03}{0.02} \\
& GPT-5.2
& \meansem{0.78}{0.04} & \meansem{0.22}{0.04}
& \meansem{0.84}{0.04} & \meansem{0.16}{0.04} \\[4pt]

\multirow{2}{*}{Permuted Levers}
& Grok 4.1 Fast
& \meansem{6.08}{0.36} & \meansem{0.02}{0.02}
& \meansem{6.28}{0.41} & \meansem{0.02}{0.02} \\
& GPT-5.2
& \meansem{7.24}{0.38} & \meansem{0.02}{0.02}
& \meansem{6.88}{0.39} & \meansem{0.02}{0.02} \\

\bottomrule
\end{tabular}
\label{tab:text_xp_summary}
\end{table*}

We report the \acs{sp} scores of the text agents with and without the hand-crafted \acs{cfd} in the fixed LLM setting for Grok 4.1 Fast and GPT-5.2 in \autoref{tab:text_xp_summary}. We find that the agents exhibit high coordination failure rates and low \acs{sp} scores across multiple environments, indicating poor coordination. Introducing the hand-crafted \acs{cfd} improves coordination in several environments, with a larger effect on Grok 4.1 Fast than on GPT-5.2. However, as discussed in \appref{sec:app/code_agents_with_cfd}, the \acs{cfd} can be overly restrictive. This is also evident in Brittle Bridge, where agents enter a deadlock, each waiting for the other to step onto the bridge to avoid collapse.

\subsection{Code Agents with \acs{cfd}}
\label{sec:app/code_agents_with_cfd}
\begin{table}[t]
\centering
\caption{Thresholds used to define coordination failures for each environment. An episode counts as a failure if its joint reward is at or below the threshold.}
\label{tab:thresholds}
\begin{tabular}{l c}
\toprule
\textbf{Environment} & \textbf{Threshold} \\
\midrule
Permuted Levers & $0.0$ \\
Cat and Dog & $0.0$ \\
Brittle Bridge & $-20.0$ \\
Narrow Tunnel & $-20.0$ \\
Ad-hoc Conventions & $49.5$ \\
Ad-hoc Conventions (Permuted Signals) & $49.5$ \\
Tiny Hanabi Fixed & $0.0$ \\
Tiny Hanabi & $0.0$ \\
Matrix Game & $0.0$ \\
\bottomrule
\end{tabular}
\end{table}

We report the \acs{sp} and \acs{xp} scores of the code agents under both the \textit{fixed} LLM setting (denoted $\text{XP}_{\text{seed}}$) and the \textit{variable} LLM setting (denoted $\text{XP}_{\text{LLM}}$) in \autoref{tab:code-xp-summary}. The scores are calculated as follows. First, we generate six independent \acl{xp} matrices, with each \acs{llm} contributing one joint policy to each matrix. Every joint-policy pairing is evaluated over 100 episodes. We then average scores across the six XP matrices to obtain a single $(N_{\text{LLM}} \times N_{\text{LLM}})$ matrix. For \acs{sp}, we report the mean and standard error over the diagonal entries. For variable-\acs{llm} $\text{XP}_{\text{LLM}}$, we compute the mean and standard error over the off-diagonal entries in each \acs{llm}$'$s corresponding row and column. For fixed-\acs{llm} $\text{XP}_{\text{seed}}$, we pair joint policies produced by the same \acs{llm} across different XP matrices into disjoint pairs (e.g., $(1,2)$, $(3,4)$, \ldots), evaluate each pair jointly, and report the resulting mean and standard error. We do the same when reporting the coordination failure rates in \autoref{tab:xp-summary-failure}.

We observe a substantial SP–XP gap across most environments, and in some cases (e.g., Brittle Bridge), agents fail to generate even high-performing self-play policies, indicating a limited understanding of the underlying coordination challenge. Introducing the hand-crafted \acp{cfd} yields only modest gains, improving performance primarily in Cat \& Dog and the semantic selection game.


\subsection{Trampoline Tiger Game}
\label{sec:app/tiger_trampline_game}
\begin{figure}
    \centering
    \begin{subfigure}[t]{0.48\textwidth}
        \centering
        \includegraphics[width=\textwidth]{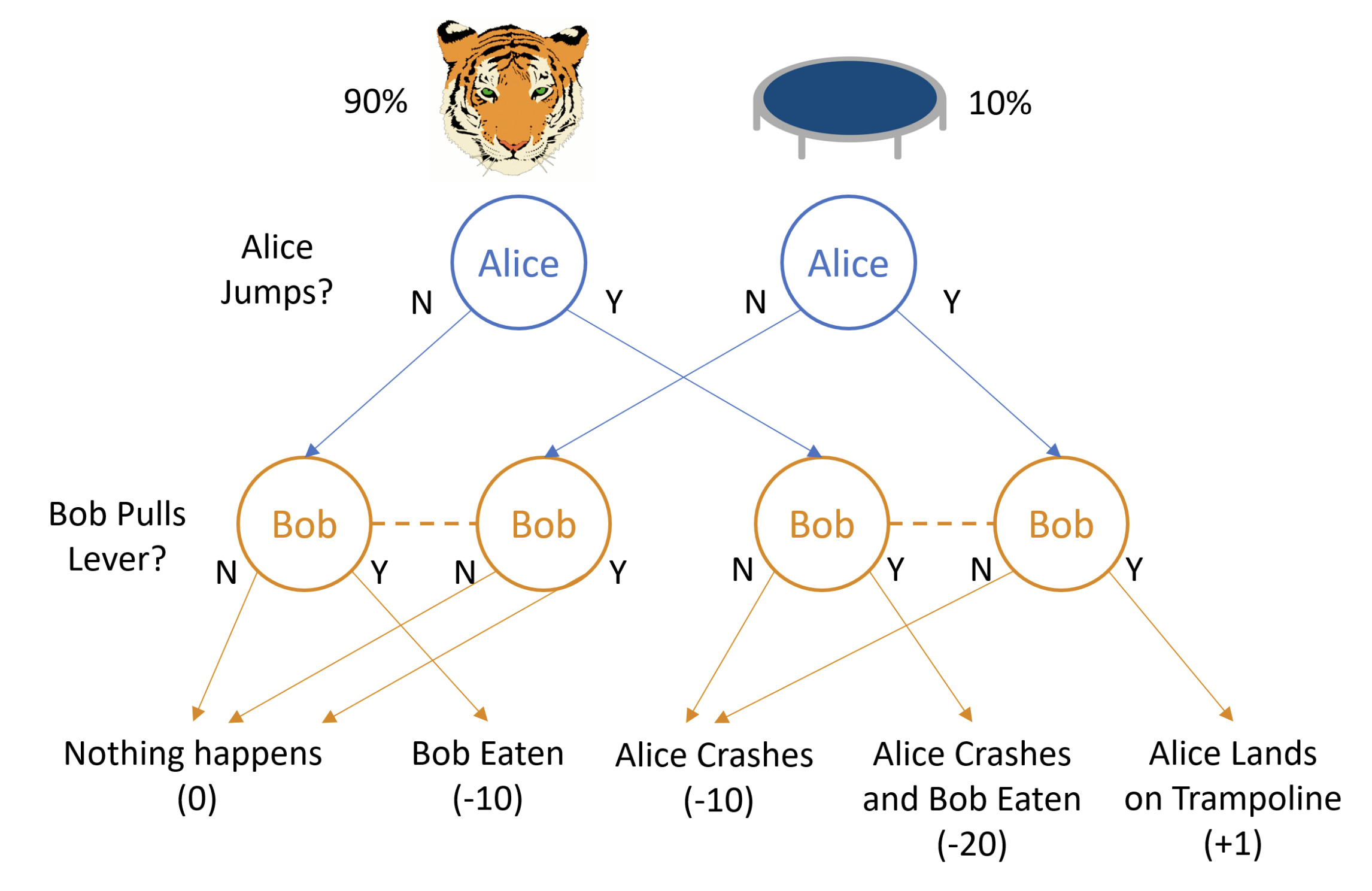}
        \caption{An illustration of the trampoline tiger environment~\cite{hu2022selfexplainingdeviationscoordination}. Alice observes whether Bob’s lever would deploy a tiger or a trampoline and then chooses whether to jump. Bob observes only Alice’s action and subsequently decides whether to pull the lever. Dotted lines between Bob’s decision nodes indicate information sets that Bob cannot distinguish.}
        \label{fig:semantic/trampoline_tiger_illustration}
    \end{subfigure}
    \hfill
    \begin{subfigure}[t]{0.48\textwidth}
        \centering
        \includegraphics[width=\textwidth]{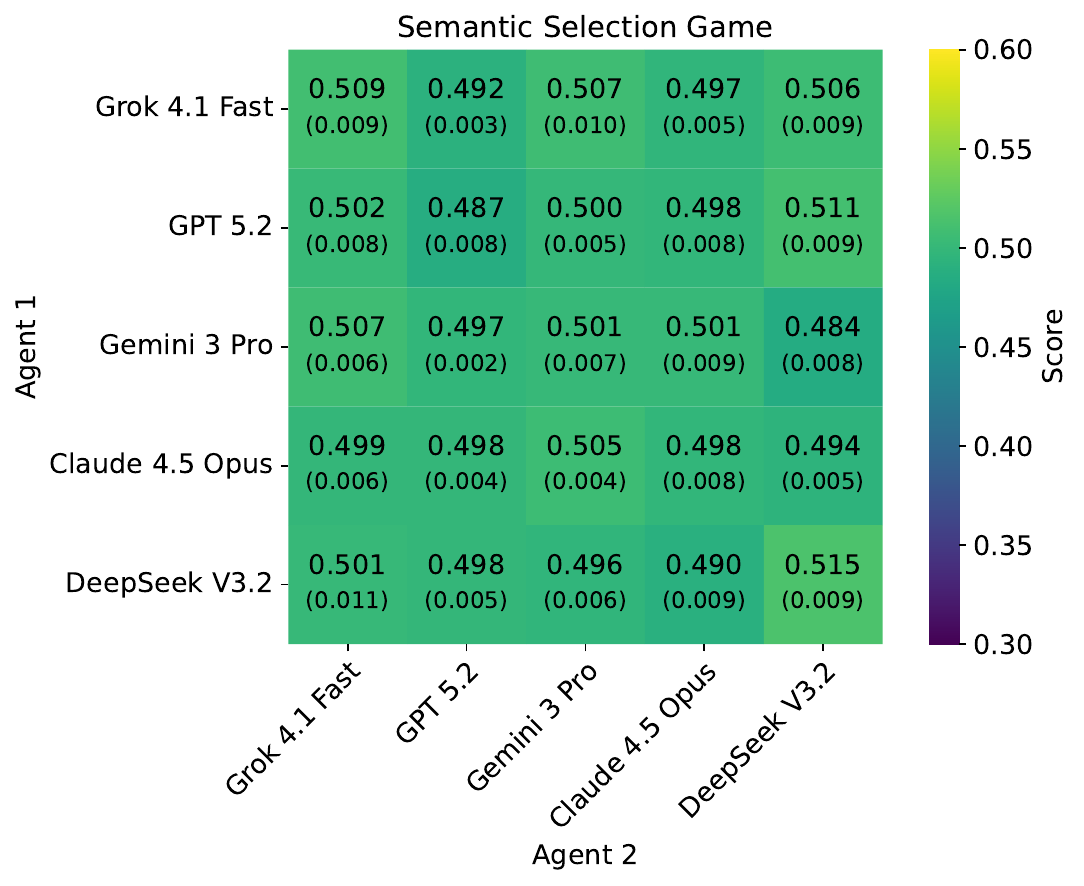}
        \caption{\acs{xp} scores of code agents in the trampoline tiger game~\cite{hu2022selfexplainingdeviationscoordination}. Scores are averaged across six different \acs{xp} matrices; each cell reports the mean return over 100 episodes. For our experiments, we use $p_{\text{tiger}} = 0.5$ for equal coverage of tiger and non-tiger states. The \acs{llm} agents are not aware of $p_{\text{tiger}}$.}
        \label{fig:semantic/trampoline_tiger_xp}
    \end{subfigure}
    \caption{\textbf{The trampoline tiger environment.}}
    \label{fig:semantic/trampoline_tiger}
\end{figure}

The tiger trampoline game, shown in \autoref{fig:semantic/trampoline_tiger_illustration}, is a coordination game introduced in~\citep{hu2022selfexplainingdeviationscoordination} to illustrate the concept of \textit{self-explaining deviations (SED)}. A self-explaining deviation is an action by a teammate in a fully cooperative setting that, at first, appears to hurt the team's performance. But if players act under the belief that their teammates' actions are rational, those actions can signal hidden information to them, ultimately aiding the team. In the trampoline tiger game, Bob controls a lever that can release a hidden object: either a tiger or a trampoline. Alice stands above and can observe the hidden state, but cannot communicate further with Bob. She has the option to jump, which yields only a small positive joint reward if the hidden state is the trampoline and Bob pulls the lever. Bob can observe whether Alice jumps before pulling the lever. If Alice acts rationally, she will only decide to jump if the hidden state is a trampoline, thereby signalling it to Bob. Notably, never jumping and never pulling the lever is a pure Nash equilibrium. Therefore, algorithms that independently learn Alice's and Bob's policies risk getting stuck in this local maximum. \citet{hu2022selfexplainingdeviationscoordination} showed that popular MARL algorithms like MAPPO~\citep{yu2022mappo} and QMIX~\citep{rashid2018qmix} converge to this Nash Equilibrium in most cases. LLM agents naturally consider bilateral deviations of Alice's and Bob's actions by directly reasoning about the environment dynamics. As shown in \autoref{fig:semantic/trampoline_tiger_xp}, this allows a variety of \acs{llm} agents to determine the optimal policy in this game.

\subsection{Local Policies of Text Agents}
\label{sec:app/local_policies}
\begin{figure*}
    \centering    \includegraphics[width=\linewidth]{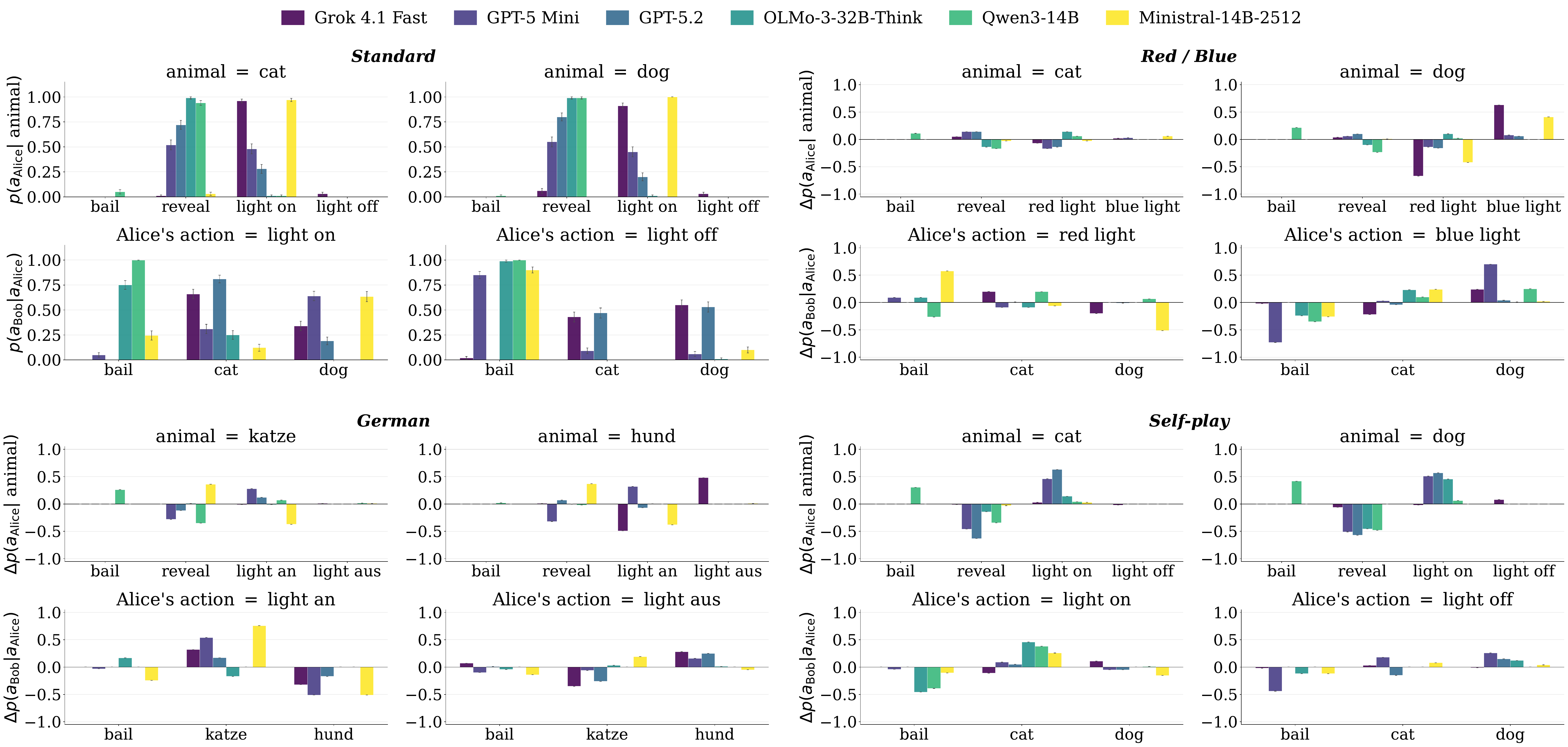}
    \caption{\textbf{Local policies of text agents in Cat \& Dog and its variants.}
The top-left plot shows action probabilities for the standard Cat \& Dog environment; each conditional distribution $p(\cdot \mid c)$ is estimated from 100 samples obtained by repeatedly prompting the model from a fixed game state.
The remaining plots show the delta, i.e., the change in action probabilities relative to the standard condition, when the environment semantics are varied (changed action labels, German translation, self-play).
}
    \label{fig:local_policies_all}
\end{figure*}
\begin{figure*}
    \centering
    \includegraphics[width=\linewidth]{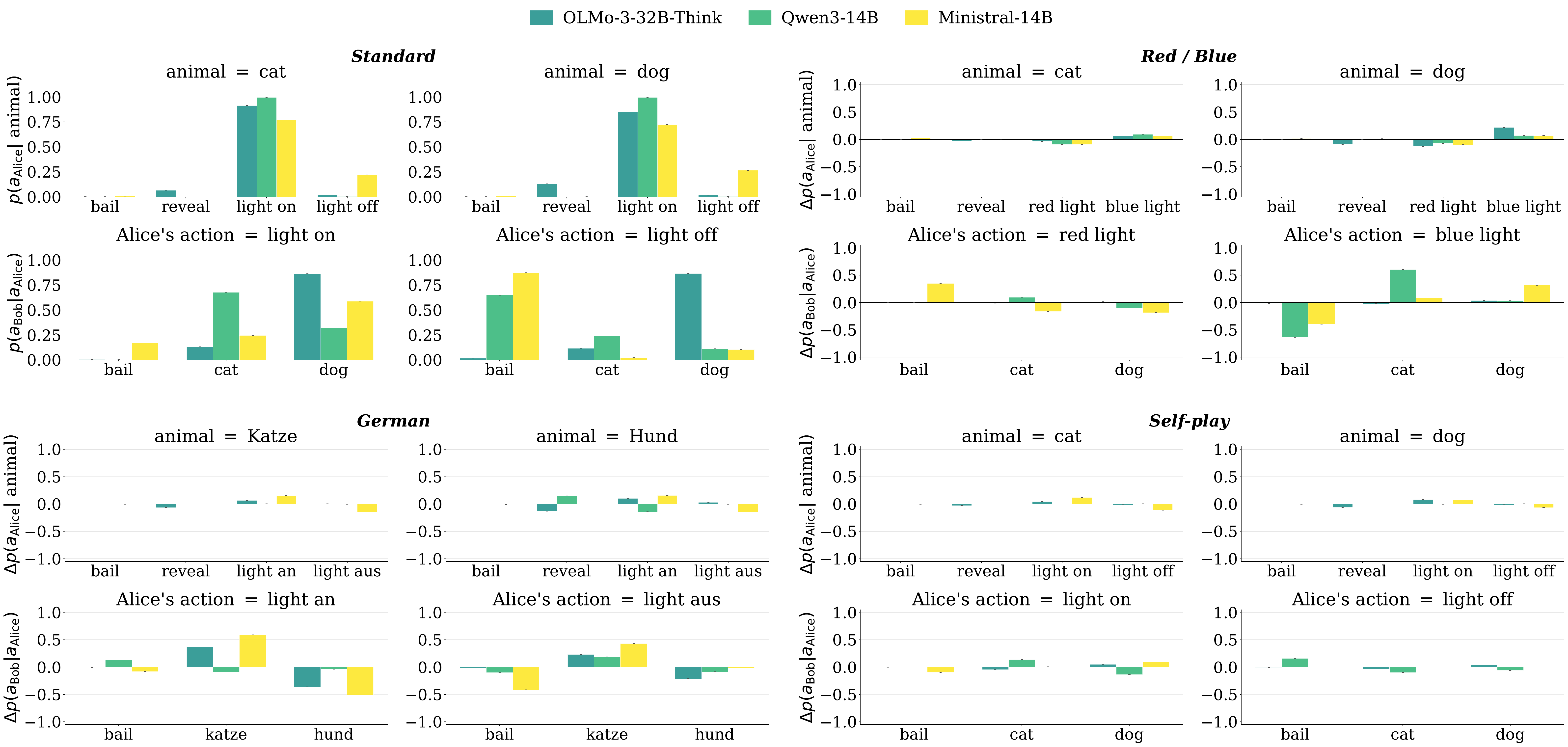}
    \caption{\textbf{Token-probability local policies of text agents in Cat \& Dog and its variants.}
The top-left plot shows action probabilities for the standard Cat \& Dog environment, computed from token-level log-probabilities rather than sampling.
The remaining plots show the delta, i.e., the change in action probabilities relative to the standard condition, when the environment semantics are varied (changed action labels, German translation, self-play).
}
    \label{fig:logprob_local_policy_all}
\end{figure*}

Since text agents interact with the environment by producing actions sequentially, rather than explicitly specifying joint policies as code, their decision-making is less interpretable. To better understand their behaviour, we empirically estimate their induced local policies.

We estimate the local policies $p_{\text{alice}}(\cdot \mid \text{animal})$ and $p_{\text{bob}}(\cdot \mid a_{\text{alice}})$ by fixing the environment state and repeatedly prompting the agent $100$ times as shown in \autoref{fig:local_policies_all}. Next, we study how these local policies change under variations in the game’s semantics $\mathcal{D}$. To this end, we consider three variants: (1) replacing the binary light-bulb signal (\text{``on/off''}) with a colour choice (\text{``red/blue''}), (2) translating the natural-language descriptions into German, (3) the agent is informed that it is being evaluated against an identical instance of itself.

Note that the agent is given complete information about the Dec-POMDP structure, including the payoffs. Consequently, the natural-language description of the MDP should not affect the induced local policy as long as the underlying Dec-POMDP structure is preserved, with the exception of the final setting, in which the agent has the option to use a convention to maximise reward. Despite this, we observe substantial differences in the induced local policies across semantic variants, as shown in \autoref{fig:local_policies_all}. In particular, the \acs{sp} variant leads to a drastic reduction in \acs{xp} scores, as discussed in \autoref{sec:zsc_performance}.

We also probe whether model internals, specifically token-level action probabilities and chain-of-thought reasoning tokens, are consistent with the empirical local policies and the explicit reasoning outputs elicited before each action. We focus on open-weight models that expose these internals: OLMo-3-32B-Think, Qwen3-14B, and Ministral-14B.

To estimate token-level action probabilities, we compute the log-probability of each legal action completion under a fixed prompt and normalise across legal actions using a softmax. This requires disabling chain-of-thought reasoning, since reasoning tokens precede the action token and would otherwise confound the estimate. We find that disabling reasoning substantially degrades policy quality: models such as OLMo-3-32B-Think and Qwen3-14B, which exhibit strong empirical local policies when reasoning is enabled, produce markedly worse token-level policies in this setting (\autoref{fig:logprob_local_policy_all}), an effect we attribute to the removal of the reasoning step.

When reasoning is re-enabled, the generated chain-of-thought tokens closely match the content of the explicit reasoning outputs, indicating that the reasoning is faithful to the model's decision process. This supports the validity of our reasoning-based analysis for closed-weight models such as GPT-5.2, for which token-level internals are inaccessible.

\subsection{Reasoning Effort}
\label{sec:app/reasoning_effort}
\begin{table*}[t]
\centering
\setlength{\tabcolsep}{8pt}
\caption{
\textbf{\acs{xp} scores of text agents under varying levels of reasoning effort on Cat \& Dog.} We evaluate GPT-5 mini and Grok 4.1 Fast under varying levels of reasoning effort, as well as with and without the use of our \acs{cfd}. XP scores are reported as mean $\pm$ standard error over 100 games.
}
\begin{tabular}{llcc}
\toprule
 &  & \multicolumn{2}{c}{Mean return} \\
\cmidrule(lr){3-4}
Model & Reasoning effort & w/o CFD & w/ CFD \\
\midrule
\multirow{3}{*}{GPT-5 mini}
 & Low    & $3.23\,{\scriptstyle \pm 0.56}$ & $4.47\,{\scriptstyle \pm 0.14}$ \\
 & Medium & $0.97\,{\scriptstyle \pm 0.81}$ & $4.88\,{\scriptstyle \pm 0.07}$ \\
 & High   & $0.86\,{\scriptstyle \pm 0.91}$ & $4.92\,{\scriptstyle \pm 0.06}$ \\
\midrule
\multirow{3}{*}{Grok 4.1 Fast}
 & Low    & $1.20\,{\scriptstyle \pm 0.98}$ & $5.00\,{\scriptstyle \pm 0.00}$ \\
 & Medium & $-0.75\,{\scriptstyle \pm 0.99}$ & $5.00\,{\scriptstyle \pm 0.00}$ \\
 & High   & $0.41\,{\scriptstyle \pm 0.99}$ & $5.00\,{\scriptstyle \pm 0.00}$ \\
\bottomrule
\end{tabular}

\label{tab:reasoning_effort}
\end{table*}

\begin{table*}[t]
\centering
\scriptsize
\setlength{\tabcolsep}{8pt}
\caption{
\textbf{Action frequencies of frontier text agents as Alice in Cat \& Dog under varying reasoning effort.} We report how often Alice bails, reveals the animal by removing the wall (the \acs{zsc}-optimal action), or uses the light as a signal. As in \autoref{fig:local_policies_cat_dog}, frequencies are estimated by repeatedly prompting each agent $100$ times from a fixed game state. Values are reported as mean $\pm$ standard error. Higher reasoning effort increases the reveal rate for all three models, most strongly for Claude Opus~5.
}
\begin{tabular}{llccc}
\toprule
Model & Reasoning effort & Bail & Reveal & Signal \\
\midrule
\multirow{3}{*}{GPT-5.6 Sol}
 & Low    & $\meansem{0.00}{0.00}$ & $\meansem{0.06}{0.02}$ & $\meansem{0.94}{0.02}$ \\
 & Medium & $\meansem{0.00}{0.00}$ & $\meansem{0.06}{0.02}$ & $\meansem{0.95}{0.02}$ \\
 & High   & $\meansem{0.00}{0.00}$ & $\meansem{0.11}{0.02}$ & $\meansem{0.89}{0.02}$ \\
\midrule
\multirow{3}{*}{Claude Opus~5}
 & Low    & $\meansem{0.00}{0.00}$ & $\meansem{0.00}{0.00}$ & $\meansem{1.00}{0.00}$ \\
 & Medium & $\meansem{0.00}{0.00}$ & $\meansem{0.13}{0.02}$ & $\meansem{0.87}{0.02}$ \\
 & High   & $\meansem{0.00}{0.00}$ & $\meansem{0.74}{0.03}$ & $\meansem{0.26}{0.03}$ \\
\midrule
\multirow{3}{*}{Grok~4.5}
 & Low    & $\meansem{0.00}{0.00}$ & $\meansem{0.26}{0.03}$ & $\meansem{0.75}{0.03}$ \\
 & Medium & $\meansem{0.00}{0.00}$ & $\meansem{0.46}{0.04}$ & $\meansem{0.54}{0.04}$ \\
 & High   & $\meansem{0.00}{0.00}$ & $\meansem{0.49}{0.04}$ & $\meansem{0.51}{0.04}$ \\
\bottomrule
\end{tabular}
\label{tab:reasoning_effort_frontier}
\end{table*}

As noted earlier, coordination requires reasoning about a partner’s behaviour (\acs{tom}). In formal reasoning tasks, increasing reasoning effort, i.e the budget of reasoning tokens available, often yields substantial performance gains. We therefore examine how varying reasoning effort affects the performance of text agents, as reported in \autoref{tab:reasoning_effort}. Surprisingly, for the smaller models (Grok 4.1 Fast and GPT-5 Mini), reasoning effort is negatively correlated with XP score. Further analysis shows that higher reasoning effort increases the likelihood that these models hallucinate a shared convention and assume that their partner will adopt the same convention. That said, we find that the hand-crafted CFD facilitates robust coordination even under increased reasoning effort.

We additionally evaluate frontier models (GPT-5.6 Sol, Claude Opus~5, and Grok~4.5) as text agents in Cat \& Dog (\autoref{tab:reasoning_effort_frontier}). In contrast to the smaller models, increasing reasoning effort \emph{improves} the reveal rate for all three, most strongly for Claude Opus~5 (from $0\%$ to $74\%$). At the same time, we observe degradation within model families: Qwen3-Max-Thinking performs worse than Qwen3-14B, and GPT-5.6 Sol reveals far less often than GPT-5.2 (\autoref{tab:reasoning_effort_frontier} and \autoref{fig:local_policies_cat_dog}), despite being positioned as a substantial advance in overall capability. Taken together, these results suggest a \textbf{mixed relationship between reasoning capacity and \acs{zsc} performance}: more capable models or higher reasoning effort do not reliably translate into stronger zero-shot coordination.

\subsection{Few-Shot Demonstrations}
\label{sec:app/few_shot}
\begin{table*}[t]
\centering
\scriptsize
\setlength{\tabcolsep}{8pt}
\caption{
\textbf{\acs{xp} scores of text agents in Cat \& Dog without a \acs{cfd}, with few-shot demonstrations, and with the hand-crafted \acs{cfd}.} Rows denote the \acs{llm} playing Alice and columns the \acs{llm} playing Bob. The few-shot prompts contain the environment description and a hand-crafted rollout of the \acs{zsc}-optimal strategy for Ad-Hoc Conventions, Tiny Hanabi Fixed, and Brittle Bridge. Scores are mean $\pm$ standard error over 5 games per cell. The safe fallback of removing the wall yields a score of 5.
}
\begin{tabular}{lcccc}
\toprule
Alice / Bob & Grok~4.5 & Gemini~3.1~Pro & DeepSeek~V3.2 & GPT-5.2 \\
\midrule
\multicolumn{5}{c}{\textit{No \acs{cfd}}} \\
\midrule
Grok~4.5       & $\meansem{0.00}{4.18}$  & $\meansem{4.00}{3.67}$ & $\meansem{6.00}{1.00}$ & $\meansem{3.00}{3.39}$ \\
Gemini~3.1~Pro & $\meansem{5.00}{0.00}$  & $\meansem{5.00}{0.00}$ & $\meansem{5.00}{0.00}$ & $\meansem{5.00}{0.00}$ \\
DeepSeek~V3.2  & $\meansem{-1.00}{3.67}$ & $\meansem{3.00}{3.39}$ & $\meansem{7.00}{1.22}$ & $\meansem{-1.00}{3.67}$ \\
GPT-5.2        & $\meansem{5.00}{0.00}$  & $\meansem{2.00}{3.00}$ & $\meansem{-1.00}{3.67}$ & $\meansem{6.00}{1.00}$ \\
\midrule
\multicolumn{5}{c}{\textit{Few-shot demonstrations}} \\
\midrule
Grok~4.5       & $\meansem{5.00}{0.00}$  & $\meansem{-1.00}{3.67}$ & $\meansem{5.00}{0.00}$ & $\meansem{5.00}{0.00}$ \\
Gemini~3.1~Pro & $\meansem{5.00}{0.00}$  & $\meansem{5.00}{0.00}$ & $\meansem{5.00}{0.00}$ & $\meansem{5.00}{0.00}$ \\
DeepSeek~V3.2  & $\meansem{7.00}{1.22}$  & $\meansem{7.00}{1.22}$ & $\meansem{-1.00}{3.67}$ & $\meansem{6.00}{1.00}$ \\
GPT-5.2        & $\meansem{2.00}{3.00}$  & $\meansem{6.00}{1.00}$ & $\meansem{6.00}{1.00}$ & $\meansem{2.00}{3.00}$ \\
\midrule
\multicolumn{5}{c}{\textit{Hand-crafted \acs{cfd}}} \\
\midrule
Grok~4.5       & $\meansem{5.00}{0.00}$ & $\meansem{5.00}{0.00}$ & $\meansem{5.00}{0.00}$ & $\meansem{5.00}{0.00}$ \\
Gemini~3.1~Pro & $\meansem{5.00}{0.00}$ & $\meansem{5.00}{0.00}$ & $\meansem{5.00}{0.00}$ & $\meansem{5.00}{0.00}$ \\
DeepSeek~V3.2  & $\meansem{5.00}{0.00}$ & $\meansem{5.00}{0.00}$ & $\meansem{5.00}{0.00}$ & $\meansem{5.00}{0.00}$ \\
GPT-5.2        & $\meansem{5.00}{0.00}$ & $\meansem{5.00}{0.00}$ & $\meansem{5.00}{0.00}$ & $\meansem{5.00}{0.00}$ \\
\bottomrule
\end{tabular}
\label{tab:few_shot_xp}
\end{table*}

A simple alternative to \acp{cfd} is to provide agents with in-context demonstrations of successful coordination. Whereas a \acs{cfd} states the principles underlying successful coordination explicitly, an agent learning from demonstrations must first infer these principles implicitly before transferring them to an unseen environment. To compare the two, we rerun the Cat \& Dog \acs{xp} experiment from \autoref{fig:combined} with prompts augmented by examples of successful coordination from three other environments. For each, we include the environment description and a hand-crafted rollout of the \acs{zsc}-optimal strategy: establishing and maintaining a stable convention across rounds in Ad-Hoc Conventions; selecting the safe fallback $(B, B)$ in Tiny Hanabi Fixed, whose payoff structure closely resembles Cat \& Dog; and avoiding bridge collapse by retreating when both players enter the bridge, while establishing an ad-hoc leader--follower convention, in Brittle Bridge. As Grok~4.1~Fast had been deprecated by the time of this experiment, we replace it with Grok~4.5 and regenerate the full \acs{xp} matrices for all three settings. As shown in \autoref{tab:few_shot_xp}, only the \acs{cfd} enables reliable coordination across all pairs of agents. While the small sample size makes it difficult to determine whether few-shot demonstrations provide any marginal benefit, they perform substantially worse than the \acs{cfd}.




\begin{table*}[ht]
\centering
\tiny
\caption{
\textbf{\acs{sp} and \acs{xp} scores of code agents across environments, under fixed ($\text{XP}_{\text{seed}}$) and variable ($\text{XP}_{\text{LLM}}$) \acs{llm} settings, with and without \acs{cfd}.}
Agents tested: GPT-5.2, Grok~4.1~Fast, Claude~Opus~4.5, Gemini~3~Pro, and Deepseek V3.2 Terminus.
}
\begin{tabular}{l l c c c c c c}
\toprule
\multicolumn{2}{c}{ } &
\multicolumn{3}{c}{w/o CFD} & \multicolumn{3}{c}{w/ CFD} \\
\cmidrule(lr){3-5} \cmidrule(lr){6-8}
Env & LLM & $\text{SP}$ & $\text{XP}_{\text{seed}}$ & $\text{XP}_{\text{LLM}}$ & $\text{SP}$ & $\text{XP}_{\text{seed}}$ & $\text{XP}_{\text{LLM}}$ \\
\midrule

\multirow{6}{*}{C.D.}
 & Grok       & \meansem{10.00}{0.00} & \meansem{3.33}{5.44} & \meansem{0.68}{1.75}   & \meansem{6.67}{0.96}  & \meansem{3.50}{0.61} & \meansem{3.94}{0.37} \\
 & GPT        & \meansem{9.58}{0.38}  & \meansem{9.54}{0.37}  & \meansem{-3.03}{1.88}  & \meansem{5.00}{0.00}  & \meansem{5.00}{0.00} & \meansem{4.50}{0.38} \\
 & Gemini     & \meansem{10.00}{0.00} & \meansem{10.00}{0.00} & \meansem{-1.63}{2.05}  & \meansem{5.00}{0.00}  & \meansem{5.00}{0.00} & \meansem{4.72}{0.18} \\
 & Claude     & \meansem{9.17}{0.76}  & \meansem{-7.50}{2.04} & \meansem{1.20}{0.55}   & \meansem{5.00}{0.00}  & \meansem{5.00}{0.00} & \meansem{4.72}{0.18} \\
 & DeepSeek   & \meansem{10.00}{0.00} & \meansem{-3.33}{5.44} & \meansem{0.26}{0.86}   & \meansem{5.83}{0.76}  & \meansem{4.25}{0.61} & \meansem{4.44}{0.18} \\
\cmidrule(lr){2-8}
 & \textit{Mean} & \meansem{9.75}{0.15}  & \meansem{2.41}{2.40}  & \meansem{-0.51}{1.04}  & \meansem{5.50}{0.30}  & \meansem{4.55}{0.23} & \meansem{4.46}{0.19} \\
\midrule

\multirow{6}{*}{A.C.}
 & Grok       & \meansem{100.00}{0.00} & \meansem{100.00}{0.00} & \meansem{9.64}{28.74}  & \meansem{54.12}{3.76} & \meansem{46.18}{3.12} & \meansem{39.53}{3.53} \\
 & GPT        & \meansem{83.33}{9.62}  & \meansem{83.33}{6.80}  & \meansem{62.20}{9.21}  & \meansem{58.33}{7.61} & \meansem{42.50}{6.12} & \meansem{38.55}{3.36} \\
 & Gemini     & \meansem{100.00}{0.00} & \meansem{100.00}{0.00} & \meansem{-5.92}{23.55} & \meansem{63.50}{7.79} & \meansem{35.00}{6.12} & \meansem{42.98}{4.41} \\
 & Claude     & \meansem{100.00}{0.00} & \meansem{33.33}{54.43} & \meansem{23.65}{16.36} & \meansem{75.00}{10.21}& \meansem{59.17}{17.49}& \meansem{36.15}{7.07} \\
 & DeepSeek   & \meansem{98.33}{0.96}  & \meansem{64.27}{28.16} & \meansem{37.90}{18.95} & \meansem{64.66}{11.10}& \meansem{18.23}{13.02}& \meansem{23.92}{4.08} \\
\cmidrule(lr){2-8}
 & \textit{Mean} & \meansem{96.33}{2.92}  & \meansem{76.19}{13.94} & \meansem{25.49}{13.96} & \meansem{63.12}{3.15} & \meansem{40.22}{5.88} & \meansem{36.23}{3.31} \\
\midrule

\multirow{6}{*}{A.C.P.S.}
 & Grok       & \meansem{69.73}{15.05} & \meansem{36.13}{3.16}  & \meansem{36.05}{7.47}  & \meansem{84.12}{2.87} & \meansem{19.33}{9.85} & \meansem{29.52}{6.05} \\
 & GPT        & \meansem{72.59}{7.33}  & \meansem{59.19}{11.77} & \meansem{64.24}{8.79}  & \meansem{57.44}{6.67} & \meansem{57.87}{3.42} & \meansem{46.69}{4.13} \\
 & Gemini     & \meansem{90.50}{0.00}  & \meansem{90.50}{0.00}  & \meansem{65.43}{9.06}  & \meansem{90.73}{0.16} & \meansem{90.17}{0.15} & \meansem{58.62}{7.08} \\
 & Claude     & \meansem{73.08}{15.17} & \meansem{79.05}{9.41}  & \meansem{59.42}{10.39} & \meansem{56.05}{19.35}& \meansem{38.35}{1.86} & \meansem{47.85}{7.97} \\
 & DeepSeek   & \meansem{61.08}{13.91} & \meansem{8.90}{4.12}   & \meansem{30.89}{7.40}  & \meansem{38.02}{9.49} & \meansem{-8.14}{9.84} & \meansem{35.99}{6.39} \\
\cmidrule(lr){2-8}
 & \textit{Mean} & \meansem{73.40}{4.28}  & \meansem{54.76}{8.25}  & \meansem{51.21}{6.41}  & \meansem{65.27}{8.70} & \meansem{39.51}{9.10} & \meansem{43.73}{4.66} \\
\midrule

\multirow{6}{*}{T. H.}
 & Grok       & \meansem{10.00}{0.00}  & \meansem{10.00}{0.00}  & \meansem{7.35}{1.00}   & \meansem{9.53}{0.31}  & \meansem{8.56}{1.18} & \meansem{7.08}{0.46} \\
 & GPT        & \meansem{9.67}{0.30}   & \meansem{8.72}{1.05}   & \meansem{7.27}{0.83}   & \meansem{9.67}{0.30}  & \meansem{8.95}{0.86} & \meansem{6.70}{0.92} \\
 & Gemini     & \meansem{10.00}{0.00}  & \meansem{10.00}{0.00}  & \meansem{7.32}{0.99}   & \meansem{10.00}{0.00} & \meansem{10.00}{0.00}& \meansem{7.18}{0.85} \\
 & Claude     & \meansem{9.32}{0.31}   & \meansem{2.64}{0.56}   & \meansem{6.03}{0.41}   & \meansem{9.00}{0.33}  & \meansem{4.01}{0.95} & \meansem{6.08}{0.49} \\
 & DeepSeek   & \meansem{8.42}{0.33}   & \meansem{5.63}{1.53}   & \meansem{4.41}{0.66}   & \meansem{7.80}{0.62}  & \meansem{6.87}{0.93} & \meansem{5.70}{0.85} \\
\cmidrule(lr){2-8}
 & \textit{Mean} & \meansem{9.48}{0.26}  & \meansem{7.40}{0.84}   & \meansem{6.48}{0.57}   & \meansem{9.20}{0.34}  & \meansem{7.68}{0.67} & \meansem{6.55}{0.49} \\
\midrule

\multirow{6}{*}{T.H.F}
 & Grok       & \meansem{10.00}{0.00}  & \meansem{10.00}{0.00}  & \meansem{9.71}{0.19}   & \meansem{9.67}{0.30}  & \meansem{8.00}{1.63} & \meansem{7.05}{0.91} \\
 & GPT        & \meansem{9.67}{0.30}   & \meansem{9.67}{0.27}   & \meansem{9.58}{0.22}   & \meansem{8.67}{0.38}  & \meansem{5.33}{1.09} & \meansem{6.57}{0.53} \\
 & Gemini     & \meansem{10.00}{0.00}  & \meansem{10.00}{0.00}  & \meansem{9.71}{0.19}   & \meansem{8.00}{0.00}  & \meansem{8.00}{0.00} & \meansem{8.42}{0.28} \\
 & Claude     & \meansem{10.00}{0.00}  & \meansem{10.00}{0.00}  & \meansem{9.71}{0.19}   & \meansem{9.67}{0.30}  & \meansem{8.00}{1.63} & \meansem{7.36}{0.89} \\
 & DeepSeek   & \meansem{9.67}{0.30}   & \meansem{8.00}{1.63}   & \meansem{8.96}{0.25}   & \meansem{8.00}{0.00}  & \meansem{8.00}{0.00} & \meansem{6.10}{0.87} \\
\cmidrule(lr){2-8}
 & \textit{Mean} & \meansem{9.87}{0.07}  & \meansem{9.53}{0.39}   & \meansem{9.53}{0.15}   & \meansem{8.80}{0.33}  & \meansem{7.47}{0.58} & \meansem{7.10}{0.50} \\
\midrule

\multirow{6}{*}{B.B}
 & Grok       & \meansem{-9.00}{0.00}   & \meansem{-9.00}{0.00}  & \meansem{-172.53}{72.54} & \meansem{-14.99}{4.88}  & \meansem{-28.40}{15.10} & \meansem{-123.77}{57.98} \\
 & GPT        & \meansem{-12.84}{0.40}  & \meansem{-12.76}{0.33} & \meansem{-24.48}{5.35}   & \meansem{-13.29}{0.70}  & \meansem{-12.23}{0.11} & \meansem{-16.68}{1.67} \\
 & Gemini     & \meansem{-10.00}{0.00}  & \meansem{-10.00}{0.00} & \meansem{-120.70}{69.36} & \meansem{-12.49}{0.09}  & \meansem{-12.51}{0.07} & \meansem{-17.44}{2.11} \\
 & Claude     & \meansem{-292.74}{102.20}& \meansem{-11.91}{0.09}& \meansem{-146.69}{36.19} & \meansem{-197.78}{55.21}& \meansem{-127.05}{31.52}& \meansem{-122.91}{50.84} \\
 & DeepSeek   & \meansem{-598.45}{180.85}& \meansem{-388.02}{153.55}& \meansem{-269.83}{76.67}& \meansem{-867.50}{121.87}& \meansem{-671.00}{269.44}& \meansem{-186.36}{68.69} \\
\cmidrule(lr){2-8}
 & \textit{Mean} & \meansem{-184.61}{104.65}& \meansem{-86.34}{49.60}& \meansem{-146.85}{41.20}& \meansem{-221.21}{147.99}& \meansem{-170.24}{85.17} & \meansem{-93.43}{32.78} \\
\midrule

\multirow{6}{*}{N.T}
 & Grok       & \meansem{-5.17}{0.15}   & \meansem{-5.17}{0.14}  & \meansem{-17.06}{5.40}  & \meansem{-5.00}{0.00}   & \meansem{-36.67}{25.86} & \meansem{-19.36}{6.98} \\
 & GPT        & \meansem{-5.33}{0.30}   & \meansem{-5.67}{0.27}  & \meansem{-13.33}{3.96}  & \meansem{-5.72}{0.34}   & \meansem{-6.03}{0.08} & \meansem{-7.88}{1.82} \\
 & Gemini     & \meansem{-5.00}{0.00}   & \meansem{-5.00}{0.00}  & \meansem{-17.00}{5.40}  & \meansem{-5.92}{0.03}   & \meansem{-5.96}{0.04} & \meansem{-6.03}{0.02} \\
 & Claude     & \meansem{-5.00}{0.00}   & \meansem{-68.33}{25.86}& \meansem{-34.83}{3.29}  & \meansem{-5.00}{0.00}   & \meansem{-68.33}{25.86} & \meansem{-29.19}{7.25} \\
 & DeepSeek   & \meansem{-36.83}{18.24}  & \meansem{-36.83}{12.79}& \meansem{-26.94}{6.20}  & \meansem{-20.83}{14.45} & \meansem{-20.83}{12.93} & \meansem{-17.31}{5.99} \\
\cmidrule(lr){2-8}
 & \textit{Mean} & \meansem{-11.47}{5.67}  & \meansem{-24.20}{8.70} & \meansem{-21.83}{3.61}  & \meansem{-8.49}{2.76}   & \meansem{-27.57}{9.82} & \meansem{-15.96}{3.84} \\
\midrule

\multirow{6}{*}{M.G}
 & Grok       & \meansem{1.00}{0.00}   & \meansem{1.00}{0.00}   & \meansem{0.96}{0.03}    & \meansem{1.00}{0.00}   & \meansem{1.00}{0.00} & \meansem{0.83}{0.10} \\
 & GPT        & \meansem{1.00}{0.00}   & \meansem{1.00}{0.00}   & \meansem{0.96}{0.03}    & \meansem{1.00}{0.00}   & \meansem{1.00}{0.00} & \meansem{0.83}{0.10} \\
 & Gemini     & \meansem{1.00}{0.00}   & \meansem{1.00}{0.00}   & \meansem{0.96}{0.03}    & \meansem{1.00}{0.00}   & \meansem{1.00}{0.00} & \meansem{0.83}{0.10} \\
 & Claude     & \meansem{1.00}{0.00}   & \meansem{1.00}{0.00}   & \meansem{0.96}{0.03}    & \meansem{1.00}{0.00}   & \meansem{1.00}{0.00} & \meansem{0.83}{0.10} \\
 & DeepSeek   & \meansem{1.00}{0.00}   & \meansem{0.67}{0.27}   & \meansem{0.83}{0.00}    & \meansem{1.00}{0.00}   & \meansem{0.01}{0.01} & \meansem{0.33}{0.00} \\
\cmidrule(lr){2-8}
 & \textit{Mean} & \meansem{1.00}{0.00}   & \meansem{0.93}{0.06}   & \meansem{0.93}{0.02}    & \meansem{1.00}{0.00}   & \meansem{0.80}{0.10} & \meansem{0.73}{0.07} \\
\midrule

\multirow{6}{*}{P.L}
 & Grok       & \meansem{5.11}{0.20}   & \meansem{3.00}{0.25}   & \meansem{4.29}{0.15}    & \meansem{5.30}{0.21}   & \meansem{4.67}{0.62} & \meansem{4.55}{0.24} \\
 & GPT        & \meansem{5.44}{1.01}   & \meansem{7.87}{0.44}   & \meansem{6.05}{0.46}    & \meansem{7.29}{0.49}   & \meansem{6.90}{0.17} & \meansem{6.09}{0.40} \\
 & Gemini     & \meansem{6.54}{0.10}   & \meansem{6.44}{0.03}   & \meansem{5.78}{0.31}    & \meansem{8.36}{0.06}   & \meansem{8.37}{0.07} & \meansem{6.40}{0.49} \\
 & Claude     & \meansem{2.27}{0.22}   & \meansem{9.18}{0.01}   & \meansem{6.20}{0.42}    & \meansem{3.57}{1.02}   & \meansem{6.56}{1.87} & \meansem{6.28}{0.51} \\
 & DeepSeek   & \meansem{3.53}{0.67}   & \meansem{4.00}{0.81}   & \meansem{5.15}{0.31}    & \meansem{4.33}{0.65}   & \meansem{5.65}{0.36} & \meansem{6.27}{0.20} \\
\cmidrule(lr){2-8}
 & \textit{Mean} & \meansem{4.58}{0.67}   & \meansem{6.10}{0.63}   & \meansem{5.50}{0.27}    & \meansem{5.77}{0.80}   & \meansem{6.43}{0.51} & \meansem{5.92}{0.29} \\
\bottomrule
\end{tabular}

\label{tab:code-xp-summary}
\end{table*}
\begin{table*}[ht]
\centering
\tiny
\caption{
\textbf{\acs{sp} and \acs{xp} failure rates of code agents across environments, under fixed- and variable-\acs{llm} settings, with and without CFD.} Agents tested: GPT-5.2, Grok~4.1~Fast, Claude~Opus~4.5, Gemini~3~Pro, and Deepseek V3.2 Terminus.}
\begin{tabular}{l l c c c c c c}
\toprule
\multicolumn{2}{c}{ } &
\multicolumn{3}{c}{w/o CFD} & \multicolumn{3}{c}{w/ CFD} \\
\cmidrule(lr){3-5} \cmidrule(lr){6-8}
Env & LLM & $\text{SP}$ & $\text{XP}_{\text{seed}}$ & $\text{XP}_{\text{LLM}}$ & $\text{SP}$ & $\text{XP}_{\text{seed}}$ & $\text{XP}_{\text{LLM}}$ \\
\midrule

\multirow{6}{*}{C.D.}
 & Grok       & \meansem{0.00}{0.00} & \meansem{0.33}{0.27} & \meansem{0.46}{0.09} & \meansem{0.00}{0.00} & \meansem{0.33}{0.27} & \meansem{0.02}{0.02} \\
 & GPT        & \meansem{0.00}{0.00} & \meansem{0.00}{0.00} & \meansem{0.63}{0.10} & \meansem{0.00}{0.00} & \meansem{0.00}{0.00} & \meansem{0.02}{0.02} \\
 & Gemini     & \meansem{0.00}{0.00} & \meansem{0.00}{0.00} & \meansem{0.57}{0.10} & \meansem{0.00}{0.00} & \meansem{0.00}{0.00} & \meansem{0.00}{0.00} \\
 & Claude     & \meansem{0.00}{0.00} & \meansem{0.83}{0.14} & \meansem{0.42}{0.03} & \meansem{0.00}{0.00} & \meansem{0.83}{0.14} & \meansem{0.00}{0.00} \\
 & DeepSeek   & \meansem{0.00}{0.00} & \meansem{0.67}{0.27} & \meansem{0.48}{0.05} & \meansem{0.00}{0.00} & \meansem{0.67}{0.27} & \meansem{0.00}{0.00} \\
\cmidrule(lr){2-8}
 & \textit{Mean} & \meansem{0.00}{0.00} & \meansem{0.37}{0.12} & \meansem{0.51}{0.05} & \meansem{0.00}{0.00} & \meansem{0.37}{0.12} & \meansem{0.01}{0.01} \\
\midrule

\multirow{6}{*}{A.C.}
 & Grok       & \meansem{0.00}{0.00} & \meansem{0.00}{0.00} & \meansem{0.46}{0.14} & \meansem{0.00}{0.00} & \meansem{0.00}{0.00} & \meansem{0.29}{0.07} \\
 & GPT        & \meansem{0.00}{0.00} & \meansem{0.00}{0.00} & \meansem{0.16}{0.04} & \meansem{0.00}{0.00} & \meansem{0.00}{0.00} & \meansem{0.29}{0.07} \\
 & Gemini     & \meansem{0.00}{0.00} & \meansem{0.00}{0.00} & \meansem{0.54}{0.13} & \meansem{0.00}{0.00} & \meansem{0.00}{0.00} & \meansem{0.26}{0.05} \\
 & Claude     & \meansem{0.00}{0.00} & \meansem{0.33}{0.27} & \meansem{0.39}{0.09} & \meansem{0.00}{0.00} & \meansem{0.33}{0.27} & \meansem{0.33}{0.08} \\
 & DeepSeek   & \meansem{0.00}{0.00} & \meansem{0.33}{0.27} & \meansem{0.39}{0.11} & \meansem{0.14}{0.13} & \meansem{0.33}{0.27} & \meansem{0.52}{0.04} \\
\cmidrule(lr){2-8}
 & \textit{Mean} & \meansem{0.00}{0.00} & \meansem{0.13}{0.09} & \meansem{0.39}{0.07} & \meansem{0.03}{0.03} & \meansem{0.13}{0.09} & \meansem{0.34}{0.05} \\
\midrule

\multirow{6}{*}{A.C.P.S.}
 & Grok       & \meansem{0.16}{0.14} & \meansem{0.60}{0.05} & \meansem{0.42}{0.05} & \meansem{0.00}{0.00} & \meansem{0.55}{0.06} & \meansem{0.47}{0.07} \\
 & GPT        & \meansem{0.08}{0.08} & \meansem{0.28}{0.17} & \meansem{0.20}{0.09} & \meansem{0.12}{0.11} & \meansem{0.28}{0.17} & \meansem{0.36}{0.06} \\
 & Gemini     & \meansem{0.00}{0.00} & \meansem{0.00}{0.00} & \meansem{0.17}{0.07} & \meansem{0.00}{0.00} & \meansem{0.00}{0.00} & \meansem{0.23}{0.08} \\
 & Claude     & \meansem{0.08}{0.08} & \meansem{0.08}{0.07} & \meansem{0.19}{0.07} & \meansem{0.14}{0.08} & \meansem{0.09}{0.07} & \meansem{0.30}{0.07} \\
 & DeepSeek   & \meansem{0.12}{0.07} & \meansem{0.62}{0.06} & \meansem{0.43}{0.07} & \meansem{0.33}{0.12} & \meansem{0.65}{0.04} & \meansem{0.45}{0.08} \\
\cmidrule(lr){2-8}
 & \textit{Mean} & \meansem{0.09}{0.02} & \meansem{0.32}{0.08} & \meansem{0.28}{0.05} & \meansem{0.12}{0.05} & \meansem{0.31}{0.08} & \meansem{0.36}{0.05} \\
\midrule

\multirow{6}{*}{T.H.}
 & Grok       & \meansem{0.00}{0.00} & \meansem{0.00}{0.00} & \meansem{0.24}{0.10} & \meansem{0.00}{0.00} & \meansem{0.00}{0.00} & \meansem{0.23}{0.05} \\
 & GPT        & \meansem{0.00}{0.00} & \meansem{0.08}{0.06} & \meansem{0.23}{0.08} & \meansem{0.00}{0.00} & \meansem{0.07}{0.06} & \meansem{0.27}{0.09} \\
 & Gemini     & \meansem{0.00}{0.00} & \meansem{0.00}{0.00} & \meansem{0.24}{0.10} & \meansem{0.00}{0.00} & \meansem{0.00}{0.00} & \meansem{0.25}{0.09} \\
 & Claude     & \meansem{0.00}{0.00} & \meansem{0.66}{0.06} & \meansem{0.34}{0.05} & \meansem{0.00}{0.00} & \meansem{0.65}{0.06} & \meansem{0.31}{0.06} \\
 & DeepSeek   & \meansem{0.05}{0.04} & \meansem{0.36}{0.18} & \meansem{0.50}{0.08} & \meansem{0.07}{0.07} & \meansem{0.34}{0.19} & \meansem{0.33}{0.10} \\
\cmidrule(lr){2-8}
 & \textit{Mean} & \meansem{0.01}{0.01} & \meansem{0.22}{0.08} & \meansem{0.31}{0.06} & \meansem{0.01}{0.01} & \meansem{0.21}{0.08} & \meansem{0.28}{0.05} \\
\midrule

\multirow{6}{*}{T.H.F}
 & Grok       & \meansem{0.00}{0.00} & \meansem{0.00}{0.00} & \meansem{0.02}{0.02} & \meansem{0.00}{0.00} & \meansem{0.00}{0.00} & \meansem{0.18}{0.10} \\
 & GPT        & \meansem{0.00}{0.00} & \meansem{0.00}{0.00} & \meansem{0.02}{0.02} & \meansem{0.00}{0.00} & \meansem{0.00}{0.00} & \meansem{0.21}{0.07} \\
 & Gemini     & \meansem{0.00}{0.00} & \meansem{0.00}{0.00} & \meansem{0.02}{0.02} & \meansem{0.00}{0.00} & \meansem{0.00}{0.00} & \meansem{0.00}{0.00} \\
 & Claude     & \meansem{0.00}{0.00} & \meansem{0.00}{0.00} & \meansem{0.02}{0.02} & \meansem{0.00}{0.00} & \meansem{0.00}{0.00} & \meansem{0.18}{0.10} \\
 & DeepSeek   & \meansem{0.00}{0.00} & \meansem{0.17}{0.14} & \meansem{0.08}{0.03} & \meansem{0.00}{0.00} & \meansem{0.17}{0.14} & \meansem{0.23}{0.11} \\
\cmidrule(lr){2-8}
 & \textit{Mean} & \meansem{0.00}{0.00} & \meansem{0.03}{0.03} & \meansem{0.03}{0.01} & \meansem{0.00}{0.00} & \meansem{0.03}{0.03} & \meansem{0.16}{0.06} \\
\midrule

\multirow{6}{*}{B.B}
 & Grok       & \meansem{0.00}{0.00} & \meansem{0.00}{0.00} & \meansem{0.19}{0.08} & \meansem{0.01}{0.01} & \meansem{0.00}{0.00} & \meansem{0.15}{0.05} \\
 & GPT        & \meansem{0.01}{0.00} & \meansem{0.01}{0.01} & \meansem{0.08}{0.03} & \meansem{0.04}{0.02} & \meansem{0.02}{0.02} & \meansem{0.04}{0.01} \\
 & Gemini     & \meansem{0.00}{0.00} & \meansem{0.00}{0.00} & \meansem{0.13}{0.07} & \meansem{0.02}{0.00} & \meansem{0.00}{0.00} & \meansem{0.04}{0.01} \\
 & Claude     & \meansem{0.62}{0.14} & \meansem{0.00}{0.00} & \meansem{0.18}{0.03} & \meansem{0.31}{0.14} & \meansem{0.00}{0.00} & \meansem{0.16}{0.06} \\
 & DeepSeek   & \meansem{0.58}{0.18} & \meansem{0.49}{0.24} & \meansem{0.34}{0.06} & \meansem{1.00}{0.00} & \meansem{0.50}{0.24} & \meansem{0.20}{0.08} \\
\cmidrule(lr){2-8}
 & \textit{Mean} & \meansem{0.24}{0.13} & \meansem{0.10}{0.07} & \meansem{0.18}{0.04} & \meansem{0.28}{0.17} & \meansem{0.10}{0.07} & \meansem{0.12}{0.03} \\
\midrule

\multirow{6}{*}{N.T}
 & Grok       & \meansem{0.00}{0.00} & \meansem{0.00}{0.00} & \meansem{0.12}{0.06} & \meansem{0.00}{0.00} & \meansem{0.00}{0.00} & \meansem{0.15}{0.07} \\
 & GPT        & \meansem{0.00}{0.00} & \meansem{0.00}{0.00} & \meansem{0.08}{0.04} & \meansem{0.00}{0.00} & \meansem{0.00}{0.00} & \meansem{0.02}{0.02} \\
 & Gemini     & \meansem{0.00}{0.00} & \meansem{0.00}{0.00} & \meansem{0.12}{0.06} & \meansem{0.00}{0.00} & \meansem{0.00}{0.00} & \meansem{0.00}{0.00} \\
 & Claude     & \meansem{0.00}{0.00} & \meansem{0.67}{0.27} & \meansem{0.31}{0.04} & \meansem{0.00}{0.00} & \meansem{0.67}{0.27} & \meansem{0.25}{0.08} \\
 & DeepSeek   & \meansem{0.33}{0.19} & \meansem{0.33}{0.14} & \meansem{0.23}{0.07} & \meansem{0.17}{0.15} & \meansem{0.33}{0.14} & \meansem{0.13}{0.06} \\
\cmidrule(lr){2-8}
 & \textit{Mean} & \meansem{0.07}{0.06} & \meansem{0.20}{0.09} & \meansem{0.17}{0.04} & \meansem{0.03}{0.03} & \meansem{0.20}{0.09} & \meansem{0.11}{0.04} \\
\midrule

\multirow{6}{*}{M.G}
 & Grok       & \meansem{0.00}{0.00} & \meansem{0.00}{0.00} & \meansem{0.04}{0.03} & \meansem{0.00}{0.00} & \meansem{0.00}{0.00} & \meansem{0.17}{0.10} \\
 & GPT        & \meansem{0.00}{0.00} & \meansem{0.00}{0.00} & \meansem{0.04}{0.03} & \meansem{0.00}{0.00} & \meansem{0.00}{0.00} & \meansem{0.17}{0.10} \\
 & Gemini     & \meansem{0.00}{0.00} & \meansem{0.00}{0.00} & \meansem{0.04}{0.03} & \meansem{0.00}{0.00} & \meansem{0.00}{0.00} & \meansem{0.17}{0.10} \\
 & Claude     & \meansem{0.00}{0.00} & \meansem{0.00}{0.00} & \meansem{0.04}{0.03} & \meansem{0.00}{0.00} & \meansem{0.00}{0.00} & \meansem{0.17}{0.10} \\
 & DeepSeek   & \meansem{0.00}{0.00} & \meansem{0.33}{0.27} & \meansem{0.17}{0.00} & \meansem{0.00}{0.00} & \meansem{0.33}{0.27} & \meansem{0.67}{0.00} \\
\cmidrule(lr){2-8}
 & \textit{Mean} & \meansem{0.00}{0.00} & \meansem{0.07}{0.06} & \meansem{0.07}{0.02} & \meansem{0.00}{0.00} & \meansem{0.07}{0.06} & \meansem{0.27}{0.07} \\
\midrule

\multirow{6}{*}{P.L}
 & Grok       & \meansem{0.20}{0.07} & \meansem{0.02}{0.01} & \meansem{0.15}{0.02} & \meansem{0.31}{0.05} & \meansem{0.02}{0.02} & \meansem{0.21}{0.04} \\
 & GPT        & \meansem{0.30}{0.15} & \meansem{0.02}{0.01} & \meansem{0.11}{0.03} & \meansem{0.07}{0.03} & \meansem{0.01}{0.01} & \meansem{0.13}{0.03} \\
 & Gemini     & \meansem{0.12}{0.01} & \meansem{0.09}{0.01} & \meansem{0.10}{0.00} & \meansem{0.00}{0.00} & \meansem{0.11}{0.01} & \meansem{0.12}{0.03} \\
 & Claude     & \meansem{0.79}{0.02} & \meansem{0.00}{0.00} & \meansem{0.11}{0.03} & \meansem{0.61}{0.13} & \meansem{0.00}{0.00} & \meansem{0.15}{0.04} \\
 & DeepSeek   & \meansem{0.49}{0.08} & \meansem{0.37}{0.18} & \meansem{0.15}{0.01} & \meansem{0.39}{0.09} & \meansem{0.40}{0.19} & \meansem{0.06}{0.02} \\
\cmidrule(lr){2-8}
 & \textit{Mean} & \meansem{0.38}{0.11} & \meansem{0.10}{0.05} & \meansem{0.12}{0.01} & \meansem{0.28}{0.10} & \meansem{0.11}{0.05} & \meansem{0.14}{0.02} \\
\bottomrule
\end{tabular}

\label{tab:xp-summary-failure}
\end{table*}

\subsection{Hanabi 3x3 variant}

\begin{figure*}
    \centering
    \includegraphics[width=\linewidth]{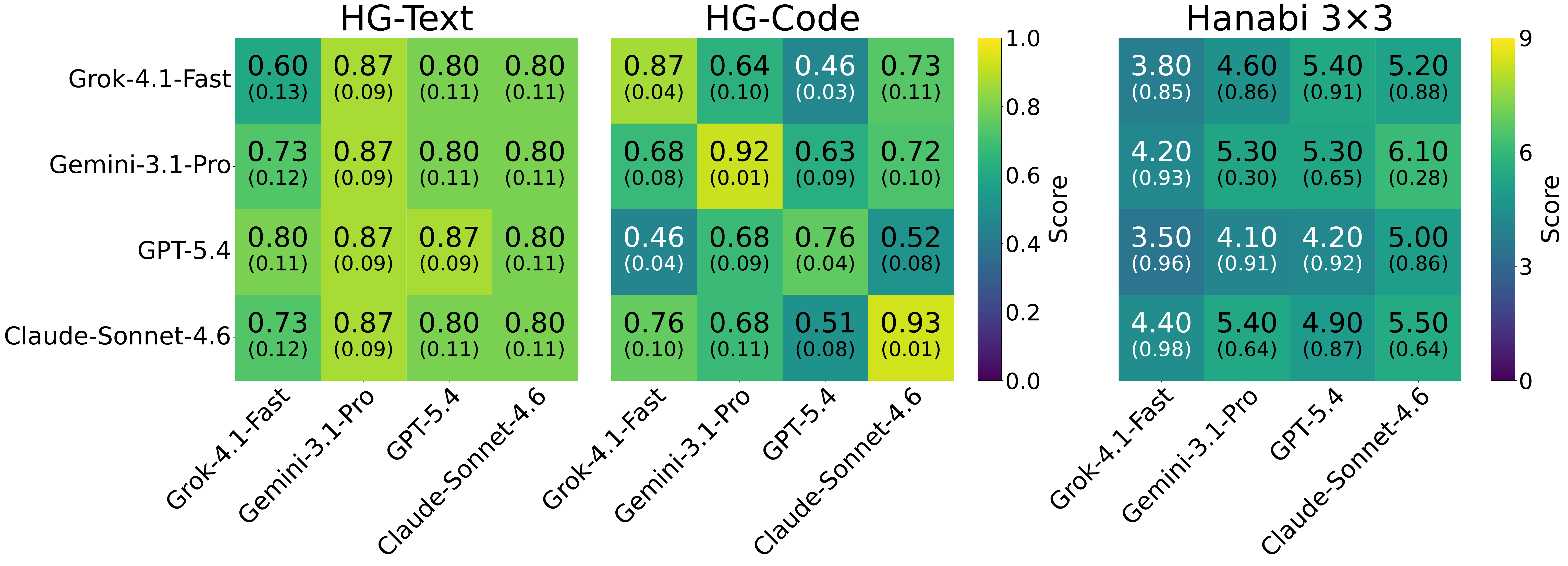}
    \caption{\textbf{XP matrices of \acs{llm} text and code agents.} (Left and Middle) XP matrix in the Hint-Guess game; text agents coordinate well, whereas code agents do not. Human players score $0.75$ on average~\citep{ma2023learningintuitivepoliciesusing}. (Right) XP matrix of text agents on the 3$\times$3 Hanabi variant; text agents fall well short of the $7.7$ achieved by IPPO with high entropy regularization~\cite{forkel2025entropy}. For text agents, each cell reports the mean and s.e.m.\ over $10$ episodes; for code agents, each cell is averaged over $100$ episodes across $6$ independent \acs{xp} matrices.}
    \label{fig:hanabi_xp}
\end{figure*}

\begin{figure*}[t]
    \centering
    \includegraphics[width=\linewidth]{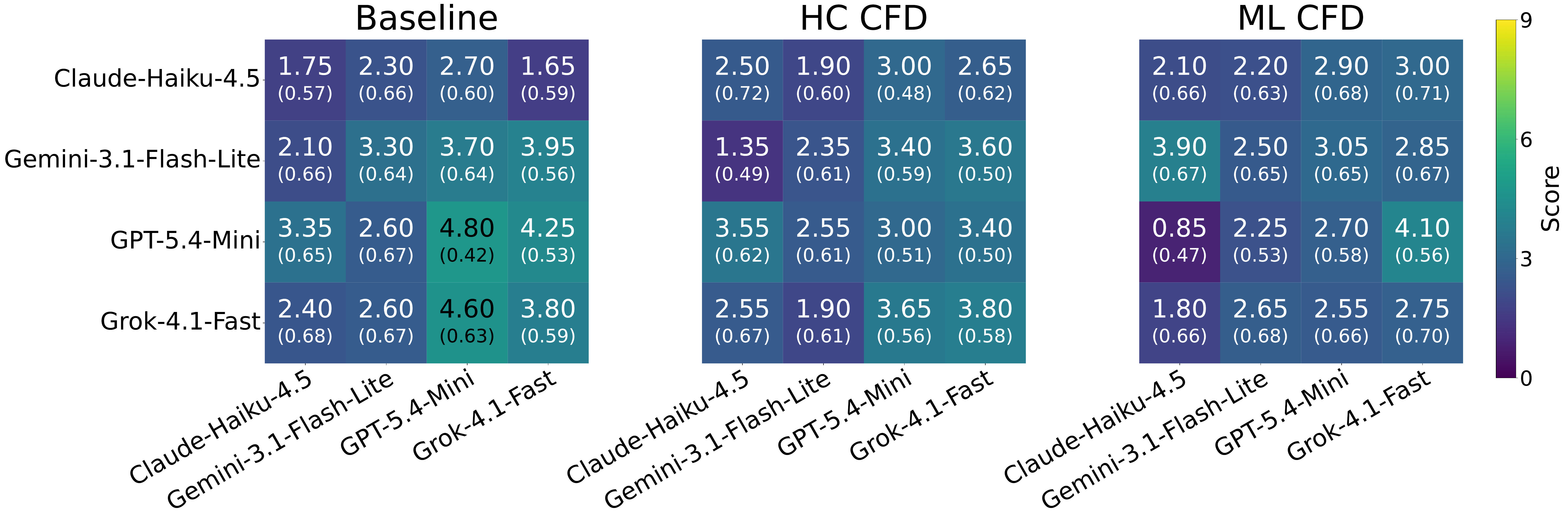}
    \caption{\textbf{Hanabi 3$\times$3 results of text agents with and without a CFD.} We find that the LLM agents achieve low scores in all $3$ cases.}
    \label{fig:hanabi_n20_conditions}
\end{figure*}

From \autoref{fig:hanabi_n20_conditions}, we observe that smaller models perform worse on Hanabi 3×3, and that neither CFD leads to score improvements for text agents.

\subsection{Hint-Guess Implicatures}
\begin{table*}[t]
\centering
\scriptsize
\setlength{\tabcolsep}{24pt}
\caption{\textbf{Hint-Guess Implicatures: Human-compatible hint rate and win rate (\%) by scenario.}
Scenarios correspond exactly to those in \autoref{fig:hint_guess_scenarios}.
Each cell shows Human\,\%\,/\,Win\,\% ($n{=}12$ games per hinter--scenario cell).
\textit{Exact}: target card available in hinter's hand.
\textit{Similarity (S)}: feature similarity required, no exact match.
\textit{Exclusivity (E)}: mutual exclusivity (Gricean implicature) required.
\textit{S+E}: combined similarity + exclusivity scenario.
All models give the human-compatible hint perfectly on Exact, Sim, and Excl; failures concentrate in the S+E condition.}
\begin{tabular}{lcccc}
\toprule
Hinter & Exact Match & Similarity (S) & Exclusivity (E) & S+E \\
\midrule
Claude Sonnet 4.6       & 100 / 100 & 100 / 100 & 100 / 100 & 100 / 100 \\
Gemini 3.1 Pro          & 100 / 100 & 100 / 100 & 100 / 100 & 100 / \phantom{0}92 \\
GPT-5.4                 & 100 / 100 & 100 / 100 & 100 / 100 & 100 / 100 \\
Grok 4.1 Fast           & 100 / 100 & 100 / 100 & 100 / 100 & \phantom{0}75 / \phantom{0}75 \\
\midrule
All                     & 100 / 100 & 100 / 100 & 100 / 100 & \phantom{0}94 / \phantom{0}92 \\
\bottomrule
\end{tabular}
\label{tab:hint_guess_implicatures}
\end{table*}

We evaluate whether text agents can match human actions, which requires human-like reasoning patterns such as implicatures \cite{implicature1,implicature2}. To this end, we pair each LLM (Grok-4.1-Fast, Gemini-3.1-Pro, GPT-5.4, and Claude-Sonnet-4.6) with every other model, sampling each game type in \autoref{fig:hint_guess_scenarios} three times per pairing. The results, reported in \autoref{tab:hint_guess_implicatures}, show that LLM agents are able to align with human-like reasoning, achieving performance comparable to state-of-the-art methods \cite{ma2023learningintuitivepoliciesusing} (Table 2).

\subsection{Automatic discovery of CFDs}
\begin{figure}
    \centering
    \includegraphics[width=\linewidth]{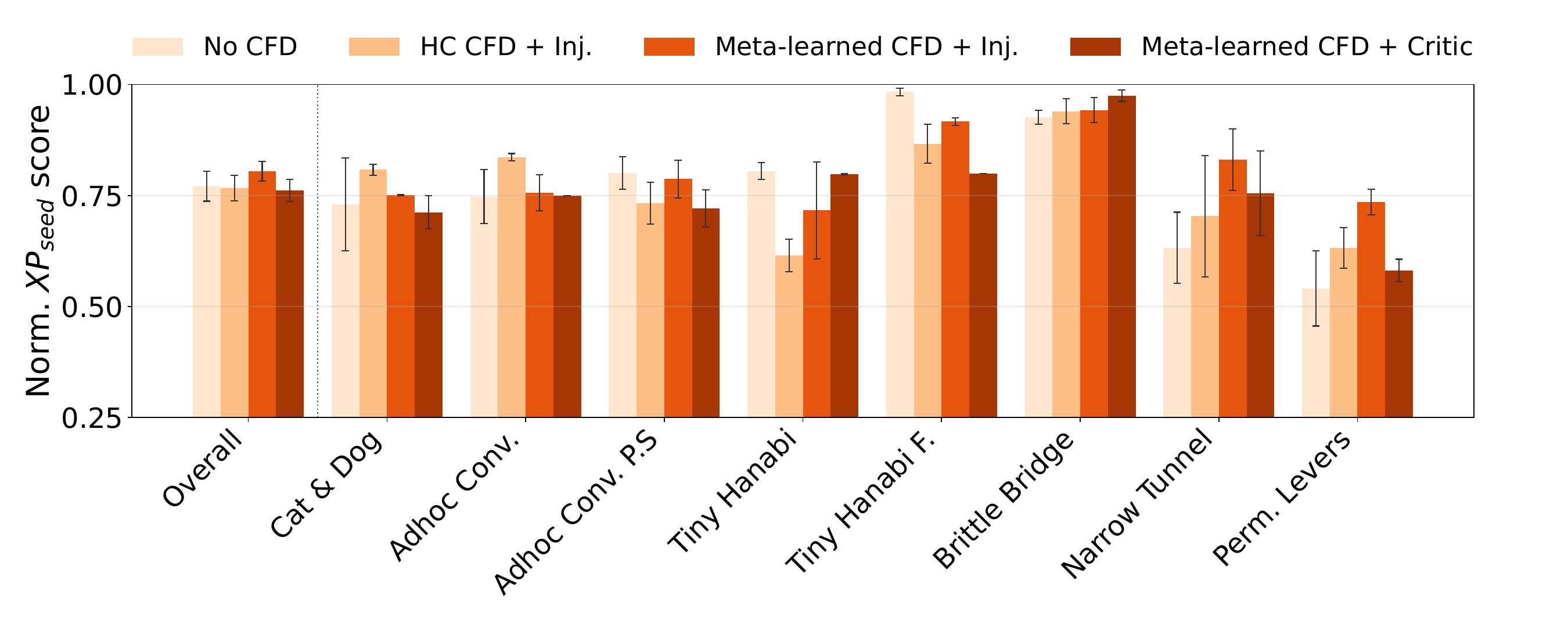}
    \caption{\textbf{$\text{XP}_{\text{seed}}$ scores of code agents aggregated across different LLM types and scaffolds.} Results are shown for Grok~4.1~Fast, GPT-5.2, Gemini~3~Pro, and Opus~4.5. Since the \acs{cfd} was meta-trained with Grok~4.1~Fast on Cat \& Dog and Brittle Bridge, Grok~4.1~Fast is excluded from these two environments in all conditions, including the overall score. Scores are min--max normalised per environment using fixed reward ranges, and error bars denote the standard error over seed pairs.}
    \label{fig:code_xp_seed_cfd}
\end{figure}
\begin{figure}
    \centering
    \includegraphics[width=\linewidth]{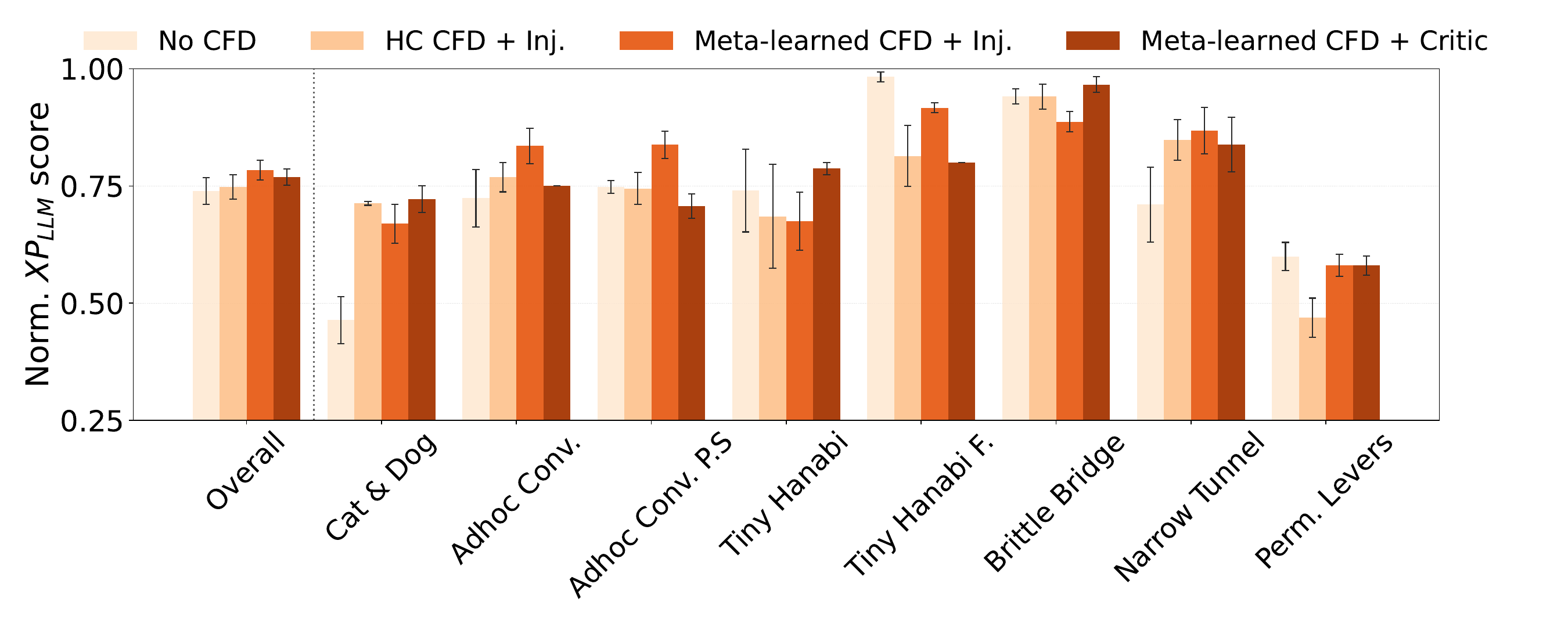}
    \caption{\textbf{$\text{XP}_{\text{LLM}}$ scores of code agents aggregated across different LLM types and scaffolds.} Results are shown for Grok~4.1~Fast, GPT-5.2, Gemini~3~Pro, and Opus~4.5.}
    \label{fig:code_xp_llm}
\end{figure}
We report the cross-LLM and scaffold generalization scores of the meta-learned CFD in \autoref{fig:code_xp_seed_cfd} and~\autoref{fig:code_xp_llm}.

\clearpage
\section{Implementation details}
\label{sec:app/implementation_details}

\subsection{Meta-Learning Loop}
\label{sec:app/implementation_details/meta_learning}

We outline the algorithm used in our pareto-evolution meta-learning loop in Algorithms~\ref{alg:meta_loop/score}, \ref{alg:meta_loop/pareto-frontier}, and~\ref{alg:meta_learning/pareto-evolution}. The algorithm is roughly based on GEPA~\citep{agrawal2026gepareflectivepromptevolution}, implementing both pareto-based candidate selection and the cheap pre-filtering of new CFDs. Still, there are notable differences to GEPA. In particular, GEPA cannot be applied directly to the \acs{zsc} setting as the number of environments is generally far smaller than the number of ``tasks'' assumed in GEPA. Compared to GEPA, we therefore compute the actual pareto frontier in Algorithm~\ref{alg:meta_loop/pareto-frontier} across our meta-train set, without any previous task-wise elite filtering. We also adopt a caching mechanism for the set of initial policies used in each scoring step. This significantly reduces the costs of running the meta-loop. Additionally, by using the same set of initial policies when comparing \acsp{cfd}, we significantly reduce the variance in \acs{xp}-scores (as we eliminate the source of variance from the initial generation step), which generally leads to a more stable meta-loop.
\\
At last, the meta-loop differs in the used mutation operators. Similar to GEPA, we tried to include environment feedback into the mutation operators used in the meta-loop, both through direct or obfuscated conditioning on observed coordination failures. In general, we found that the inclusion of such operators unhelpful, often leading to either overfitting or low-quality uni-modal proposals. Instead, we found that ensuring the creation of diverse proposals in the mutation operators was of key importance to the success of the meta-loop. To facilitate such proposals, we apply the following modifications: Mutation operators generally prompt an \acs{llm} for a specific kind of modification of the \acs{cfd} instead of leaving the modification to the \acs{llm}. This is similar to the mutation prompts used in PromptBreeder~\cite{fernando2023promptbreederselfreferentialselfimprovementprompt}. In our case, the \acs{llm} is prompted to either add, remove,  reformulate, or freely modify a bullet point in the \acs{cfd}. Orthogonal to this, we found that prompting an \acs{llm} for a set of diverse proposals and then sampling the final proposal uniformly at random from this set can significantly increase diversity, as it allows us to sample proposals from the long tail, which would normally not make it into the \acs{llm}'s first response. These two axes yield a total of eight mutation operators. We also add the re-initialization of \acsp{cfd} as an additional mutation operator. 
\paragraph{Hyperparameter choices.} We outline in this paragraph the hyperparameter choices used in~\autoref{alg:meta_learning/pareto-evolution} to obtain the \acs{cfd} described in~\autoref{sec:cfd}.
The distribution over the mutation operators $\nu$ assigns each a sample weight of $1$, except the re-initialization operator, which receives a weight of $2$. To increase the diversity of proposals and generated policies, the meta-optimisation agent and the code agents generating policies during meta-training use a temperature of $1.0$, while the policy critic uses a temperature of $0$. The remaining hyperparameter choices are outlined in \autoref{tab:meta_train:hyperparams}.

\begin{table}[H]
\centering
\small
\begin{tabular}{@{}llp{0.55\linewidth}@{}}
\toprule
Symbol & Value & Quick explanation \\
\midrule
$\mathcal{E}$        & $\{\textsc{CatAndDog},\ \textsc{BrittleBridge}\}$ & Meta-train coordination environments \\
$K$                  & $5$    & Number of seed \acsp{cfd} sampled at initialisation \\
$T_{\mathrm{evo}}$   & $20$   & Outer-loop evolution steps \\
$n_c$                & $10$   & Size of the \acs{xp}-matrix in the cheap single-env gate \\
$n_f$                & $15$   & Size of the \acs{xp}-matrix in the full all-envs evaluation \\
$L$                  & $3$    & Max retries to draw a non-duplicate \acs{cfd} \\
$T_{\mathrm{crit}}$  & $4$    & Critic iteration budget per starting policy \\
\bottomrule
\end{tabular}
\caption{Inputs to~\autoref{alg:meta_learning/pareto-evolution} used in our experiments.}
\label{tab:meta_train:hyperparams}
\end{table}

\clearpage

\begin{algorithm}[H]
\caption{\textsc{Score}: evaluate a CFD on one environment via $n \times n$ cross-play}
\label{alg:meta_loop/score}
\begin{algorithmic}[1]
\Require CFD $c$; Dec-POMDP $\mathcal{M}$ with natural-text descriptions $(\mathcal{D}_{\mathcal{M}}, \mathcal{O}_{\mathcal{M}}, \mathcal{A}_{\mathcal{M}})$ of the game, observations, and actions; preset starting policies $\Pi = (\pi^{(1)}_0, \ldots, \pi^{(n)}_0)$, one per seed; critic iteration budget $T_{\mathrm{crit}}$
\Ensure mean off-diagonal cross-play score $\sigma \in \mathbb{R}$
\Procedure{Score}{$c, \mathcal{M}, \Pi, T_{\mathrm{crit}}$}
    \State $n \gets |\Pi|$
    \For{$i = 1$ \textbf{to} $n$}
        \State $(\pi^{(i)},\ \cdot) \gets \textsc{Refine}(\pi^{(i)}_0,\ \mathcal{M},\ c,\ T_{\mathrm{crit}})$ \Comment{see \autoref{alg:policy_critic/full}}
    \EndFor
    \State $X \in \mathbb{R}^{n \times n}$ with $X[i,j] = \mathbb{E}[J(\pi^{(i)}, \pi^{(j)})]$ \Comment{cross-play return matrix}
    \State \Return $\dfrac{1}{n^2 - n} \sum_{i \neq j} X[i,j]$
\EndProcedure
\end{algorithmic}
\end{algorithm}

\begin{algorithm}[H]
\caption{\textsc{ParetoFrontier}: non-dominated subset of a candidate set}
\label{alg:meta_loop/pareto-frontier}
\begin{algorithmic}[1]
\Statex \textbf{Definition (Pareto dominance).} Given candidates $c, c'$ with per-environment scores $\sigma_c, \sigma_{c'} \in \mathbb{R}^{|\mathcal{E}|}$, $c'$ \emph{dominates} $c$ iff $\sigma_{c'}(\mathcal{M}) \geq \sigma_c(\mathcal{M})$ for all $\mathcal{M} \in \mathcal{E}$ and $\sigma_{c'}(\mathcal{M}) > \sigma_c(\mathcal{M})$ for at least one $\mathcal{M} \in \mathcal{E}$.
\Statex
\Require candidate set $P$; environment set $\mathcal{E}$
\Ensure non-dominated subset $F \subseteq P$
\Procedure{ParetoFrontier}{$P, \mathcal{E}$}
    \State $F \gets \varnothing$
    \For{$c \in P$}
        \If{no $c' \in P \setminus \{c\}$ dominates $c$}
            \State $F \gets F \cup \{c\}$
        \EndIf
    \EndFor
    \State \Return $F$
\EndProcedure
\end{algorithmic}
\end{algorithm}

\begin{algorithm}[H]
\caption{\textsc{ParetoEvolution}: meta-learning over CFDs across environments}
\label{alg:meta_learning/pareto-evolution}
\begin{algorithmic}[1]
\Require environment set $\mathcal{E}$ (Dec-POMDPs); seed population size $K$; evolution steps $T_{\mathrm{evo}}$; cheap-eval seeds $n_c$; full-eval seeds $n_f$; mutation operator distribution $\nu$; max mutation retries $L$; critic iteration budget $T_{\mathrm{crit}}$
\Ensure selected CFD $c^\star$
\Procedure{ParetoEvolution}{$\mathcal{E}, K, T_{\mathrm{evo}}, n_c, n_f, \nu, L, T_{\mathrm{crit}}$}
    \Statex \textbf{Phase 1: Initialization.}
    \State pre-sample fixed starting policies $\hat{\Pi}_{\mathcal{M}}$ for each $\mathcal{M} \in \mathcal{E}$ \Comment{$|\hat{\Pi}_{\mathcal{M}}| = n_f$ (one per seed)}
    \State sample $K$ diverse seed CFDs $\{c_1, \ldots, c_K\}$
    \State $P \gets \varnothing$
    \For{$c \in \{c_1, \ldots, c_K\}$}
        \State $\sigma_c(\mathcal{M}) \gets \textsc{Score}(c, \mathcal{M}, \hat{\Pi}_{\mathcal{M}}, T_{\mathrm{crit}})$ for all $\mathcal{M} \in \mathcal{E}$
        \State $P \gets P \cup \{c\}$
    \EndFor
    \State $P_{\mathrm{alive}} \gets P$ \Comment{all seeds initially alive}

    \Statex
    \Statex \textbf{Phase 2: Evolution.}
    \For{$t = 1$ \textbf{to} $T_{\mathrm{evo}}$}
        \State $F \gets \textsc{ParetoFrontier}(P_{\mathrm{alive}}, \mathcal{E})$
        \State $c_p \gets$ uniform sample from $F$ \Comment{parent}

        \Statex \quad \textit{Mutation with duplicate retry:}
        \State $c_x \gets \mathtt{nil}$
        \For{$\ell = 1$ \textbf{to} $L$}
            \State $m \sim \nu$;\ \ $c \gets m(c_p)$
            \If{$c \notin \{\text{CFDs in } P\}$}
                \State $c_x \gets c$
                \State \textbf{break}
            \EndIf
        \EndFor
        \If{$c_x = \mathtt{nil}$}
            \State \textbf{continue} \Comment{no novel child after $L$ tries}
        \EndIf

        \Statex \quad \textit{Cheap gate: same fresh starting policies for child and parent.}
        \State $\mathcal{M} \gets$ uniform sample from $\mathcal{E}$
        \State $\tilde{\Pi} \gets$ fresh sample of $n_c$ starting policies for $\mathcal{M}$
        \If{$\textsc{Score}(c_x, \mathcal{M}, \tilde{\Pi}, T_{\mathrm{crit}}) \leq \textsc{Score}(c_p, \mathcal{M}, \tilde{\Pi}, T_{\mathrm{crit}})$}
            \State \textbf{continue} \Comment{discard child}
        \EndIf

        \Statex \quad \textit{Full evaluation on all environments:}
        \State $\sigma_{c_x}(\mathcal{M}) \gets \textsc{Score}(c_x, \mathcal{M}, \hat{\Pi}_{\mathcal{M}}, T_{\mathrm{crit}})$ for all $\mathcal{M} \in \mathcal{E}$

        \Statex \quad \textit{Frontier admission:}
        \State $F' \gets \textsc{ParetoFrontier}(P_{\mathrm{alive}} \cup \{c_x\}, \mathcal{E})$
        \If{$c_x \in F'$}
            \State $P \gets P \cup \{c_x\}$
            \State $P_{\mathrm{alive}} \gets F'$ \Comment{dominated candidates removed}
        \EndIf
    \EndFor

    \Statex
    \Statex \textbf{Phase 3: Final Selection.}
    \Statex \quad \textit{From the final frontier, return the candidate with the highest mean score after per-environment min--max normalisation. This balances performance across environments without privileging those with larger absolute score ranges.}
    \State $F^\star \gets \textsc{ParetoFrontier}(P_{\mathrm{alive}}, \mathcal{E})$
    \For{$\mathcal{M} \in \mathcal{E}$}
        \State $\mathit{lo}_{\mathcal{M}} \gets \min_{c \in F^\star} \sigma_c(\mathcal{M})$;\quad $\mathit{hi}_{\mathcal{M}} \gets \max_{c \in F^\star} \sigma_c(\mathcal{M})$
    \EndFor
    \State \Return $\displaystyle \arg\max_{c \in F^\star} \frac{1}{|\mathcal{E}|} \sum_{\mathcal{M} \in \mathcal{E}} \frac{\sigma_c(\mathcal{M}) - \mathit{lo}_{\mathcal{M}}}{\mathit{hi}_{\mathcal{M}} - \mathit{lo}_{\mathcal{M}}}$
\EndProcedure
\end{algorithmic}
\end{algorithm}

\subsection{Policy Critic}
\label{sec:app/implementation_details/policy_critic}

We outline in \autoref{alg:policy_critic/one_step} and \autoref{alg:policy_critic/full} how the policy critic can be used to iteratively refine starting policies $\pi_0$. Starting policies are generated using the code-agent generation process as outlined in \autoref{fig:text_and_code_agents}, but without conditioning on any \acs{cfd}. We provide the prompt used for the policy critic in \autoref{fig:prompt_critic}. Separating generation from verfication can increase diversity \citep{ahmed2025intentfactoredgenerationunleashing}, as well as reduce the occurrence of false positives in the policy creation pipeline.

\begin{algorithm}[H]
\caption{\textsc{CriticStep}: single critic pass over a policy}
\label{alg:policy_critic/one_step}
\begin{algorithmic}[1]
\Require candidate policy $\pi$; Dec-POMDP $\mathcal{M}$ with natural-text descriptions $(\mathcal{D}_{\mathcal{M}}, \mathcal{O}_{\mathcal{M}}, \mathcal{A}_{\mathcal{M}})$ of the game, observations, and actions; coordination-friendly definition $c$; critic language model $\mathrm{LLM}$
\Ensure verdict $cf \in \{\text{true}, \text{false}\}$; revised policy $\pi'$ (with $\pi' = \pi$ when $cf = \text{true}$)
\Procedure{CriticStep}{$\pi, \mathcal{M}, c$}
    \State $p \gets \textsc{BuildCriticPrompt}(\pi, \mathcal{M}, c)$
    \State $\mathrm{resp} \gets \mathrm{LLM}(p,\ \text{temperature} = 0)$
    \State parse $\mathrm{resp}$ as JSON with fields \texttt{coordination\_friendly}, \texttt{modified\_policy}
    \If{$\mathrm{resp}.\texttt{coordination\_friendly}$}
        \State $\pi' \gets \pi$
    \Else
        \State $\pi' \gets \textsc{Parse}(\mathrm{resp}.\texttt{modified\_policy})$
    \EndIf
    \State \Return $(\mathrm{resp}.\texttt{coordination\_friendly},\ \pi')$
\EndProcedure
\end{algorithmic}
\end{algorithm}

\begin{algorithm}[H]
\caption{\textsc{Refine}: iterative policy refinement via the critic}
\label{alg:policy_critic/full}
\begin{algorithmic}[1]
\Require starting policy $\pi_0$; Dec-POMDP $\mathcal{M}$; coordination-friendly definition $c$; iteration budget $T$
\Ensure final policy $\pi^\star$; flag $\mathit{approved}$ indicating whether $\pi^\star$ was confirmed coordination-friendly
\Procedure{Refine}{$\pi_0, \mathcal{M}, c, T$}
    \State $\pi \gets \pi_0$
    \For{$t = 1$ \textbf{to} $T$}
        \State $(cf,\ \pi') \gets \textsc{CriticStep}(\pi, \mathcal{M}, c)$
        \If{$cf$}
            \State \Return $(\pi,\ \text{true})$
        \EndIf
        \State $\pi \gets \pi'$ \Comment{adopt the proposed revision}
    \EndFor
    \State \Return $(\pi,\ \text{false})$ \Comment{budget exhausted; return last revision}
\EndProcedure
\end{algorithmic}
\end{algorithm}

\clearpage

\subsection{Prompts}
\label{app:prompt}
The prompt template for text agents is shown in \autoref{fig:prompt_text}, for code agents in \autoref{fig:prompt_code}, and for the policy critic in \autoref{fig:prompt_critic}. To ensure consistent policies and reasonable actions, we use a temperature of $0$ for text and code agents unless otherwise specified. We note that even at these settings, most frontier models are non-deterministic~\citep{he_nondeterminism}.

\begin{figure*}[!ht]
\centering
\begin{tcolorbox}[
  colback=white,
  colframe=black!60,
  sharp corners,
  boxrule=0.8pt,
  width=\textwidth,
  title={Prompt Template — Text Agents},
  fonttitle=\bfseries\small,
  colbacktitle=gray!15,
  coltitle=black,
  before skip=6pt,
  after skip=6pt,
  enhanced,
]
\scriptsize\setlength{\parindent}{0pt}\raggedright

You are playing a cooperative game with another player. The game is generally turn-based. While you cannot communicate with the other player, you can observe their actions and the joint rewards from previous turns. Your objective is to maximise the joint reward.

\smallskip

You will play only a single round of the game (a round may consist of multiple turns).

\smallskip

Below is a detailed description of the game and, if applicable, the history of previous turns.

\smallskip
\colorbox{blue!8}{\strut\texttt{\$\,Game Description\,\$}} \quad
\colorbox{blue!8}{\strut\texttt{\$\,Game History\,\$}}

\medskip\noindent
You must now select an action for the current turn from the following set of legal actions:

\smallskip
\colorbox{blue!8}{\strut\texttt{\$\,Legal Actions\,\$}} \quad
\colorbox{blue!8}{\strut\texttt{\$\,CFD Instructions (if applicable)\,\$}}

\medskip\noindent
Respond using the following format:

\smallskip\noindent
\texttt{REASONING:}\\[1pt]
\texttt{<REASONING>}

\smallskip\noindent
\texttt{ACTION JSON:}
\begin{verbatim}
```json
<JSON>
```
\end{verbatim}

\end{tcolorbox}
\caption{Template of the prompt used for text agents. Blue-shaded fields are filled in at runtime.}
\label{fig:prompt_text}
\end{figure*}

\begin{figure*}[!ht]
\centering
\begin{tcolorbox}[
  colback=white,
  colframe=black!60,
  sharp corners,
  boxrule=0.8pt,
  width=\textwidth,
  title={Prompt Template — Code Agents},
  fonttitle=\bfseries\small,
  colbacktitle=gray!15,
  coltitle=black,
  before skip=6pt,
  after skip=6pt,
  enhanced,
]
\scriptsize\setlength{\parindent}{0pt}\raggedright

\textbf{General Instructions}

\smallskip
You will need to generate policies for a collaborative two-player game.
The game is generally turn-based. Agents cannot communicate with each other.
Agents can see their actions and the (joint) reward from previous turns.
Generally, agents play each game only once; if the game is iterated this will be stated explicitly.
\textbf{Ensure the policy is coordination-friendly.}

\tcbline

\textbf{Zero-Shot Coordination (ZSC) Setting}

\smallskip
The policy will be evaluated against independently generated policies (potentially from different LLMs) with similar instructions.
Poor coordination often arises when policies adopt incompatible conventions or break symmetry arbitrarily.
\textbf{Ultimate goal:} maximise coordination in the ZSC setting.

\smallskip
\colorbox{blue!8}{\strut\texttt{\$\,CFD Instructions (if applicable)\,\$}}

\tcbline

\textbf{Policy Specification}

\smallskip
Each player has a local policy: an \texttt{act} function with the following signature:

\smallskip
\texttt{def act(observation\_dict: dict, memory: dict) -> tuple[str, dict]}

\smallskip
The first return value is the action (string); the second is the memory dictionary (initialised to \texttt{\{\}} at game start, not shared between players). You may use Python standard libraries \texttt{random} and \texttt{math}.

\smallskip
\textbf{Observation dict} — index values using the \emph{exact} keys shown below (case-sensitive):

\smallskip
\colorbox{blue!8}{\strut\texttt{\$\,Sample Observation Dict\,\$}}

\smallskip
\textbf{Legal actions} — the returned action must exactly match one of:

\smallskip
\colorbox{blue!8}{\strut\texttt{\$\,Legal Actions Dict\,\$}}

\smallskip
Policies that return an invalid action end the game immediately with a penalty.

\smallskip
\textbf{Game description:}

\smallskip
\colorbox{blue!8}{\strut\texttt{\$\,Game Description\,\$}}

\tcbline

\textbf{Output Specification}

\smallskip
Propose 1 joint policy (one \texttt{act} function per player). Format the output as JSON under the key \texttt{"joint\_policies"}:

\smallskip
\begin{minipage}{\linewidth}
\ttfamily\small
\{\\
\quad "joint\_policies": [\{\\
\qquad "name":   \textit{<name shared by all players>},\\
\qquad "reason": \textit{<why this design>},\\
\qquad "player\_0": \{ "code": \textit{<escaped Python string for \texttt{act}>} \},\\
\qquad "player\_1": \{ "code": \textit{<escaped Python string for \texttt{act}>} \}\\
\quad \}]\\
\}
\end{minipage}

\medskip
\texttt{REASONING:}\\[1pt]
\texttt{<REASONING>}

\smallskip
\texttt{Policy JSON:}
\begin{verbatim}
```json
<JSON>
```
\end{verbatim}

\end{tcolorbox}
\caption{Template of the prompt used for code agents. Blue-shaded fields are filled in at runtime.}
\label{fig:prompt_code}
\end{figure*}

\begin{figure*}[!ht]
\centering
\begin{tcolorbox}[
  colback=white,
  colframe=black!60,
  sharp corners,
  boxrule=0.8pt,
  width=\textwidth,
  title={Prompt Template — Policy Critic},
  fonttitle=\bfseries\small,
  colbacktitle=gray!15,
  coltitle=black,
  before skip=6pt,
  after skip=6pt,
  enhanced,
]
\scriptsize\setlength{\parindent}{0pt}\raggedright

\textbf{General Instructions}

\smallskip
You are a policy critic. Your goal is to judge whether a given policy is \emph{coordination-friendly} according to the definition below.
``Coordination-friendly'' is a defined term: do not add your own interpretation, even if the definition feels short or incomplete --- reason strictly from it.
It is critical that you never classify a coordination-unfriendly policy as coordination-friendly.

\smallskip
\textbf{Judge each local policy independently.}
Evaluate \texttt{player\_0} and \texttt{player\_1} as if each existed in isolation, without knowledge of the other.
At deployment, each local policy is paired with an independently generated partner --- not with the partner shown here.
A joint policy is coordination-friendly \emph{only if} both local policies are individually compliant.

\smallskip
\textbf{Also return a modified policy.}
\begin{itemize}\setlength{\itemsep}{1pt}\setlength{\topsep}{1pt}
  \item If the policy is not coordination-friendly, propose a modification that is, ensuring each local policy independently satisfies the CFD.
  \item Among compliant modifications, prefer ones that maximise expected joint reward.
  \item If the policy is already coordination-friendly, return it unchanged as the modified policy.
\end{itemize}

\tcbline

\textbf{Coordination-Friendly Definition (CFD)}

\smallskip
\colorbox{blue!8}{\strut\texttt{\$\,Coordination-Friendly Definition\,\$}}

\tcbline

\textbf{Policy under Review}

\smallskip
\colorbox{blue!8}{\strut\texttt{\$\,Policy (player\_0 and player\_1 code)\,\$}}

\tcbline

\textbf{Game Description}

\smallskip
The policy was generated for the game described below.

\smallskip
\colorbox{blue!8}{\strut\texttt{\$\,Game Description\,\$}}

\tcbline

\textbf{Policy Specification}

\smallskip
Each player has a local policy: an \texttt{act} function with the following signature:

\smallskip
\texttt{def act(observation\_dict: dict, memory: dict) -> tuple[str, dict]}

\smallskip
The first return value is the action (string); the second is the memory dictionary (initialised to \texttt{\{\}} at game start, not shared between players). Policies may use Python standard libraries \texttt{random} and \texttt{math}.

\smallskip
\textbf{Observation dict} --- index values using the \emph{exact} keys shown below (case-sensitive):

\smallskip
\colorbox{blue!8}{\strut\texttt{\$\,Sample Observation Dict\,\$}}

\smallskip
\textbf{Legal actions} --- the returned action must exactly match one of:

\smallskip
\colorbox{blue!8}{\strut\texttt{\$\,Legal Actions Dict\,\$}}

\smallskip
Policies that return an invalid action end the game immediately with a penalty.

\tcbline

\textbf{Output Specification}

\smallskip
Return a single JSON object under the structure below:

\smallskip
\begin{minipage}{\linewidth}
\ttfamily\small
\{\\
\quad "coordination\_friendly": \textit{<true or false --- verdict on the original policy>},\\
\quad "reasoning": \textit{<why the original policy is or is not coordination-friendly>},\\
\quad "modified\_policy": \{\\
\qquad "name": \textit{<name shared by both players>},\\
\qquad "reasoning": \textit{<why the modification is designed this way>},\\
\qquad "player\_0": \{ "code": \textit{<escaped Python string for \texttt{act}>} \},\\
\qquad "player\_1": \{ "code": \textit{<escaped Python string for \texttt{act}>} \}\\
\quad \}\\
\}
\end{minipage}

\medskip
\texttt{REASONING:}\\[1pt]
\texttt{<REASONING>}

\smallskip
\texttt{JSON:}
\begin{verbatim}
```json
<JSON>
```
\end{verbatim}

\end{tcolorbox}
\caption{Template of the prompt used for the policy critic. Blue-shaded fields are filled in at runtime.}
\label{fig:prompt_critic}
\end{figure*}

\medskip
\noindent The prompt templates used by the mutation operators in our meta-learning loop are shown in \autoref{fig:prompt_mutation_background}, \autoref{fig:prompt_mutation_unary}, \autoref{fig:prompt_mutation_thinking_styles}, \autoref{fig:prompt_mutation_diverse}, and \autoref{fig:prompt_mutation_reinit}. All mutation prompts begin with the shared background block in \autoref{fig:prompt_mutation_background}. The four \texttt{unary\_*} operators (\texttt{add}, \texttt{remove}, \texttt{reformulate}, \texttt{atomic}) share the template in \autoref{fig:prompt_mutation_unary}; the thinking-style block from \autoref{fig:prompt_mutation_thinking_styles} specifies which atomic edit the \acs{llm} is asked to perform. The four \texttt{diverse\_*} operators reuse the same thinking-style blocks but ask for $n_{\text{proposals}}$ diverse candidates per call (\autoref{fig:prompt_mutation_diverse}); we then sample one proposal uniformly at random. The \texttt{reinit} operator generates fresh single-bullet \acsp{cfd} from scratch using the seed template in \autoref{fig:prompt_mutation_reinit}.

\begin{figure*}[!ht]
\centering
\begin{tcolorbox}[
  colback=white,
  colframe=black!60,
  sharp corners,
  boxrule=0.8pt,
  width=\textwidth,
  title={Prompt Template: Shared Mutation Background},
  fonttitle=\bfseries\small,
  colbacktitle=gray!15,
  coltitle=black,
  before skip=6pt,
  after skip=6pt,
  enhanced,
]
\scriptsize\setlength{\parindent}{0pt}\raggedright

\textbf{Background}

\smallskip
In zero-shot coordination (ZSC), independently trained policies must coordinate successfully without any shared training or communication.

\smallskip
\textbf{Game Structure:}
\begin{itemize}\setlength{\itemsep}{1pt}\setlength{\topsep}{1pt}
  \item Fully cooperative two-player games
  \item Each player's behavior is defined by a Python function (local policy)
  \item Two local policies together form a joint policy
\end{itemize}

\smallskip
\textbf{Training Process:}
\begin{itemize}\setlength{\itemsep}{1pt}\setlength{\topsep}{1pt}
  \item $N$ independent training runs (seeds) are conducted
  \item In each seed, an LLM generates a policy candidate
  \item A \textbf{policy critic} then reviews each candidate against the CFD. The critic judges each local policy (\texttt{player\_0}, \texttt{player\_1}) \textbf{independently} --- as if it exists in isolation, without knowledge of the other player's policy. A joint policy is only approved if both local policies individually satisfy the CFD. The critic also tries to maximize the \textbf{self-play (SP) reward} --- but it has no knowledge of cross-play or other seeds.
  \item The CFD is the \textbf{only mechanism} through which you can influence compatibility across seeds. The critic will enforce whatever the CFD says on each local policy independently, so the CFD must be specific and unambiguous enough that independent critics produce compatible local policies.
  \item There is no shared training or communication between the seeds/policies.
\end{itemize}

\smallskip
\textbf{Evaluation - Cross-Play (XP) Score:}
\begin{itemize}\setlength{\itemsep}{1pt}\setlength{\topsep}{1pt}
  \item After $N$ independent runs, we have $N$ final policies
  \item Each policy is paired with every other policy ($N \times N$ matchups)
  \item XP score = average joint reward across all pairwise combinations
\end{itemize}

\smallskip
\textbf{Goal:} Evolve CFDs that maximize XP scores by encouraging policies to adopt compatible, coordination-friendly behaviors. Priortize avoiding total coordination failures, then think about raising the XP scores in general.

\end{tcolorbox}
\caption{Shared background block prepended to every mutation prompt.}
\label{fig:prompt_mutation_background}
\end{figure*}

\begin{figure*}[!ht]
\centering
\begin{tcolorbox}[
  colback=white,
  colframe=black!60,
  sharp corners,
  boxrule=0.8pt,
  width=\textwidth,
  title={Prompt Template: Unary Mutation Operators},
  fonttitle=\bfseries\small,
  colbacktitle=gray!15,
  coltitle=black,
  before skip=6pt,
  after skip=6pt,
  enhanced,
]
\scriptsize\setlength{\parindent}{0pt}\raggedright

You are helping to evolve a ``coordination-friendly definition'' (CFD) for zero-shot coordination (ZSC).

\smallskip
\colorbox{blue!8}{\strut\texttt{\$\,Shared Background (\autoref{fig:prompt_mutation_background})\,\$}}

\smallskip
\textbf{Current CFD}

\smallskip
\colorbox{blue!8}{\strut\texttt{\$\,CFD\,\$}}

\smallskip
\textbf{Your Task}

\smallskip
\colorbox{blue!8}{\strut\texttt{\$\,Task Instructions (see \autoref{fig:prompt_mutation_thinking_styles})\,\$}}

\smallskip
\textbf{Constraints}
\begin{itemize}\setlength{\itemsep}{1pt}\setlength{\topsep}{1pt}
  \item The CFD must have at most \colorbox{blue!8}{\strut\texttt{\$\,max\_bullet\_points\,\$}} bullet points
  \item Each bullet point should be a single, concise sentence
  \item \colorbox{blue!8}{\strut\texttt{\$\,Extra Constraints (see \autoref{fig:prompt_mutation_thinking_styles})\,\$}}
\end{itemize}

\smallskip
\textbf{Response Format}

\smallskip
Respond with valid JSON:

\smallskip
\begin{minipage}{\linewidth}
\ttfamily\small
\{\\
\quad "reasoning": \textit{<reasoning prompt, varies by thinking style>},\\
\quad "new\_cfd": "- First bullet point\textbackslash n- Second bullet point\textbackslash n..."\\
\}
\end{minipage}

\end{tcolorbox}
\caption{Combined template for the four \texttt{unary\_*} mutation operators (\texttt{unary\_add}, \texttt{unary\_remove}, \texttt{unary\_reformulate}, \texttt{unary\_atomic}). The thinking-style blocks substituted into the Task Instructions, Extra Constraints, and Reasoning Prompt placeholders are listed in \autoref{fig:prompt_mutation_thinking_styles}.}
\label{fig:prompt_mutation_unary}
\end{figure*}

\begin{figure*}[!ht]
\centering
\begin{tcolorbox}[
  colback=white,
  colframe=black!60,
  sharp corners,
  boxrule=0.8pt,
  width=\textwidth,
  title={Thinking-Style Blocks for Mutation Operators},
  fonttitle=\bfseries\small,
  colbacktitle=gray!15,
  coltitle=black,
  before skip=6pt,
  after skip=6pt,
  enhanced,
]
\scriptsize\setlength{\parindent}{0pt}\raggedright

The blocks below are substituted into the Task Instructions, Extra Constraints, and Reasoning Prompt placeholders in the unary template (\autoref{fig:prompt_mutation_unary}) and the diverse template (\autoref{fig:prompt_mutation_diverse}) to specialise each prompt for one of the four thinking styles.

\smallskip
\textbf{ADD} \quad (\texttt{unary\_add}, \texttt{diverse\_add})

\smallskip
\textit{Task Instructions:} \textbf{ADD} a new bullet point to the CFD. The new bullet point should:
\begin{itemize}\setlength{\itemsep}{1pt}\setlength{\topsep}{1pt}
  \item Address an aspect of coordination not yet covered by existing points
  \item Be complementary to (not redundant with) existing points
  \item Help policies coordinate better across different game scenarios
\end{itemize}

\textit{Extra Constraints:} (none)

\smallskip
\textit{Reasoning Prompt:} Explain why you're adding this bullet point and what coordination aspect it addresses

\smallskip
\textbf{REMOVE} \quad (\texttt{unary\_remove}, \texttt{diverse\_remove})

\smallskip
\textit{Task Instructions:} \textbf{REMOVE} one bullet point from the CFD. Choose to remove a bullet point that:
\begin{itemize}\setlength{\itemsep}{1pt}\setlength{\topsep}{1pt}
  \item Is redundant or overlaps significantly with other points
  \item Is too restrictive and may hurt coordination flexibility
  \item Is less essential than other points for achieving high XP scores
\end{itemize}

\textit{Extra Constraints:}
\begin{itemize}\setlength{\itemsep}{1pt}\setlength{\topsep}{1pt}
  \item The resulting CFD must have at least 1 bullet point. If the CFD has only 1 bullet point, reformulate it instead of removing it.
\end{itemize}

\textit{Reasoning Prompt:} Explain which bullet point you removed and why removing it improves the CFD

\smallskip
\textbf{REFORMULATE} \quad (\texttt{unary\_reformulate}, \texttt{diverse\_reformulate})

\smallskip
\textit{Task Instructions:} \textbf{REFORMULATE or REPLACE} one bullet point in the CFD. You may either:
\begin{itemize}\setlength{\itemsep}{1pt}\setlength{\topsep}{1pt}
  \item Reformulate: refine the wording to be clearer, more precise, or more actionable
  \item Replace: swap the bullet point for an entirely different principle that better encourages coordination
  \item Adapt: Change some part of the bullet point while keepint the rest the same
\end{itemize}
Choose whichever approach you believe will most improve the CFD's effectiveness.

\smallskip
\textit{Extra Constraints:} (none)

\smallskip
\textit{Reasoning Prompt:} Explain what you changed and why it improves the CFD

\smallskip
\textbf{ATOMIC} \quad (\texttt{unary\_atomic}, \texttt{diverse\_atomic})

\smallskip
\textit{Task Instructions:} Make exactly \textbf{ONE atomic change} to the CFD. Choose ONE of the following operations:
\begin{enumerate}\setlength{\itemsep}{1pt}\setlength{\topsep}{1pt}
  \item \textbf{ADD}: Add exactly one new bullet point that addresses a missing coordination aspect
  \item \textbf{REMOVE}: Remove exactly one bullet point that is redundant, too restrictive, or less essential
  \item \textbf{CHANGE}: Modify exactly one existing bullet point (rewording, refining, or replacing it)
\end{enumerate}
Pick the operation that you believe will most improve the CFD's effectiveness for zero-shot coordination.

\smallskip
\textit{Extra Constraints:}
\begin{itemize}\setlength{\itemsep}{1pt}\setlength{\topsep}{1pt}
  \item You must make exactly ONE change (add one, remove one, or change one bullet point)
  \item The resulting CFD must have at least 1 bullet point
\end{itemize}

\textit{Reasoning Prompt:} Explain which atomic operation you chose (add/remove/change), what specific change you made, and why

\end{tcolorbox}
\caption{Thinking-style blocks substituted into the unary (\autoref{fig:prompt_mutation_unary}) and diverse (\autoref{fig:prompt_mutation_diverse}) mutation prompt templates.}
\label{fig:prompt_mutation_thinking_styles}
\end{figure*}

\begin{figure*}[!ht]
\centering
\begin{tcolorbox}[
  colback=white,
  colframe=black!60,
  sharp corners,
  boxrule=0.8pt,
  width=\textwidth,
  title={Prompt Template: Diverse Mutation Operators},
  fonttitle=\bfseries\small,
  colbacktitle=gray!15,
  coltitle=black,
  before skip=6pt,
  after skip=6pt,
  enhanced,
]
\scriptsize\setlength{\parindent}{0pt}\raggedright

You are helping to evolve a ``coordination-friendly definition'' (CFD) for zero-shot coordination (ZSC).

\smallskip
\colorbox{blue!8}{\strut\texttt{\$\,Shared Background (\autoref{fig:prompt_mutation_background})\,\$}}

\smallskip
\textbf{Current CFD}

\smallskip
\colorbox{blue!8}{\strut\texttt{\$\,CFD\,\$}}

\smallskip
\textbf{Your Task}

\smallskip
\colorbox{blue!8}{\strut\texttt{\$\,Task Instructions (see \autoref{fig:prompt_mutation_thinking_styles})\,\$}}

\smallskip
\textbf{IMPORTANT:} Provide \colorbox{blue!8}{\strut\texttt{\$\,n\_proposals\,\$}} DIVERSE proposals for the modified CFD. Each proposal should take a meaningfully different approach to the mutation. Maximize diversity across your proposals.

\smallskip
Note: For some operations (e.g., removing a bullet point from a short CFD), fewer than \colorbox{blue!8}{\strut\texttt{\$\,n\_proposals\,\$}} meaningfully different options may exist. In that case, return as many distinct proposals as possible --- do not pad with near-duplicates.

\smallskip
\textbf{Constraints}
\begin{itemize}\setlength{\itemsep}{1pt}\setlength{\topsep}{1pt}
  \item Each CFD must have at most \colorbox{blue!8}{\strut\texttt{\$\,max\_bullet\_points\,\$}} bullet points
  \item Each bullet point should be a single, concise sentence
  \item \colorbox{blue!8}{\strut\texttt{\$\,Extra Constraints (see \autoref{fig:prompt_mutation_thinking_styles})\,\$}}
\end{itemize}

\smallskip
\textbf{Response Format}

\smallskip
Respond with valid JSON:

\smallskip
\begin{minipage}{\linewidth}
\ttfamily\small
\{\\
\quad "reasoning": \textit{<reasoning prompt, varies by thinking style>},\\
\quad "proposals": [\\
\qquad "- First bullet point\textbackslash n- Second bullet point\textbackslash n...",\\
\qquad "- First bullet point\textbackslash n- Second bullet point\textbackslash n...",\\
\qquad ...\\
\quad ]\\
\}
\end{minipage}

\end{tcolorbox}
\caption{Combined template for the four \texttt{diverse\_*} mutation operators (\texttt{diverse\_add}, \texttt{diverse\_remove}, \texttt{diverse\_reformulate}, \texttt{diverse\_atomic}). The differences from the unary template (\autoref{fig:prompt_mutation_unary}) are that the prompt asks for $n_{\text{proposals}}$ diverse candidates and that the singleton \texttt{new\_cfd} field is replaced with a \texttt{proposals} list. After parsing, we sample one proposal uniformly at random as the child CFD. Thinking-style blocks (\autoref{fig:prompt_mutation_thinking_styles}) are reused unchanged.}
\label{fig:prompt_mutation_diverse}
\end{figure*}

\begin{figure*}[!ht]
\centering
\begin{tcolorbox}[
  colback=white,
  colframe=black!60,
  sharp corners,
  boxrule=0.8pt,
  width=\textwidth,
  title={Prompt Template: Reinitialization Operator (\texttt{reinit})},
  fonttitle=\bfseries\small,
  colbacktitle=gray!15,
  coltitle=black,
  before skip=6pt,
  after skip=6pt,
  enhanced,
]
\scriptsize\setlength{\parindent}{0pt}\raggedright

You are helping to create diverse ``coordination-friendly definitions'' (CFDs) for zero-shot coordination (ZSC).

\smallskip
\colorbox{blue!8}{\strut\texttt{\$\,Shared Background (\autoref{fig:prompt_mutation_background})\,\$}}

\smallskip
\textbf{Your Task}

\smallskip
Generate \colorbox{blue!8}{\strut\texttt{\$\,n\,\$}} DIVERSE seed CFDs. Each CFD should:
\begin{itemize}\setlength{\itemsep}{1pt}\setlength{\topsep}{1pt}
  \item Be a single bullet point (one concise sentence)
  \item Encourage coordination in a unique way
  \item Be simple and general enough to apply to various multi-agent games
  \item NOT overlap significantly with other CFDs you generate
\end{itemize}

\smallskip
\textbf{Constraints}
\begin{itemize}\setlength{\itemsep}{1pt}\setlength{\topsep}{1pt}
  \item Generate exactly \colorbox{blue!8}{\strut\texttt{\$\,n\,\$}} CFDs
  \item Each CFD is a single bullet point (one sentence)
  \item Maximize diversity across the CFDs
\end{itemize}

\smallskip
\textbf{Response Format}

\smallskip
Respond with valid JSON:

\smallskip
\begin{minipage}{\linewidth}
\ttfamily\small
\{\\
\quad "reasoning": \textit{<Explain your strategy for creating diverse CFDs>},\\
\quad "cfds": [\\
\qquad "- First CFD",\\
\qquad "- Second CFD",\\
\qquad ...\\
\quad ]\\
\}
\end{minipage}

\end{tcolorbox}
\caption{Template used by the \texttt{reinit} operator. The operator calls this prompt with $n$ candidates per invocation and samples one uniformly at random as the child CFD. The same template is also used to create the initial seed population at the start of the meta-learning loop.}
\label{fig:prompt_mutation_reinit}
\end{figure*}

\subsection{Experimental Resources}
\label{sec:app/experimental_resources}

Running all experiments reported in this paper incurred approximately \$1{,}000 in API costs across the evaluated \acs{llm} agents. The token-level probability experiments (\appref{sec:app/local_policies}) for the open-weight models were run locally on a single H100 GPU.








\clearpage
\clearpage
\section{Reasoning Traces}
\label{sec:app/reasoning_traces}
This section contains reasoning traces of various LLM agents across the environments studied in this paper, illustrating both successful coordination and common failure modes.

\begin{figure*}[t]
\centering
\begin{tcolorbox}[
  colback=white,
  colframe=black!70,
  sharp corners,
  boxrule=0.8pt,
  width=\textwidth,
  left=1mm,
  right=1mm,
  top=1mm,
  bottom=1mm,
  before skip=6pt,
  after skip=6pt,
  enhanced,
]

\begin{tcolorbox}[
  breakable,
  geministyle,
  title={Gemini 3 Pro (Turn 2) --- giving H1},
  fonttitle=\bfseries\small,
  boxrule=0.6pt,
  left=1mm,
  right=1mm,
  top=0.8mm,
  bottom=0.8mm,
  enhanced,
]
\scriptsize\ttfamily
\setlength{\parindent}{0pt}
\raggedright
Fireworks: Y=1 \quad Info: 2/3 \quad Lives: 1/1 \\[2pt]
\begin{tabular}{@{}ll@{\hspace{8mm}}ll@{}}
\textbf{Own hand (hidden):} & & \textbf{Partner's hand:} & \\
pos 0: {[}RYG123{]} & & pos 0: Y2 & \\
pos 1: {[}RYG123{]} & & pos 1: Y3 & \\
pos 2: {[}RYG123{]} & & pos 2: G2 & \\
pos 3: {[}RYG123{]} & & pos 3: \textbf{Y1}~~hint:1 & \textit{(duplicate of placed Y1)} \\
pos 4: {[}RYG123{]} & & pos 4: \textbf{R1}~~no hint & \textit{(playable)} \\
\end{tabular}
\\[4pt]
The other player has a playable R1 at position 4. By giving a rank hint of `1', they will know that the card at position 4 is a 1 (since position 3 was already hinted as a 1 previously). This will allow them to safely play the R1 to start the Red fireworks.

\textbf{Action: H1 $\to$ touches positions 3 and 4.}
\end{tcolorbox}

\begin{tcolorbox}[
  breakable,
  gptstyle,
  title={GPT 5.2 (Turn 3) --- responding to H1},
  fonttitle=\bfseries\small,
  boxrule=0.6pt,
  left=1mm,
  right=1mm,
  top=0.8mm,
  bottom=0.8mm,
  enhanced,
]
\scriptsize\ttfamily
\setlength{\parindent}{0pt}
\raggedright
Fireworks: Y=1 \quad Info: 1/3 \quad Lives: 1/1 \\[2pt]
\begin{tabular}{@{}ll@{\hspace{8mm}}ll@{}}
\textbf{Own hand (hidden):} & & \textbf{Partner's hand:} & \\
pos 0: {[}RYG23{]}          & & pos 0: R1 & \\
pos 1: {[}RYG23{]}          & & pos 1: G3 & \\
pos 2: {[}RYG23{]}          & & pos 2: G1 & \\
pos 3: hint:1~~{[}RYG1{]}   & \textit{(actually Y1, duplicate)} & pos 3: R1 & \\
pos 4: hint:1~~{[}RYG1{]}   & \textit{(actually R1, playable)}  & pos 4: G1 & \\
\end{tabular}
\\[4pt]
My cards at positions 3 and 4 were just hinted as 1s after Y1 was already played, so this strongly suggests they are the other playable 1s. Playing the older hinted card is the best safe progress.

\textbf{Action: play position 3.}
\textrightarrow{} \textbf{Card was Y1 (duplicate of placed Y1). Life lost. Final score: 0.}
\end{tcolorbox}

\end{tcolorbox}
\caption{\textbf{Coordination failure between Gemini~3~Pro and GPT~5.2 in Hanabi 3$\times$3 due to mismatched assumptions about the ``Good Touch'' principle (i.e, all clued cards will be played in the future), a convention widely adopted by human Hanabi players.}
Gemini provides a rank-1 hint to signal a newly drawn R1 at position~4, but the hint also re-touches an already-played duplicate Y1 at position~3, violating the ``Good Touch'' principle. GPT, assuming that a cooperative partner “would not hint cards that are already played,” infers that both touched cards must be playable 1s and plays the older hinted card (the Y1 duplicate), costing the team its only life. This failure is compounded by an implicit convention, playing the oldest hinted card, that was not mutually established; had the clue giver anticipated this assumption, they would have avoided giving the hint.}
\label{trace:hanabi_good_touch_failure}
\end{figure*}

\begin{figure*}[t]
\centering
\begin{tcolorbox}[
  colback=white,
  colframe=black!70,
  sharp corners,
  boxrule=0.8pt,
  width=\textwidth,
  left=1mm,
  right=1mm,
  top=1mm,
  bottom=1mm,
  before skip=6pt,
  after skip=6pt,
  enhanced,
]

\begin{tcolorbox}[
  breakable,
  gptstyle,
  title={GPT 5.2},
  fonttitle=\bfseries\small,
  boxrule=0.6pt,
  left=1mm,
  right=1mm,
  top=0.8mm,
  bottom=0.8mm,
  enhanced,
]
\scriptsize\ttfamily
\setlength{\parindent}{0pt}
\raggedright
To maximize zero-shot coordination, use an (essentially forced) perfect-signaling convention: when player\_0 has Card 2, they must play "A" to enable a payoff-10 outcome regardless of player\_1’s card (since Case 3 only allows 10 via (A,C)). When player\_0 has Card 1, they can safely play "C" because it still allows payoff-10 outcomes for both possible player\_1 cards ((C,A) in Case 1 and (C,C) in Case 2). Player\_1 then best-responds deterministically based on (player\_0\_action, own card) to always reach payoff 10. Additionally, if player\_1 ever observes "B", responding with "B" is a robust fallback since (B,B)=8 in all cases.
\end{tcolorbox}

\begin{tcolorbox}[
  breakable,
  opusstyle,
  title={Opus 4.5},
  fonttitle=\bfseries\small,
  boxrule=0.6pt,
  left=1mm,
  right=1mm,
  top=0.8mm,
  bottom=0.8mm,
  enhanced,
]
\scriptsize\ttfamily
\setlength{\parindent}{0pt}
\raggedright
Looking at this game, I need to analyze the payoff structure and find a coordination-friendly strategy.

Analysis:

Key observations from the payoff tables:
\begin{enumerate}
\item (B, B) always gives 8 regardless of cards -- this is a ``safe'' play
\item 10-point payoffs exist but require knowing which case we're in
\item Player 0 can signal their card through their action choice, enabling Player 1 to know the exact case
\end{enumerate}

Signaling strategy:
\begin{itemize}
\item If Player 0 plays A $\rightarrow$ signals Card 1
\item If Player 0 plays B $\rightarrow$ signals Card 2
\end{itemize}

Player 1's optimal responses:
\begin{itemize}
\item When P0 plays A (P0 has Card 1):
  \begin{itemize}
  \item P1 Card 1 (Case 1): Play A $\rightarrow$ 10
  \item P1 Card 2 (Case 2): Play C $\rightarrow$ 10
  \end{itemize}
\item When P0 plays B (P0 has Card 2):
  \begin{itemize}
  \item P1 Card 1 (Case 3): Play B $\rightarrow$ 8 (best available)
  \item P1 Card 2 (Case 4): Play B $\rightarrow$ 8 (best available)
  \end{itemize}
\end{itemize}

Expected value: $(10 + 10 + 8 + 8)/4 = 9$

This is better than always playing (B, B), which gives 8, and the convention ``A for 1, B for 2'' is natural and coordination-friendly.

REASONING:

The optimal ZSC strategy uses Player 0's action to signal their card. Using ``A'' for Card 1 and ``B'' for Card 2 is a natural, memorable convention. Player 1 can then deduce the exact game case and respond optimally. When P0 has Card 1 (plays A), P1 can always achieve 10 points. When P0 has Card 2 (plays B), P1's best option is B, giving 8 points. This yields an expected value of 9.
\end{tcolorbox}

\end{tcolorbox}
\caption{\textbf{Reasoning traces of GPT 5.2 and Opus 4.5 agents in Tiny Hanabi.} GPT 5.2 chooses a joint policy based on SP optimality wheres Opus picks a joint policy based on \textit{lexicographical} ordering.}
\label{trace:signaling_failure}
\end{figure*}
\begin{figure*}[t]
\centering
\begin{tcolorbox}[
  colback=white,
  colframe=black!70,
  sharp corners,
  boxrule=0.8pt,
  width=\textwidth,
  left=1mm,
  right=1mm,
  top=1mm,
  bottom=1mm,
  before skip=6pt,
  after skip=6pt,
  enhanced,
]

\begin{tcolorbox}[
  breakable,
  grokstyle,
  title={Grok 4.1 Fast},
  fonttitle=\bfseries\small,
  boxrule=0.6pt,
  left=1mm,
  right=1mm,
  top=0.8mm,
  bottom=0.8mm,
  enhanced,
]
\scriptsize\ttfamily
\setlength{\parindent}{0pt}
\raggedright
The optimal coordination strategy leverages the asymmetry: player\_0 signals its card (A for 1, C for 2). Player\_1 interprets this signal to infer player\_0's card, determines if cards match (both 1 or both 2) or mismatch, and responds to achieve the 10 payoff:

Match (signaled same as own): Play A (yields A,A or C,A = 10).

Mismatch (signaled different): Play C (yields A,C or C,C = 10). This works in all cases and is robust for ZSC as it uses a natural, symmetric signaling convention (A$\approx$1, C$\approx$2, B safe fallback). B,B=8 always as fallback if signal invalid. Expected payoff 10 if both use this policy. Memory unused (single turn).
\end{tcolorbox}

\begin{tcolorbox}[
  breakable,
  gptstyle,
  title={GPT 5.2},
  fonttitle=\bfseries\small,
  boxrule=0.6pt,
  left=1mm,
  right=1mm,
  top=0.8mm,
  bottom=0.8mm,
  enhanced,
]
\scriptsize\ttfamily
\setlength{\parindent}{0pt}
\raggedright
This game has a ``safe'' coordination point: (B,B) yields 8 regardless of both cards, so it is highly compatible with many independently-designed policies and avoids catastrophic 0s. However, if player\_0 uses A/C as a signal for their card (a natural convention: A$\rightarrow$Card1, C$\rightarrow$Card2), player\_1 can often achieve 10 by responding conditionally on their own card. The joint design below therefore uses a risk-dominant anchor (player\_0 always plays B), while player\_1 is compatible with B and also opportunistically decodes A/C in the most likely signaling scheme.
\end{tcolorbox}

\end{tcolorbox}
\caption{\textbf{Reasoning traces of Grok 4.1 Fast and GPT 5.2 as code agents in Tiny Hanabi Fixed.} Both agents independently converge on the lexicographic mapping (A$\,\approx\,$1, C$\,\approx\,$2): Grok uses it as an active signalling convention, while GPT uses it to decode the partner's signal. This shared semantic bias enables agents to coordinate via a convention rather than defaulting to the safe fallback.}
\label{trace:tiny_hanabi_fixed_convergence}
\end{figure*}

\begin{figure*}[t]
\centering
\begin{tcolorbox}[
  colback=white,
  colframe=black!70,
  sharp corners,
  boxrule=0.8pt,
  width=\textwidth,
  left=1mm,
  right=1mm,
  top=1mm,
  bottom=1mm,
  before skip=6pt,
  after skip=6pt,
  enhanced,
]

\begin{tcolorbox}[
  breakable,
  grokstyle,
  title={Grok 4.1 Fast},
  fonttitle=\bfseries\small,
  boxrule=0.6pt,
  left=1mm,
  right=1mm,
  top=0.8mm,
  bottom=0.8mm,
  enhanced,
]
\scriptsize\ttfamily
\setlength{\parindent}{0pt}
\raggedright
The optimal strategy for zero-shot coordination is a fixed, symmetric signaling convention that both policies can independently derive from the shared prompt. The levers are listed as ['blue', 'red'] and legal actions list 'blue' before 'red'; signals as ['square', 'circle'] with 'square' before 'circle' in legal actions. This suggests a canonical 1:1 mapping: 'square'$\leftrightarrow$'blue', 'circle'$\leftrightarrow$'red'. Alice always encodes the correct lever into the matching signal (free), expecting +10 joint reward per round if Bob decodes correctly. Bob decodes the signal to the matching lever or follows a reveal if present (-5 +10 = +5). No bailing (low reward: +1 or +0.5) or revealing (costly unless needed). This yields +100 expected if coordinating, robust as Bob safely follows reveals/signals. Policies are simple, use no arbitrary choices/memory, ignore history (no need if convention holds), and always valid. Excellent ZSC as different LLMs likely infer the same prompt-order mapping.
\end{tcolorbox}

\begin{tcolorbox}[
  breakable,
  geministyle,
  title={Gemini 3 Pro},
  fonttitle=\bfseries\small,
  boxrule=0.6pt,
  left=1mm,
  right=1mm,
  top=0.8mm,
  bottom=0.8mm,
  enhanced,
]
\scriptsize\ttfamily
\setlength{\parindent}{0pt}
\raggedright
The game is a pure coordination game where the agents must agree on a mapping between signals ('circle', 'square') and levers ('blue', 'red'). Since the agents cannot communicate and are evaluated in a Zero-Shot Coordination (ZSC) setting, they must rely on a "Schelling point" — a solution that is naturally salient to both parties without prior agreement.

The most robust Schelling point for mapping distinct sets of strings is Lexicographical Ordering (Alphabetical Sort). This method is external, objective, and independent of arbitrary presentation order in prompts.

Sorting Levers: 'blue' comes before 'red' alphabetically.
Index 0: 'blue'
Index 1: 'red'
Sorting Signals: 'circle' comes before 'square' alphabetically.
Index 0: 'circle'
Index 1: 'square'
Therefore, the proposed policy adopts the convention:

'circle' maps to 'blue'
'square' maps to 'red'
Alice's policy simply observes the correct lever and sends the corresponding alphabetic signal. Bob's policy checks for a revealed lever first (taking the guaranteed points), and if none, interprets the signal using the same alphabetic sorting convention. This maximizes the joint reward (10 per round) and minimizes the risk of miscoordination.
\end{tcolorbox}

\begin{tcolorbox}[
  breakable,
  deepseekstyle,
  title={Deepseek V3.2},
  fonttitle=\bfseries\small,
  boxrule=0.6pt,
  left=1mm,
  right=1mm,
  top=0.8mm,
  bottom=0.8mm,
  enhanced,
]
\scriptsize\ttfamily
\setlength{\parindent}{0pt}
\raggedright
We design a policy that establishes a clear convention for signaling the correct lever without direct communication. The key idea is to use the previous known correct lever as a reference point. Alice reveals the correct lever in the first round to establish a baseline. In subsequent rounds, Alice signals "square" if the current correct lever is the same as the last known correct lever, and "circle" if it changes. Bob uses the same rule to interpret signals and picks the lever accordingly. This approach minimizes ambiguity and ensures coordinated guesses, maximizing the chance of +10 rewards while avoiding costly mistakes. Both policies maintain a memory of the last known correct lever, updated based on revealed levers and outcomes, ensuring robustness across rounds.
\end{tcolorbox}

\end{tcolorbox}
\caption{\textbf{Reasoning traces of Grok 4.1 Fast, Gemini 3 Pro and Deepseek V3.2 when generating the most common observed strategies in Adhoc conventions.} Note that the three strategies are mutually incompatible.}
\label{trace:adhoc_conventions_failure}
\end{figure*}
\begin{figure*}[t]
\centering
\begin{tcolorbox}[
  colback=white,
  colframe=black!70,
  sharp corners,
  boxrule=0.8pt,
  width=\textwidth,
  left=1mm,
  right=1mm,
  top=1mm,
  bottom=1mm,
  before skip=6pt,
  after skip=6pt,
  enhanced,
]

\begin{tcolorbox}[
  breakable,
  geministyle,
  title={Gemini 3 Pro},
  fonttitle=\bfseries\small,
  boxrule=0.6pt,
  left=1mm,
  right=1mm,
  top=0.8mm,
  bottom=0.8mm,
  enhanced,
]
\scriptsize\ttfamily
\setlength{\parindent}{0pt}
\raggedright

So, the challenge is to establish a communication protocol over a noisy channel (unknown button -> signal mapping) without talking. The key is to realize that while Bob cannot interpret the signal in Round 1, he receives the "ground truth" (the correct lever) at the end of the round. This allows him to perform "supervised learning" on Alice's signaling strategy.

Strategy:

Exploration (Round 1):
Alice sends a signal corresponding to the correct lever using a fixed convention (Alphabetical: Apple=Blue, Orange=Red). Since Bob does not know the mapping yet, he cannot safely guess a lever (risk of -10 is too high compared to guessing reward). Instead, Bob plays "bail" to secure a small positive reward (0.5).

Learning:
At the start of subsequent rounds, Bob looks at the history. He sees what signal Alice sent in previous rounds and what the actual correct lever was. He deduces the mapping (e.g., "Ah, last round I saw a Square and the lever was Red, so Square means Red").

Exploitation (Round 2-10):
Once the mapping is learned, Bob can perfectly decode Alice's signals and pick the correct lever for +10 reward.

This yields an expected score of 0.5 + 9 * 10 = 90.5, which is far superior to blind guessing or bailing every round.

Policy Details:

Alice:
Simply maps levers to buttons Alphabetically (Blue->Apple, Red->Orange). This is a strong focal point.

Bob:
Checks if he has historical data. If yes, he deduces the mapping and acts on the current signal. If no (Round 1), he bails. He also handles the edge case where Alice "reveals" the lever (though she won't in this policy, he should handle it for robustness).

\end{tcolorbox}

\begin{tcolorbox}[
  breakable,
  grokstyle,
  title={Grok 4.1 Fast},
  fonttitle=\bfseries\small,
  boxrule=0.6pt,
  left=1mm,
  right=1mm,
  top=0.8mm,
  bottom=0.8mm,
  enhanced,
]
\scriptsize\ttfamily
\setlength{\parindent}{0pt}
\raggedright
The optimal strategy in ZSC requires a canonical, symmetry-breaking protocol to learn the fixed button-signal bijection using the first two rounds for calibration.

Calibration (Rounds 1-2):
Alice presses buttons in lexical order ('apple' then 'orange'), ignoring the levers. Bob observes the signals in these rounds (synchronized via global round count), stores the mapping from signal to button, and bails safely (+0.5 each round).

Exploitation (Round 3+):
Alice encodes the current lever lexically ('blue' -> 'apple', 'red' -> 'orange'). Bob decodes the signal to the intended lever and selects it (+10).

Synchronization:
Both agents use the global round number (len(correct\_levers) for Alice, +1 for Bob) to remain perfectly synchronized, regardless of bails.

Robustness:
The strategy includes handling for reveals/None cases.

Convention:
Canonical lexical ordering (apple < orange, blue < red) ensures coordination across LLMs.

Expected reward:
0.5 * 2 + 10 * 8 = 81
\end{tcolorbox}

\end{tcolorbox}
\caption{\textbf{Reasoning Traces of Gemini 3 Pro and Grok 4.1 Fast in Ad-hoc conventions permuted signals}. Although both models converge on a scheme where Alice remains fixed and Bob adapts, they adopt different calibration strategies for Bob, rendering them incompatible.}
\label{trace:ad_hoc_conventions_permuted_signals}
\end{figure*}

\begin{figure*}[t]
\centering
\begin{tcolorbox}[
  colback=white,
  colframe=black!70,
  sharp corners,
  boxrule=0.8pt,
  width=\textwidth,
  left=1mm,
  right=1mm,
  top=1mm,
  bottom=1mm,
  before skip=6pt,
  after skip=6pt,
  enhanced,
]

\begin{tcolorbox}[
  breakable,
  grokstyle,
  title={Grok 4.1 Fast},
  fonttitle=\bfseries\small,
  boxrule=0.6pt,
  left=1mm,
  right=1mm,
  top=0.8mm,
  bottom=0.8mm,
  enhanced,
]
\scriptsize\ttfamily
\setlength{\parindent}{0pt}
\raggedright

The game requires crossing positions 0 to 6 while ensuring both players are not on the bridge (positions 1-5) for more than 1 consecutive turn. Identical policies fail to coordinate due to symmetry (both would forward on turn 2, breaking the bridge). Thus, use asymmetric roles with a clear convention: player\_0 (leader) always "forward" to cross first in 6 turns. player\_1 (follower) stays at 0 until observing leader at position 4 (after turn 4), then "forward". This results in both on bridge only on turn 5 (leader at 5, follower at 1; first instance), turn 6 has leader off at 6. Follower reaches 6 on turn 10 (-10 reward). Safe, minimal backs, leverages ZSC by converging on player\_0=leader, player\_1=follower across LLMs. Robust as actions always legal.

\end{tcolorbox}

\begin{tcolorbox}[
  breakable,
  deepseekstyle,
  title={Deepseek V3.2 (poor SP)},
  fonttitle=\bfseries\small,
  boxrule=0.6pt,
  left=1mm,
  right=1mm,
  top=0.8mm,
  bottom=0.8mm,
  enhanced,
]
\scriptsize\ttfamily
\setlength{\parindent}{0pt}
\raggedright
The game requires coordination to maximize reward and avoid the catastrophic bridge collapse. Key insights:

The bridge can only support both players for 1 consecutive turn. This implies we must not have both players on the bridge (positions 1-5) for two turns in a row.
To minimize turns, we need to cross as quickly as possible, but safety is paramount due to the large penalty (-1000).
Without communication, players must rely on observing past actions to coordinate. A simple and robust strategy is to have one player wait at the start while the other crosses completely, then the waiting player crosses. This avoids simultaneous bridge occupancy entirely and is coordination-friendly in ZSC.
To break symmetry (both players might otherwise adopt the same policy and both wait or both go), we use a deterministic rule based on comparing paths: the player whose historical action sequence is lexicographically smaller becomes the "leader" who goes first. This is a common symmetry-breaking technique in ZSC that doesn't require randomness (which could lead to miscoordination) and ensures both players compute the same leader/follower roles.
Memory is used to store the computed role after the first turn to avoid recomputation.
The policy ensures valid actions at all times and handles edge cases (e.g., already at goal).
\end{tcolorbox}

\begin{tcolorbox}[
  breakable,
  gptstyle,
  title={GPT 5.2},
  fonttitle=\bfseries\small,
  boxrule=0.6pt,
  left=1mm,
  right=1mm,
  top=0.8mm,
  bottom=0.8mm,
  enhanced,
]
\scriptsize\ttfamily
\setlength{\parindent}{0pt}
\raggedright
To maximize zero-shot coordination, the policy uses a widely compatible "single-file bridge" convention: never enter the bridge if the other player is currently on it, and if you are on the bridge while the other is not, keep moving forward to clear it quickly. The only hard case is simultaneous entry (both move from start to position 1); to resolve this without communication, the policy uses randomized backoff at the start so that eventually only one player enters first while the other waits. If an unexpected overlap on the bridge occurs, an emergency rule tries to separate the players by sending the one closer to the start backward and the one closer to the goal forward (a symmetric, observation-based convention).

\end{tcolorbox}

\end{tcolorbox}
\caption{\textbf{Reasoning Traces of Grok 4.1 Fast, Deepseek V3.2 and GPT 5.2 in Brittle Bridge.} DeepSeek produces a complex role-assignment strategy that performs poorly even in self-play. In contrast, Grok 4.1 Fast and GPT 5.2 each learn role-assignment strategies that are mutually incompatible.}
\label{trace:brittle_bridge}
\end{figure*}
\begin{figure*}[t]
\centering
\begin{tcolorbox}[
  colback=white,
  colframe=black!70,
  sharp corners,
  boxrule=0.8pt,
  width=\textwidth,
  left=1mm,
  right=1mm,
  top=1mm,
  bottom=1mm,
  before skip=6pt,
  after skip=6pt,
  enhanced,
]

\begin{tcolorbox}[
  breakable,
  opusstyle,
  title={Opus 4.5},
  fonttitle=\bfseries\small,
  boxrule=0.6pt,
  left=1mm,
  right=1mm,
  top=0.8mm,
  bottom=0.8mm,
  enhanced,
]
\scriptsize\ttfamily
\setlength{\parindent}{0pt}
\raggedright
Looking at this game, I need to design a policy for a narrow tunnel coordination problem.

The key challenge is:

- Player 0 starts at position 0, needs to reach position 4
- Player 1 starts at position 4, needs to reach position 0
- There is a narrow tile at position 2 that creates a bottleneck
- Both players cannot be on the narrow tile simultaneously
- Players cannot swap positions when one is on the narrow tile

For ZSC, I need a deterministic convention that independently generated policies would converge on. A natural symmetry-breaking approach is to use the goal direction (which is observable) to determine who yields.

Chosen convention:
The player going RIGHT (goal = ``right'') yields by waiting at position 1 until the player going LEFT has passed through the narrow section. This is a natural, deterministic convention based on observable information.

Optimal sequence with this convention:

Turn 1: P0->1, P1->3  
Turn 2: P0 stays at 1, P1->2  
Turn 3: P0 stays at 1, P1->1 (both on wide tile 1)  
Turn 4: P0->2, P1->0  
Turn 5: P0->3  
Turn 6: P0->4  

This achieves the goal in 6 turns, which is optimal given the bottleneck constraint.
\end{tcolorbox}

\begin{tcolorbox}[
  breakable,
  gptstyle,
  title={GPT 5.2},
  fonttitle=\bfseries\small,
  boxrule=0.6pt,
  left=1mm,
  right=1mm,
  top=0.8mm,
  bottom=0.8mm,
  enhanced,
]
\scriptsize\ttfamily
\setlength{\parindent}{0pt}
\raggedright
The key coordination problem is the single narrow tile (index 2): if both try to enter it from opposite sides on the same turn, nobody moves; and if they try to “swap” across it, the swap is blocked. A ZSC-friendly solution is to use a simple, role-inferable convention that both agents will independently follow: the agent whose goal is "right" (i.e., the left-starting agent) has priority to enter the narrow tile first when both are poised to enter from opposite sides (positions 1 and 3). The other agent yields by waiting. This convention is symmetric from the game description (it depends only on the publicly inferable goal direction), avoids deadlock, and achieves the standard efficient 6-turn crossing.
\end{tcolorbox}

\end{tcolorbox}
\caption{\textbf{Reasoning Traces of Opus 4.5 and GPT 5.2 in Narrow Tunnel.} These traces illustrate the two most common yet incompatible role-assignment strategies: “left goes first” versus “right goes first.”}
\label{trace:narrow_tunnel}
\end{figure*}

\begin{figure*}[t]
\centering
\begin{tcolorbox}[
  colback=white,
  colframe=black!70,
  sharp corners,
  boxrule=0.8pt,
  width=\textwidth,
  left=1mm,
  right=1mm,
  top=1mm,
  bottom=1mm,
  before skip=6pt,
  after skip=6pt,
  enhanced,
]

\begin{tcolorbox}[
  breakable,
  opusstyle,
  title={Opus 4.5},
  fonttitle=\bfseries\small,
  boxrule=0.6pt,
  left=1mm,
  right=1mm,
  top=0.8mm,
  bottom=0.8mm,
  enhanced,
]
\scriptsize\ttfamily
\setlength{\parindent}{0pt}
\raggedright
Looking at the 11x11 matrix game, I need to find a focal point that both players would independently converge on for zero-shot coordination.

Key observations about the matrix:

- The entry (0,0) = 1.0 (maximum reward)
- Entry (1,0) = 1.0 (also maximum reward) -- notably, rows 0 and 1 both have their maximum at column 0
- The diagonal entries (i,i) for i >= 2 all equal 1.0, with small 0.02 values on adjacent off-diagonal entries
- Most other entries are 0

For zero-shot coordination, I need the most natural focal point. The (0,0) cell is the strongest candidate because:

- Smallest index convention: ``Pick the first/smallest option'' is a universal coordination heuristic
- Symmetry: both players choosing 0 is symmetric
- Column 0 is special: both rows 0 and 1 give reward 1.0 only at column 0, making column 0 feel more ``focal'' than other columns
- Maximum reward: the payoff of 1.0 is the best achievable

Other focal points like ``meet in the middle'' (5,5) or ``go to the end'' (10,10) are possible but less universally natural than ``start at the beginning.''

Therefore, the optimal coordination strategy is for both players to always choose action ``0''.

\end{tcolorbox}

\begin{tcolorbox}[
  breakable,
  deepseekstyle,
  title={Deepseek V3.2},
  fonttitle=\bfseries\small,
  boxrule=0.6pt,
  left=1mm,
  right=1mm,
  top=0.8mm,
  bottom=0.8mm,
  enhanced,
]
\scriptsize\ttfamily
\setlength{\parindent}{0pt}
\raggedright
The game is a one-turn matrix game with a fixed payoff matrix. The optimal joint action is to coordinate on a cell with the highest possible reward (1.0). However, there are multiple such cells: (0,0), (1,0), and the main diagonal from (2,2) to (10,10). Since agents cannot communicate and the game is played only once, the key challenge is symmetry-breaking: both players must independently choose the same convention to coordinate on one of these high-payoff cells.

Given the matrix structure, the diagonal cells (i,i) for i=2..10 are particularly salient because they form a clear, symmetric pattern. Among these, the most natural focal point is the center of the diagonal—index 6—as it is the midpoint of the range 2-10. This is a common human convention (e.g., "meet in the middle") and likely to be chosen independently by other LLM-generated policies, maximizing zero-shot coordination.

Choosing 0 or 1 carries risk: if one player picks 0 and the other picks any index except 0, the payoff is 0. The diagonal strategy is more robust because even if coordination fails slightly (e.g., one picks 5 and the other 6), the payoff is still 0.02, which is better than 0. Thus, the policy for both players should be to choose action 6.

Since the game is one-turn, memory is unused but must be returned unchanged.

\end{tcolorbox}

\end{tcolorbox}
\caption{\textbf{Reasoning Traces of Opus 4.5 and Deepseek V3.2 in Matrix Game.} Opus 4.5 successfully identify different focal points from the payoff matrix.}
\label{trace:matrix_game_failure}
\end{figure*}

\begin{figure*}[t]
\centering
\begin{tcolorbox}[
  colback=white,
  colframe=black!70,
  sharp corners,
  boxrule=0.8pt,
  width=\textwidth,
  left=1mm,
  right=1mm,
  top=1mm,
  bottom=1mm,
  before skip=6pt,
  after skip=6pt,
  enhanced,
]

\begin{tcolorbox}[
  breakable,
  opusstyle,
  title={Opus 4.5},
  fonttitle=\bfseries\small,
  boxrule=0.6pt,
  left=1mm,
  right=1mm,
  top=0.8mm,
  bottom=0.8mm,
  enhanced,
]
\scriptsize\ttfamily
\setlength{\parindent}{0pt}
\raggedright
This is a coordination game with hidden permutations -- each player sees levers numbered 1--5, but the underlying mapping is different for each player.

Key insight:

When I observe the other player's action, I see it in MY coordinate system. If they picked their lever X and I observe it as lever Y (in my view), then if I pick lever Y, I will be selecting the same underlying lever they chose.

Naively, if both players "follow" (i.e., switch to what they observed), they will keep swapping and never converge. We therefore need exactly ONE player to follow and ONE to stay.

Symmetry-breaking rule:

- If you observe a HIGHER number than your own action, you follow
- If you observe LOWER or EQUAL, you stay

Why this works:

- If I see higher (X > my\_action): I follow → pick X
- If they see lower (my\_action in their view < their action): they stay  
- Result: I move to their action, they stay → convergence

Initialization:

Start with lever 1 (a Schelling focal point -- the lowest number), then apply the rule above. Once both players match (reward = 1), repeat the successful action.

Edge case:

If both players observe the same number as their own action (X = 1, Y = 1), both will stay. This case has low probability under random permutations, and the system will eventually resolve once observations differ.

\end{tcolorbox}

\begin{tcolorbox}[
  breakable,
  deepseekstyle,
  title={Deepseek V3.2},
  fonttitle=\bfseries\small,
  boxrule=0.6pt,
  left=1mm,
  right=1mm,
  top=0.8mm,
  bottom=0.8mm,
  enhanced,
]
\scriptsize\ttfamily
\setlength{\parindent}{0pt}
\raggedright
This is a zero-shot coordination problem where two players must choose the same lever under a random, fixed permutation unknown to them. The key challenge is establishing a robust convention without communication. Observing each other's actions (though permuted) and joint rewards provides limited feedback.

Since both players see the same structure, we can adopt a symmetry-breaking strategy that uses the first few turns to infer a common reference point. A reliable approach is to treat the first turn as a "coordination attempt": both players choose lever 1 initially. If they receive a reward of 1, they continue choosing lever 1. If not, they switch to "mirroring" the other player's last action, which accounts for the unknown permutation by aligning on the lever the other player just chose in their own perspective. This adaptive rule maximizes the chance of converging quickly, as mirroring is a natural coordination heuristic when permutations are unknown. Storing the initial lever in memory ensures consistency.

The policy uses a simple state machine:

Start by choosing lever "1".
If the last reward was 1, stick with the last successful lever.
Otherwise, mirror the other player's last action. This balances exploration and exploitation over the 10 turns and is likely to be independently chosen by another LLM, promoting coordination.
\end{tcolorbox}

\end{tcolorbox}
\caption{\textbf{Reasoning Traces of Opus 4.5 and Deepseek V3.2 in Permuted Levers.} Opus 4.5 achieves a low self-play score but exhibits strong cross-play performance when paired with DeepSeek V3.2.}
\label{trace:permuted_levers}
\end{figure*}

\begin{figure*}[t]
\centering
\begin{tcolorbox}[
  colback=white,
  colframe=black!70,
  sharp corners,
  boxrule=0.8pt,
  width=\textwidth,
  left=1mm,
  right=1mm,
  top=1mm,
  bottom=1mm,
  before skip=6pt,
  after skip=6pt,
  enhanced,
]

\begin{tcolorbox}[
  breakable,
  grokstyle,
  title={Grok 4.1 Fast (Hinter) — hand: 1B, 3A, 1B, 3C, 2B \quad target: 1B \quad hint: \textbf{1B}},
  fonttitle=\bfseries\small,
  boxrule=0.6pt,
  left=1mm,
  right=1mm,
  top=0.8mm,
  bottom=0.8mm,
  enhanced,
]
\scriptsize\ttfamily
\setlength{\parindent}{0pt}
\raggedright
The target is 1B, an exact match in my hand. Hinting 1B maximizes similarity (score 2). Using a strategy of max similarity (exact=2 > feature match=1), with ties broken by preferring suit match over number match, this hint is unique to the target 1B: other targets prefer different hints (1A->3A, 1C->3C, 3A->3A, 3C->3C). The guesser, assuming the same strategy, will identify 1B uniquely.
\end{tcolorbox}

\begin{tcolorbox}[
  breakable,
  gptstyle,
  title={GPT-5.4 (Guesser) — hand: 3C, 3A, 1A, 1B, 1C \quad hint received: 1B \quad guess: \textbf{1A} \textrightarrow{} reward 0},
  fonttitle=\bfseries\small,
  boxrule=0.6pt,
  left=1mm,
  right=1mm,
  top=0.8mm,
  bottom=0.8mm,
  enhanced,
]
\scriptsize\ttfamily
\setlength{\parindent}{0pt}
\raggedright
A cooperative optimal code can distinguish 4 of the 5 possible targets. The most natural high-value mapping is:\\
- 3A hint -> 3A\\
- 3C hint -> 3C\\
- 2B hint -> 1B\\
- 1B hint -> one of the remaining 1X cards

\smallskip
So seeing \textbf{1B} makes \textbf{1A/1C} the most likely targets, not \textbf{1B}. With no further distinction, I use the natural tie-break toward \textbf{1A}.
\end{tcolorbox}

\end{tcolorbox}
\caption{\textbf{Convention mismatch in Hint-Guess.} The hinter gives the trivially correct exact-match hint \texttt{1B} for target \texttt{1B}. The guesser, however, applies an over-engineered bijective code in which \texttt{1B} is re-mapped to a residual slot (\texttt{1A}).}
\label{trace:hint_guess_grok_gpt_failure}
\end{figure*}

\begin{figure*}[t]
\centering
\begin{tcolorbox}[
  colback=white,
  colframe=black!70,
  sharp corners,
  boxrule=0.8pt,
  width=\textwidth,
  left=1mm,
  right=1mm,
  top=1mm,
  bottom=1mm,
  before skip=6pt,
  after skip=6pt,
  enhanced,
]

\begin{tcolorbox}[
  breakable,
  opusstyle,
  title={Claude Sonnet 4.6 (Hinter) — hand: 1B, 1C, 1C, 2A, 2C \quad target: 1C \quad hint: \textbf{1C}},
  fonttitle=\bfseries\small,
  boxrule=0.6pt,
  left=1mm,
  right=1mm,
  top=0.8mm,
  bottom=0.8mm,
  enhanced,
]
\scriptsize\ttfamily
\setlength{\parindent}{0pt}
\raggedright
The target card is 1C, and I need the Guesser to pick 1C from their hand.

The Guesser's hand contains: 1C, 1C, 2B, 2A, 3A.

If I hint 1C (which is in my hand), the Guesser will see that my hint card exactly matches 1C, which is also in their hand. The most natural cooperative strategy is: if the hint card appears in the Guesser's hand, pick that card. Since 1C is in both my hand and the Guesser's hand, hinting 1C directly and unambiguously signals that the target is 1C.

Other hints would be less direct:\\
- hint\_1B: shares ``1'' with 1C but also ``B'' with 2B --- ambiguous\\
- hint\_2A: shares features with 2A, 3A, or 2B --- points away from 1C\\
- hint\_2C: shares ``C'' with 1C but ``2'' with 2B/2A --- ambiguous

Hinting 1C is the clearest possible signal.
\end{tcolorbox}

\begin{tcolorbox}[
  breakable,
  grokstyle,
  title={Grok 4.1 Fast (Guesser) — hand: 1C, 1C, 2B, 2A, 3A \quad hint received: 1C \quad guess: \textbf{2A} \textrightarrow{} reward 0},
  fonttitle=\bfseries\small,
  boxrule=0.6pt,
  left=1mm,
  right=1mm,
  top=0.8mm,
  bottom=0.8mm,
  enhanced,
]
\scriptsize\ttfamily
\setlength{\parindent}{0pt}
\raggedright
Unique card types in my hand, sorted lexicographically: 1C, 2A, 2B, 3A.\\
Unique card types in Hinter's hand, sorted lexicographically: 1B, 1C, 2A, 2C.\\
Use position-based mapping: 1st hint signals 1st guess, 2nd hint signals 2nd guess, etc.\\
Hinter hinted 1C (2nd in sorted list), so guess 2A (2nd in my sorted list).\\
This bijection allows perfect signaling in general.
\end{tcolorbox}

\end{tcolorbox}
\caption{\textbf{Convention mismatch in Hint-Guess.} The hinter uses a simple exact-match strategy: hint \texttt{1C} because \texttt{1C} is in both hands. The guesser ignores the direct exact-match signal and instead applies a lexicographic position-based bijection, mapping the 2nd hint in the sorted hinter hand (\texttt{1C}) to the 2nd card in the sorted guesser hand (\texttt{2A}).}
\label{trace:hint_guess_sonnet_grok_failure}
\end{figure*}

\begin{figure*}[t]
\centering
\begin{tcolorbox}[
  colback=white,
  colframe=black!70,
  sharp corners,
  boxrule=0.8pt,
  width=\textwidth,
  left=1mm,
  right=1mm,
  top=1mm,
  bottom=1mm,
  before skip=6pt,
  after skip=6pt,
  enhanced,
]

\begin{tcolorbox}[
  breakable,
  gptstyle,
  title={GPT-5.4},
  fonttitle=\bfseries\small,
  boxrule=0.6pt,
  left=1mm,
  right=1mm,
  top=0.8mm,
  bottom=0.8mm,
  enhanced,
]
\scriptsize\ttfamily
\setlength{\parindent}{0pt}
\raggedright
Use a deterministic canonical ordering over cards (1A < 1B < \ldots{} < 3C) and ignore hand permutation by deduplicating and sorting both public hands.

Then coordinate via aligned indices:

\textbf{Hinter:} find the target's index in the sorted unique Guesser hand, and hint the Hinter card at the same index modulo the number of unique Hinter cards.\\
\textbf{Guesser:} recover the hinted card's index in the sorted unique Hinter hand, then guess the Guesser card(s) whose index has the same modulo class; choose the first such card.

Why this is coordination-friendly:
\begin{itemize}
  \item Fully deterministic.
  \item Uses only public information plus the hidden target.
  \item Independent of arbitrary hand order.
  \item Very simple and likely to be rediscovered by another policy.
  \item Perfectly identifies the target whenever the Hinter has at least as many unique hint cards as the Guesser has unique candidate cards.
\end{itemize}
\end{tcolorbox}

\end{tcolorbox}
\caption{\textbf{Reasoning trace of GPT-5.4 as a code agent in Hint Guess.} The policy sorts both hands canonically (1A $<$ 1B $<$ \ldots $<$ 3C), deduplicates, and coordinates by mapping target indices modulo the number of unique hinter cards. Although internally consistent, the convention is essentially arbitrary — another code agent is unlikely to independently arrive at the same convention, hence cross-play coordination breaks down.}
\label{trace:hint_guess_gpt_canonical_index}
\end{figure*}

\clearpage
\section*{NeurIPS Paper Checklist}

The checklist is designed to encourage best practices for responsible machine learning research, addressing issues of reproducibility, transparency, research ethics, and societal impact. Do not remove the checklist: {\bf The papers not including the checklist will be desk rejected.} The checklist should follow the references and follow the (optional) supplemental material.  The checklist does NOT count towards the page
limit. 

Please read the checklist guidelines carefully for information on how to answer these questions. For each question in the checklist:
\begin{itemize}
    \item You should answer \answerYes{}, \answerNo{}, or \answerNA{}.
    \item \answerNA{} means either that the question is Not Applicable for that particular paper or the relevant information is Not Available.
    \item Please provide a short (1--2 sentence) justification right after your answer (even for \answerNA). 
\end{itemize}

{\bf The checklist answers are an integral part of your paper submission.} They are visible to the reviewers, area chairs, senior area chairs, and ethics reviewers. You will also be asked to include it (after eventual revisions) with the final version of your paper, and its final version will be published with the paper.

The reviewers of your paper will be asked to use the checklist as one of the factors in their evaluation. While \answerYes{} is generally preferable to \answerNo{}, it is perfectly acceptable to answer \answerNo{} provided a proper justification is given (e.g., error bars are not reported because it would be too computationally expensive'' or ``we were unable to find the license for the dataset we used''). In general, answering \answerNo{} or \answerNA{} is not grounds for rejection. While the questions are phrased in a binary way, we acknowledge that the true answer is often more nuanced, so please just use your best judgment and write a justification to elaborate. All supporting evidence can appear either in the main paper or the supplemental material, provided in appendix. If you answer \answerYes{} to a question, in the justification please point to the section(s) where related material for the question can be found.

IMPORTANT, please:
\begin{itemize}
    \item {\bf Delete this instruction block, but keep the section heading ``NeurIPS Paper Checklist"},
    \item  {\bf Keep the checklist subsection headings, questions/answers and guidelines below.}
    \item {\bf Do not modify the questions and only use the provided macros for your answers}.
\end{itemize}


\begin{enumerate}

\item {\bf Claims}
    \item[] Question: Do the main claims made in the abstract and introduction accurately reflect the paper's contributions and scope?
    \item[] Answer: \answerYes{} 
    \item[] Justification: Empirical claims are supported by experiments in \autoref{sec:zsc_performance}, \autoref{sec:semantics_impact}, and~\autoref{sec:cfd}, with additional results in \appref{sec:app/code_agents_with_cfd}. Non-empirical claims, such as the relation to prior work, are addressed in \appref{sec:relatedwork}; out-of-scope items and limitations are discussed in \autoref{sec:conclusion}.

\item {\bf Limitations}
    \item[] Question: Does the paper discuss the limitations of the work performed by the authors?
    \item[] Answer: \answerYes{} 
    \item[] Justification: We discuss limitations in \autoref{sec:conclusion}.

\item {\bf Theory assumptions and proofs}
    \item[] Question: For each theoretical result, does the paper provide the full set of assumptions and a complete (and correct) proof?
    \item[] Answer: \answerNA{} 
    \item[] Justification: The paper does not contain any theoretical results.

    \item {\bf Experimental result reproducibility}
    \item[] Question: Does the paper fully disclose all the information needed to reproduce the main experimental results of the paper to the extent that it affects the main claims and/or conclusions of the paper (regardless of whether the code and data are provided or not)?
    \item[] Answer: \answerYes{} 
    \item[] Justification: Environments are described in \appref{app:environments}, prompts are given in \appref{app:prompt}, and the remaining implementation details required to reproduce our results are provided in \appref{sec:app/implementation_details}.

\item {\bf Open access to data and code}
    \item[] Question: Does the paper provide open access to the data and code, with sufficient instructions to faithfully reproduce the main experimental results, as described in supplemental material?
    \item[] Answer: \answerYes{}{} 
    \item[] Justification: Anonymised code with run instructions to reproduce all experiments is available at \url{https://anonymous.4open.science/r/zsc_llm-EAE1/README.md}.

\item {\bf Experimental setting/details}
    \item[] Question: Does the paper specify all the training and test details (e.g., data splits, hyperparameters, how they were chosen, type of optimizer) necessary to understand the results?
    \item[] Answer: \answerYes{} 
    \item[] Justification: Experimental settings and hyperparameters are described in \autoref{sec:zsc_performance}, and full implementation details for the meta-learning optimisation are provided in \appref{sec:app/implementation_details}.

\item {\bf Experiment statistical significance}
    \item[] Question: Does the paper report error bars suitably and correctly defined or other appropriate information about the statistical significance of the experiments?
    \item[] Answer: \answerYes{} 
    \item[] Justification: We report the mean and standard error over multiple random seeds for all results; the protocol used to compute these statistics is detailed in \appref{sec:app/code_agents_with_cfd}.

\item {\bf Experiments compute resources}
    \item[] Question: For each experiment, does the paper provide sufficient information on the computer resources (type of compute workers, memory, time of execution) needed to reproduce the experiments?
    \item[] Answer: \answerYes{} 
    \item[] Justification: API costs and the GPU used for the token-level probability experiments are reported in \appref{sec:app/experimental_resources}.
    
\item {\bf Code of ethics}
    \item[] Question: Does the research conducted in the paper conform, in every respect, with the NeurIPS Code of Ethics \url{https://neurips.cc/public/EthicsGuidelines}?
    \item[] Answer: \answerYes{} 
    \item[] Justification: The research conforms to the NeurIPS Code of Ethics.

\item {\bf Broader impacts}
    \item[] Question: Does the paper discuss both potential positive societal impacts and negative societal impacts of the work performed?
    \item[] Answer: \answerYes{} 
    \item[] Justification: Broader impacts are discussed in \appref{sec:app/impact_statement}.
    
\item {\bf Safeguards}
    \item[] Question: Does the paper describe safeguards that have been put in place for responsible release of data or models that have a high risk for misuse (e.g., pre-trained language models, image generators, or scraped datasets)?
    \item[] Answer: \answerNA{} 
    \item[] Justification: This paper poses no such risks.

\item {\bf Licenses for existing assets}
    \item[] Question: Are the creators or original owners of assets (e.g., code, data, models), used in the paper, properly credited and are the license and terms of use explicitly mentioned and properly respected?
    \item[] Answer: \answerYes{} 
    \item[] Justification: Licenses for the figures and environments reused from prior work are listed in \appref{sec:app/licenses}.

\item {\bf New assets}
    \item[] Question: Are new assets introduced in the paper well documented and is the documentation provided alongside the assets?
    \item[] Answer: \answerYes{} 
    \item[] Justification: We introduce seven new coordination environments (Narrow Tunnel, Brittle Bridge, Ad-Hoc Conventions, Ad-Hoc Conventions with Permuted Signals, Permuted Levers, Semantic Selection Game, and Semantic Convention Game), all provided with documentation in the anonymised code repository (\url{https://anonymous.4open.science/r/zsc_llm-EAE1/README.md}). The same repository also includes our re-implementations of the reused environments (Cat \& Dog, Tiny Hanabi, Hanabi 3$\times$3, Matrix Game, Hint--Guess, and Trampoline--Tiger), each augmented with natural-language descriptions for interaction with \acs{llm} agents.

\item {\bf Crowdsourcing and research with human subjects}
    \item[] Question: For crowdsourcing experiments and research with human subjects, does the paper include the full text of instructions given to participants and screenshots, if applicable, as well as details about compensation (if any)?
    \item[] Answer: \answerNA{} 
    \item[] Justification: The paper does not involve crowdsourcing or human subjects.

\item {\bf Institutional review board (IRB) approvals or equivalent for research with human subjects}
    \item[] Question: Does the paper describe potential risks incurred by study participants, whether such risks were disclosed to the subjects, and whether Institutional Review Board (IRB) approvals (or an equivalent approval/review based on the requirements of your country or institution) were obtained?
    \item[] Answer: \answerNA{} 
    \item[] Justification: The paper does not involve human subjects.

\item {\bf Declaration of LLM usage}
    \item[] Question: Does the paper describe the usage of LLMs if it is an important, original, or non-standard component of the core methods in this research? Note that if the LLM is used only for writing, editing, or formatting purposes and does \emph{not} impact the core methodology, scientific rigor, or originality of the research, declaration is not required.
    \item[] Answer: \answerYes{} 
    \item[] Justification: \acsp{llm} are central to this work in two ways: (i) they are the subject of study, evaluated in the sections \textit{ZSC performance of LLM agents}, \textit{Semantics and Coordination}, and \textit{Coordination Friendly Definitions}; and (ii) an \acs{llm} (GPT-5.4) serves as the meta-optimisation agent in the GEPA-based loop that discovers \acsp{cfd} (\textit{Coordination Friendly Definitions}, with implementation details in \appref{sec:app/implementation_details}). The full list of \acsp{llm} accessed in this work and their licenses is provided in \appref{sec:app/licenses}.

\end{enumerate}

\end{document}